%% file: main.tex
\documentclass{article}
\usepackage[margin=1in]{geometry}

\usepackage{graphicx}
\usepackage{booktabs}
\usepackage{amsmath}
\usepackage{amssymb}
\usepackage{amsfonts}
\usepackage{multirow}
\usepackage{verbatim}
\usepackage{caption}
\usepackage{longtable}
\usepackage{supertabular}
\usepackage{float}

\usepackage{enumitem}
\usepackage{tablefootnote}
\usepackage[round,semicolon]{natbib}
\usepackage{xcolor}
\usepackage{xspace}
\usepackage{textcomp}
\usepackage{makecell}
\usepackage{multirow}
\usepackage{lscape} 
\usepackage{siunitx}

\usepackage{amssymb}
\usepackage{pifont}
\usepackage{scrextend}

\usepackage{array}
\usepackage{tgpagella}
\usepackage{latexsym}
\usepackage[T1]{fontenc}
\usepackage[utf8]{inputenc}
\usepackage{microtype}
\definecolor{mydarkblue}{rgb}{0,0.08,0.45}
\usepackage[colorlinks,citecolor=mydarkblue,urlcolor=mydarkblue,linkcolor=mydarkblue]{hyperref}
\usepackage{url}            
\usepackage{nicefrac}       
\usepackage{changepage}
\usepackage{xargs}          
\usepackage{wrapfig,lipsum,booktabs}
\usepackage{longtable}
\usepackage{endnotes}

\usepackage{pgfplots}
\usetikzlibrary{pgfplots.groupplots}
\pgfplotsset{compat=1.3}
\usepackage{tikz}
\usetikzlibrary{patterns}

\usepackage[most]{tcolorbox}

\usepackage[capitalize,noabbrev]{cleveref}
\crefname{section}{Section}{\S\S}
\Crefname{section}{Section}{\S\S}
\crefname{table}{Table}{Tables}
\crefname{figure}{Figure}{Figures}
\crefname{algorithm}{Algorithm}{}
\crefname{equation}{eq.}{}
\crefname{appendix}{Appendix}{}
\crefformat{section}{Section #2#1#3}
\usepackage{multicol}

\usepackage{minitoc}

\title{
\vspace{-10mm}
\textbf{
Scaling Vision Transformers to 22 Billion Parameters
}
\vspace{-3mm}
  }

\author{
\normalsize{}
\textbf{Mostafa Dehghani\thanks{Core contributors. Correspondence: \url{dehghani@google.com}}} \hspace{5mm} 
\textbf{Josip Djolonga\footnotemark[1]} \hspace{3mm} 
\textbf{Basil Mustafa\footnotemark[1]} \hspace{3mm} 
\textbf{Piotr Padlewski\footnotemark[1]} \hspace{3mm} 
\textbf{Jonathan Heek\footnotemark[1]} 
\\
\normalsize{}
\textbf{Justin Gilmer} \hspace{3mm} 
\textbf{Andreas Steiner} \hspace{3mm} 
\textbf{Mathilde Caron} \hspace{3mm} 
\textbf{Robert Geirhos} \hspace{3mm} 
\textbf{Ibrahim  Alabdulmohsin}

\\
\normalsize{}
\textbf{Rodolphe Jenatton} \hspace{3mm} 
\textbf{Lucas Beyer} \hspace{3mm} 
\textbf{Michael Tschannen} \hspace{3mm} 
\textbf{Anurag Arnab} \hspace{3mm} 
\textbf{Xiao Wang}
\\
\normalsize{}
\textbf{Carlos Riquelme} \hspace{3mm} 
\textbf{Matthias Minderer} \hspace{3mm} 
\textbf{Joan Puigcerver} \hspace{3mm} 
\textbf{Utku Evci} \hspace{3mm} 
\textbf{Manoj Kumar}
\\
\normalsize{}
\textbf{Sjoerd van Steenkiste} \hspace{3mm} 
\textbf{Gamaleldin F. Elsayed} \hspace{3mm} 
\textbf{Aravindh Mahendran} \hspace{3mm} 
\textbf{Fisher Yu}
\\
\normalsize{}
\textbf{Avital Oliver} \hspace{3mm} 
\textbf{Fantine Huot} \hspace{3mm} 
\textbf{Jasmijn Bastings} \hspace{3mm} 
\textbf{Mark Patrick Collier} \hspace{3mm} 
\textbf{Alexey A. Gritsenko}
\\
\normalsize{}
\textbf{Vighnesh Birodkar} \hspace{3mm} 
\textbf{Cristina Vasconcelos} \hspace{3mm} 
\textbf{Yi Tay} \hspace{3mm} 
\textbf{Thomas Mensink} \hspace{3mm} 
\textbf{Alexander Kolesnikov}
\\
\normalsize{}
\textbf{Filip Pavetić} \hspace{3mm} 
\textbf{Dustin Tran} \hspace{3mm} 
\textbf{Thomas Kipf} \hspace{3mm} 
\textbf{Mario Lučić} \hspace{3mm} 
\textbf{Xiaohua Zhai} \hspace{3mm} 
\textbf{Daniel Keysers}
\\
\normalsize{}
\textbf{Jeremiah Harmsen} \hspace{3mm} 
\textbf{Neil Houlsby\footnotemark[1]}
\\ 
\vspace{7mm}
\normalsize{}
Google Research
\vspace{-7mm}
}

\date{}

\usepackage{microtype}
\usepackage{graphicx}
\usepackage{booktabs} 
\usepackage{hyperref}
\usepackage{siunitx}
\usepackage{amsmath}
\usepackage{amssymb}
\usepackage{mathtools}
\usepackage{amsthm}
\usepackage{fontawesome5}
\usepackage{longtable}
\usepackage{multirow}
\usepackage{subfigure}
\usepackage{caption}
\usepackage{xfrac}
\usepackage{setspace}

\usepackage{wrapfig}
\usepackage{tabulary}
\usepackage{tabularx}
\usepackage[export]{adjustbox}

\usepackage[capitalize,noabbrev]{cleveref}

\newcommand{\chonk}{\mbox{ViT-22B}\xspace}
\usepackage{xspace}
\usepackage{xcolor}

\begin{document}
\setlength{\abovedisplayskip}{4pt}
\setlength{\belowdisplayskip}{4pt}
\setlength{\abovedisplayshortskip}{0pt}
\setlength{\belowdisplayshortskip}{0pt}

\doparttoc 
\faketableofcontents 
\maketitle

\input{arxiv/text/0-abstract}
\input{arxiv/text/1-introduction}

\input{arxiv/text/2-model}
\input{arxiv/text/3-infrastructure}
\input{arxiv/text/4-0-evaluation}

\input{arxiv/text/5-conclusion}

\section*{Acknowledgment}
We would like to thank Jasper Uijlings, Jeremy Cohen, Arushi Goel, Radu Soricut, Xingyi Zhou, Lluis Castrejon, Adam Paszke, Joelle Barral, Federico Lebron, Blake Hechtman, and Peter Hawkins. Their expertise and unwavering support played a crucial role in the completion of this paper. We also acknowledge the collaboration and dedication of the talented researchers and engineers at Google Research.

\bibliography{ref}
\bibliographystyle{plainnat}

\input{arxiv/text/6-appendix}

\end{document}

%% file: arxiv/text/0-abstract.tex
\begin{abstract}

The scaling of Transformers has driven breakthrough capabilities for language models.
At present, the largest large language models (LLMs) contain upwards of 100B parameters.
Vision Transformers (ViT) have introduced the same architecture to image and video modelling, but these have not yet been successfully scaled to nearly the same degree; the largest dense ViT contains 4B parameters~\citep{chen2022pali}.
We present a recipe for highly efficient and stable training of a 22B-parameter ViT (\chonk) and perform a wide variety of experiments on the resulting model.
When evaluated on downstream tasks (often with a lightweight linear model on frozen features), \chonk demonstrates increasing performance with scale.
We further observe other interesting benefits of scale, including an improved tradeoff between fairness and performance, state-of-the-art alignment to human visual perception in terms of shape/texture bias, and improved robustness.
\chonk demonstrates the potential for ``LLM-like'' scaling in vision, and provides key steps towards getting there.
\end{abstract}

%% file: arxiv/text/1-introduction.tex
\section{Introduction}
Similar to natural language processing, transfer of pre-trained vision backbones has improved performance on a wide variety of vision tasks~\citep{pan2009survey,zhai2019large,kolesnikov2020big}.
Larger datasets, scalable architectures, and new training methods~\citep{mahajan2018exploring,dosovitskiy2020image,clip,zhai2022scaling} have accelerated this growth. Despite this, vision models have trailed far behind language models, which have demonstrated emergent capabilities at massive scales~\citep{chowdhery2022palm, wei2022emergent}. Specifically, the largest dense vision model to date is a mere $4$B parameter ViT~\citep{chen2022pali}, while a modestly parameterized model for an entry-level competitive language model typically contains over 10B parameters~\citep{t5paper,tay2022unifying,chung2022scaling}, and the largest dense language model has 540B parameters~\citep{chowdhery2022palm}. Sparse models demonstrate the same trend, where language models go beyond a trillion parameters~\citep{fedus2021switch} but the largest reported sparse vision models are only 15B~\citep{riquelme2021scaling}.


This paper presents~\chonk{}, the largest dense ViT model to date. En route to 22B parameters, we uncover pathological training instabilities which prevent scaling the default recipe, and demonstrate architectural changes which make it possible. Further, we carefully engineer the model to enable model-parallel training at unprecedented efficiency.
\chonk{}'s quality is assessed via a comprehensive evaluation suite of tasks, ranging from (few-shot) classification to dense output tasks, where it reaches or advances the current state-of-the-art.
For example, even when used as a frozen visual feature extractor, \chonk achieves an accuracy of 89.5\% on ImageNet.
With a text tower trained to match these visual features~\citep{lit}, it achieves 85.9\% accuracy on ImageNet in the zero-shot setting.
The model is furthermore a great teacher --- used as a distillation target, we train a ViT-B student that achieves 88.6\% on ImageNet, state-of-the-art at this scale.

This performance comes with improved out of distribution behaviour, reliability, uncertainty estimation and fairness tradeoffs.
Finally, the model's features are better aligned with humans perception, achieving previously unseen shape bias of 87\%.

%% file: arxiv/text/2-model.tex
\section{Model Architecture}
\label{sec:model_architecture}

\chonk is a Transformer-based encoder model that resembles the architecture of the original Vision Transformer~\citep{dosovitskiy2020image} but incorporates the following three main modifications to improve efficiency and training stability at scale: parallel layers, query/key (QK) normalization, and omitted biases.

\paragraph{Parallel layers.}
As in \citet{gpt-j}, \chonk applies the Attention and MLP blocks in parallel, instead of sequentially as in the standard Transformer:
 \begin{align*} 
 y' &= \text{LayerNorm}(x),\\
 y\hphantom{'} &= x + \text{MLP}(y')+\text{Attention}(y').
 \end{align*}
This enables additional parallelization via combination of linear projections from the MLP and attention blocks. In particular, the matrix multiplication for query/key/value-projections and the first linear layer of the MLP are fused into a single operation, and the same is done for the attention out-projection and second linear layer of the MLP. This approach is also used by PaLM~\citep{chowdhery2022palm}, where this technique sped up the largest model's training by 15\% without performance degradation. 
    
\begin{figure}[tbp]
    \centering
    \includegraphics[width=\textwidth]{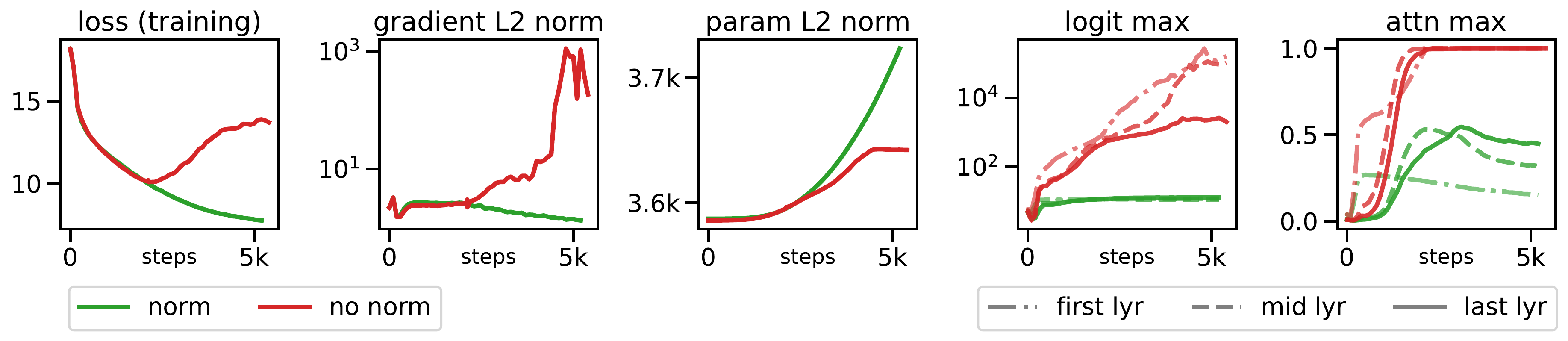}
    \caption{Effect of query/key normalization on an 8B parameter model.}
    \label{fig:stability_8b}
\end{figure}

\paragraph{QK Normalization.}
In scaling ViT beyond prior works, we observed divergent training loss after a few thousand steps. In particular, this instability was observed for models with around 8B parameters (see \cref{app:scalability}). It was caused by extremely large values in attention logits, which lead to (almost one-hot) attention weights with near-zero entropy.
To solve this, we adopt the approach of \citet{gilmer2023intriguing}, which applies LayerNorm~\citep{ba2016layer} to the queries and keys before the dot-product attention computation.
Specifically, the attention weights are computed as
\begin{align*}
\text{softmax}\left[  \tfrac{1}{\sqrt{d}}
{\text{LN}(X W^Q) ( \text{LN}(X W^K))^T} \right],
\end{align*}
where $d$ is query/key dimension, $X$ is the input, $\text{LN}$ stands for layer normalization, and $W^Q$ is the query weight matrix, and $W^K$ is the key weight matrix.
The effect on an 8B parameter model is shown in \cref{fig:stability_8b}, where normalization prevents divergence due to uncontrolled attention logit growth.
    
\paragraph{Omitting biases on QKV projections and LayerNorms.}
Following PaLM~\citep{chowdhery2022palm}, the bias terms were removed from the QKV projections and all LayerNorms were applied without bias and centering~\citep{zhang2019root}. This improved accelerator utilization (by 3\%), without quality degradation. However, unlike PaLM, we use bias terms for the (in- and out-) MLP dense layers as we have observed improved quality and no speed reduction. 

\begin{figure}[t]
    \centering
    \includegraphics[width=0.6\textwidth]{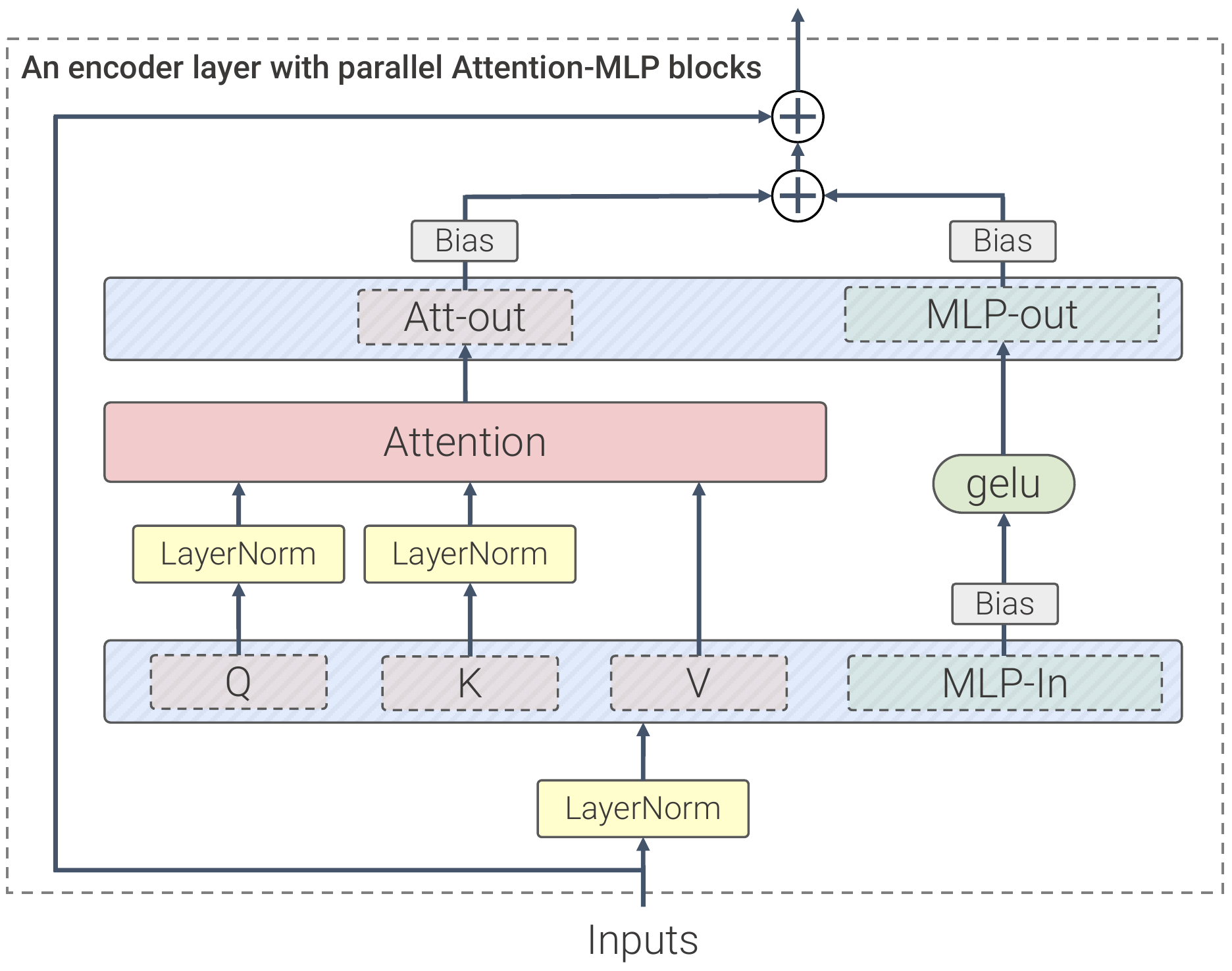}
    \caption{Parallel \chonk layer with QK normalization.}
    \label{fig:vit22b_schema}
\end{figure}

\cref{fig:vit22b_schema} illustrates a \chonk encoder block. The embedding layer, which includes extracting patches, linear projection, and the addition of position embedding follow those used in the original ViT. We use multi-head attention pooling~\citep{cordonnier2019relationship,zhai2022scaling} to aggregate the per-token representations in the head. 

\chonk is uses patch size of $14 \times 14$ with  images at resolution $224 \times 224$ (pre-processed by inception crop followed by random horizontal flip). 
Similar to the original ViT~\citep{dosovitskiy2020image}, \chonk employs a learned 1D positional embedding. 
During fine-tuning on high-resolution images (different number of visual tokens), we perform a 2D interpolation of the pre-trained position embeddings, according to their location in the original image. 

Other hyperparameters for the \chonk model architecture are presented in \cref{tbl:model_arch}, compared to the previously reported largest ViT models, ViT-G~\citep{zhai2022scaling} and ViT-e~\citep{chen2022pali}.

\begin{table}[h]
  \caption{\chonk model architecture details.}
  \centering
  \begin{tabulary}{1.0\linewidth}{@{}LCCCCR@{}}
    \toprule
    \bf{Name} & \bf{Width} & \bf{Depth} & \bf{MLP} & \bf{Heads} & \bf{Params [M]} \\
          ViT-G & 1664 & 48 & \phantom{0}8192  & 16 & 1843  \\
          ViT-e & 1792 & 56 & 15360 & 16 & 3926  \\
    \textbf{\chonk} & 6144 & 48 & 24576 & 48 & 21743 \\
    \bottomrule
  \end{tabulary}
  \label{tbl:model_arch}
\end{table}

Following the template in~\citet{mitchell2019model}, we provide the model card in \Cref{tab:model_card} (\Cref{app:model_card}).

%% file: arxiv/text/3-infrastructure.tex
\section{Training Infrastructure and Efficiency}

\chonk is implemented in JAX~\citep{jax2018github} using the FLAX library~\citep{flax2020github} and built within Scenic~\citep{dehghani2021scenic}. It leverages both model and data parallelism. In particular, we used the \texttt{jax.xmap} API, which provides explicit control over both the sharding of all intermediates (e.g.\ weights and activations) as well as inter-chip communication.
We organized the chips into a 2D logical mesh of size $t\times k$, where $t$ is the size of the data-parallel axis and $k$ is the size of the model axis.
Then, for each of the $t$ groups, $k$ devices get the same batch of images, each device keeps only $1/k$ of the activations and is responsible for computing $1/k$ of the output of all linear layers (detailed below).

\begin{figure}[ht]
    \centering
    \subfigure[The matrix $A$ is row-sharded across the devices.]{
    \includegraphics[width=0.6\textwidth]{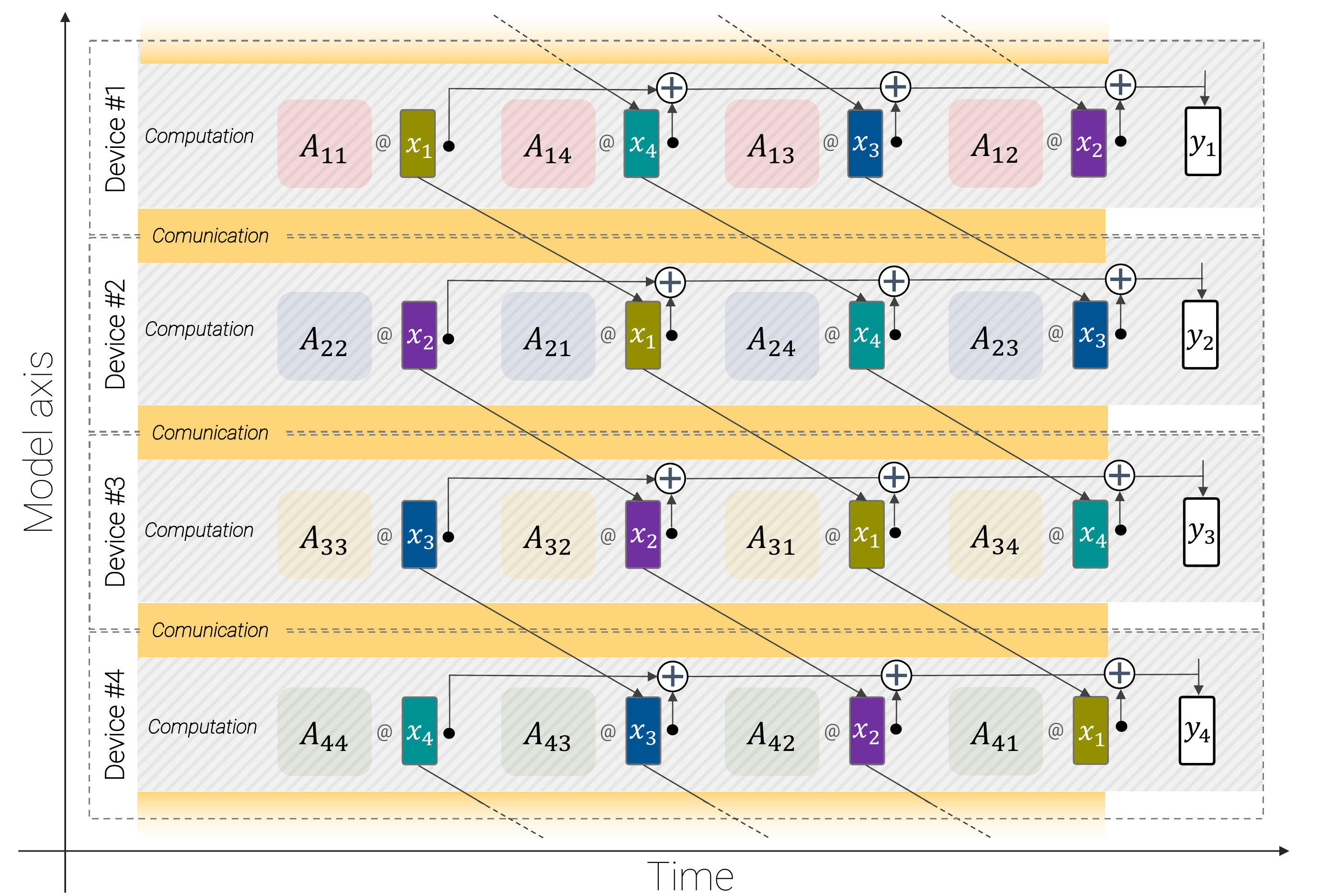}}
    \subfigure[The matrix $A$ is column-sharded across the devices.]{
    \includegraphics[width=0.6\textwidth]{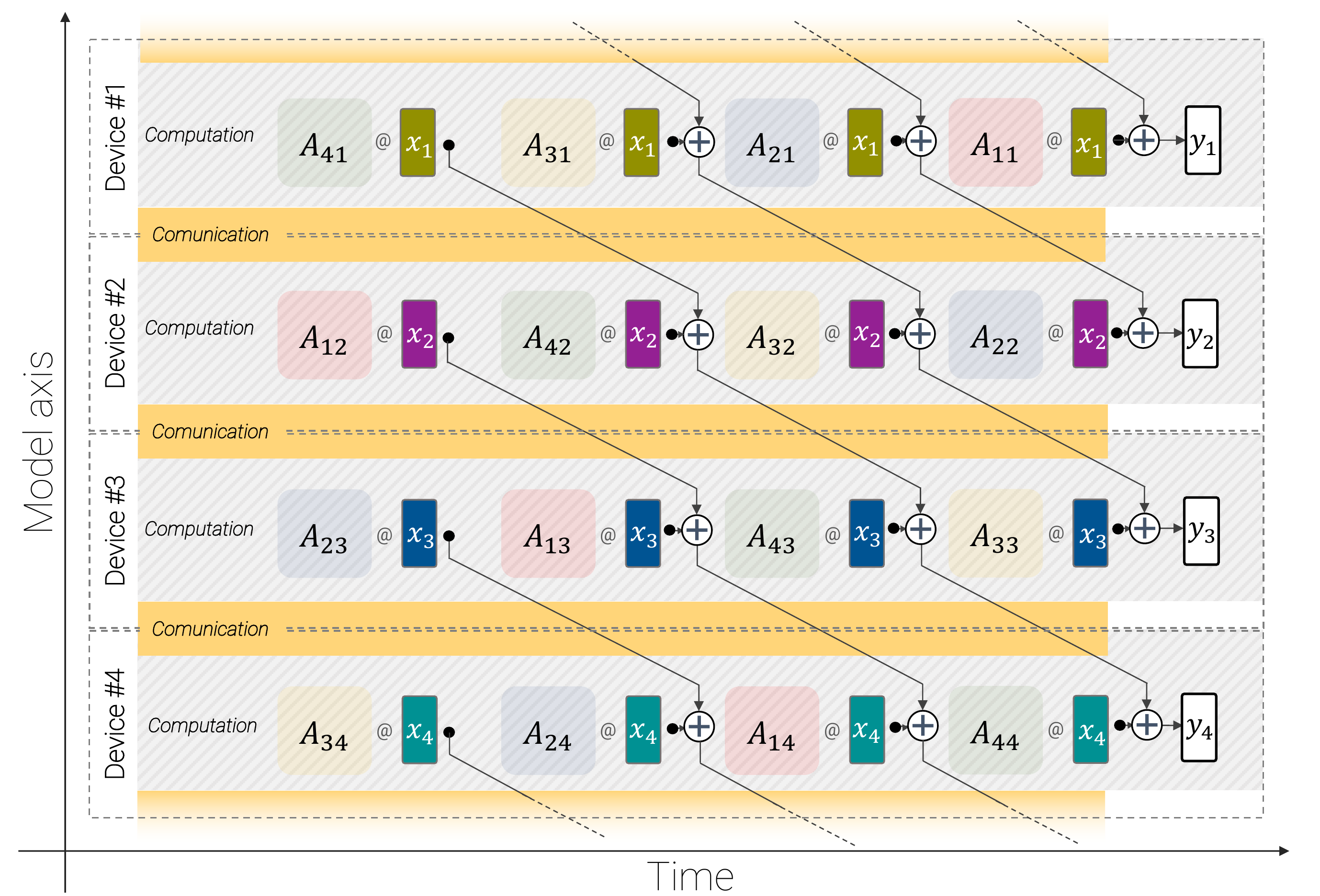}}
    \vspace{-10pt}
    \caption{Asynchronized parallel linear operation ($y=Ax$): model parallel matrix multiplication with overlapping communication and computation across devices.}
    \label{fig:vit22b_async_linear}
\end{figure}

\paragraph{Asynchronous parallel linear operations.}
As we use explicit sharding, we built a wrapper around the dense layers in FLAX that adapts them to the setting where their inputs are split across $k$ devices.
To maximize throughput, two aspects have to be considered --- computation and communication.
Namely, we want the operations to be analytically equivalent to the unsharded case, to communicate as little as possible, and ideally to have them overlap~\citep{wang2022overlap} so that we can keep the matrix multiply unit, where most of the FLOP capacity is, busy at all times.

To illustrate the process, consider the problem of computing $y=Ax$ under the constraint that the $i$-th block of $x$ and $y$ both reside  on the $i$-th device. We denote the blocks of $A\in\mathbb{R}^{m\times n}$ by $A_{i,j}\in\mathbb{R}^{\frac{m}{k}\times\frac{n}{k}}$, and analogously $x_i\in\mathbb{R}^{\frac{n}{k}}$ and  $y_j\in\mathbb{R}^{\frac{m}{k}}$, with $i,j\in\{1,\ldots,k\}$.
The first option is to have device $i$ hold the $i$-th block of rows, necessary for computation of $y_i$, so that to compute $y_i$ the chip needs to communicate $k-1$ times to complete $x$, a total of $(k-1)(n/k)$ floats.
Alternatively, device $i$ can hold the $i$-th block of columns, all acting on $x_i$.
This way, the device computes the vectors $y_{ji}=A_{ji}x_i$, which have to be communicated (scatter-reduced) with the other devices.
Note that here the communicated vectors belong to the output space, a total of $(k-1)(m/k)$ floats.
This \emph{asymmetry} is leveraged in communication costs when $n\neq m$; column-sharding is used in the computation of the output of the MLP in a Transformer, where $n=4m$, and row-sharding elsewhere.

Furthermore, matrix multiplications are overlapped with the communication with the neighbours.
This asynchronous approach allows for high matrix core utilization and increased device efficiency, while minimizing waiting on incoming communication.
\Cref{fig:vit22b_async_linear} presents the overlapping communication and computation across $4$ devices with the parallel linear operation in row-sharding and column-sharding modes.
The general case of this technique is presented in \citet{wang2022overlap}, who
also introduce the XLA operations we leverage here.

\paragraph{Parameter sharding.}
The model is data-parallel on the first axis.
Each parameter can be either fully replicated over this axis, or have each device hold a chunk of it.
We opted to shard some large tensors from the model parameters to be able to fit larger models and batch sizes.
This means that the device would have to gather the parameters before computing of the forward and scatter on the backward pass, but again, note that this happens asynchronous with computation.
In particular, while computing one layer the device can start communicating the weights of the next one, thus minimizing the communication overhead.

Using these techniques, \chonk processes $1.15$k tokens per second per core during training (forward and backward pass) on TPUv4~\citep{jouppi2020domain}. 
\chonk's model flops utilization (MFU)~\citep{chowdhery2022palm,dehghani2021efficiency} is \emph{54.9\%}, indicating a very efficient use of the hardware. Note that PaLM reports 46.2\% MFU~\citep{chowdhery2022palm, pope2022efficiently} and we measured 44.0\% MFU for ViT-e (data-parallel only) on the same hardware.

%% file: arxiv/text/4-0-evaluation.tex
\section{Experiments}
\label{sec:evaluation}

\subsection{Training details}
\label{sec:training}
\paragraph{Dataset.}
\chonk is trained on a version of JFT~\citep{sun2017revisiting}, extended to around 4B images~\citep{zhai2022scaling}.
These images have been semi-automatically annotated with a class-hierarchy of 30k labels.
Following the original Vision Transformer, we flatten the hierarchical label structure and use all the assigned labels in a multi-label classification fashion employing the sigmoid cross-entropy loss.

\paragraph{Hyperparameters.}
\chonk was trained using 256 visual tokens per image, where each token represents a $14 \times 14$ patch extracted from $224 \times 224$ sized images. 
\chonk is trained for 177k steps with batch size of 65k: approximately 3 epochs.
We use a reciprocal square-root learning rate schedule with a peak of $10^{-3}$, and linear warmup (first 10k steps) and cooldown (last 30k steps) phases. For better few-shot adaptation, we use a higher weight decay on the head ($3.0$) than body ($0.03$) for upstream training~\citep{zhai2022scaling, abnar2021exploring}.

\input{arxiv/text/4-1-eval_transfer_classification}
\input{arxiv/text/4-2-eval_transfer_dense_prediction}

\input{arxiv/text/4-3-eval_transfer_video}

\input{arxiv/text/4-4-eval_beyond_accuracy}

%% file: arxiv/text/4-1-eval_transfer_classification.tex
\subsection{Transfer to image classification}
Efficient transfer learning with large scale backbones is often achieved by using them as frozen feature extractors. This section presents the evaluation results of \chonk for image classification using linear probing and locked-image tuning as well as out-of-distribution transfer. Additional results for  Head2Toe transfer, few-shot transfer, and linear probing with L-BFGS can be found in \cref{app:image_classification}.

\subsubsection{Linear probing}


We explored various ways of training a linear probe, 
our final setup on ImageNet uses SGD with momentum for 10 epochs at 224px resolution, with mild random cropping and horizontal flipping as the only data augmentations, and no further regularizations.

The results presented in \cref{tab:linear_probe_i1k} show that while the returns are diminishing, there is still a notable improvement at this scale\footnote{We repeated a subset of the experiments multiple times and the results are almost identical.}.
Furthermore, we show that linear probing of larger models like \chonk can approach or exceed performance of full fine-tuning of smaller models with high-resolution, which can be often cheaper or easier to do.

\begin{table}[ht]
    \caption{Linear evaluation on ImageNet-1k~\citep{deng2009imagenet} with varying scale. All models pre-trained on large datasets. 
    Performances of a few high-resolution fine-tuned models from are provided for reference.}
    \centering
    \setlength{\tabcolsep}{4pt}
    \resizebox{.6\columnwidth}{!}{%
    \begin{tabular}{@{} l l c c c c c @{}}
        \toprule
        Model & IN & ReaL & INv2 & ObjectNet & IN-R & IN-A \\
        \midrule
        \multicolumn{7}{l}{\textit{224px linear probe (frozen)}} \\
        \midrule
        B/32 & 80.18 & 86.00 & 69.56 & 46.03 & 75.03 & 31.2 \\
        B/16 & 84.20 & 88.79 & 75.07 & 56.01 & 82.50 & 52.67 \\
        ALIGN (360px) & 85.5 & - & - & - & - & - \\
        L/16 & 86.66 & 90.05 & 78.57 & 63.84 & 89.92 & 67.96 \\
        g/14 & 88.51 & 90.50 & 81.10 & 68.84 & 92.33 & 77.51 \\
        G/14 & 88.98 & 90.60 & 81.32 & 69.55 & 91.74 & 78.79 \\
        e/14 & 89.26 & 90.74 & 82.51 & 71.54 & \textbf{94.33} & 81.56 \\
        22B  & \textbf{89.51} & \textbf{90.94} & \textbf{83.15} & \textbf{74.30} & 94.27 & \textbf{83.80} \\
        \midrule
        \multicolumn{7}{l}{\textit{High-res fine-tuning}} \\
        \midrule
        L/16 & 88.5 & 90.4 & 80.4 & - & - & - \\
        FixNoisy-L2 & 88.5 & 90.9 & 80.8 & - & - & - \\
        ALIGN-L2 & 88.64 & - & - & - & - & - \\
        MaxViT-XL & 89.53 & - & - & - & - & - \\
        G/14 & 90.45 & 90.81 & 83.33 & 70.53 & - & - \\
        e/14 & 90.9 & 91.1 & 84.3 & 72.0 & - & - \\
        \bottomrule
    \end{tabular}
    }
    \label{tab:linear_probe_i1k}
\end{table}

\begin{figure}[tbp]
    \centering
    \subfigure{\includegraphics[width=0.35\textwidth]{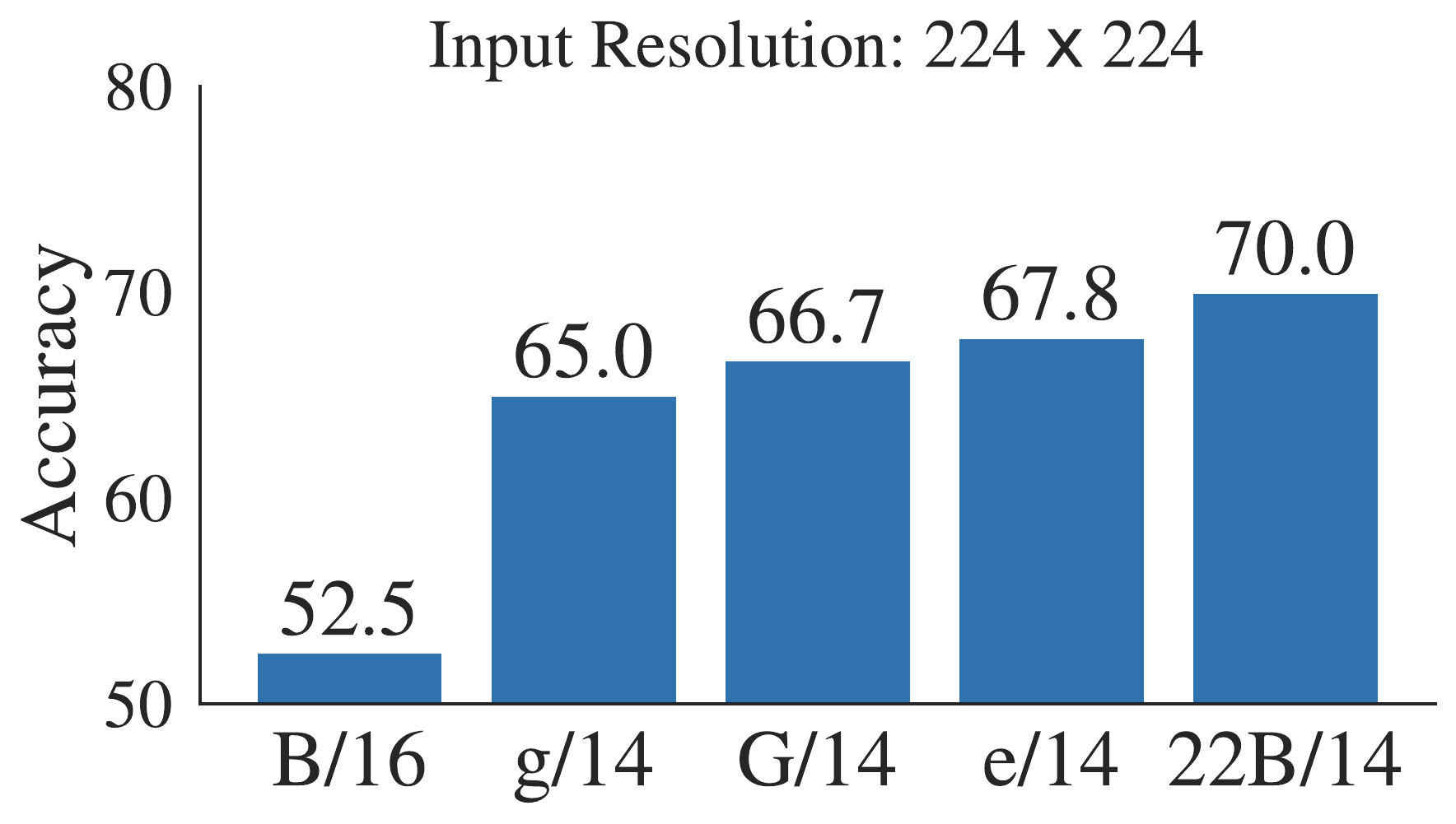}}
    \hspace{15pt} 
    \subfigure{\includegraphics[width=0.35\textwidth]{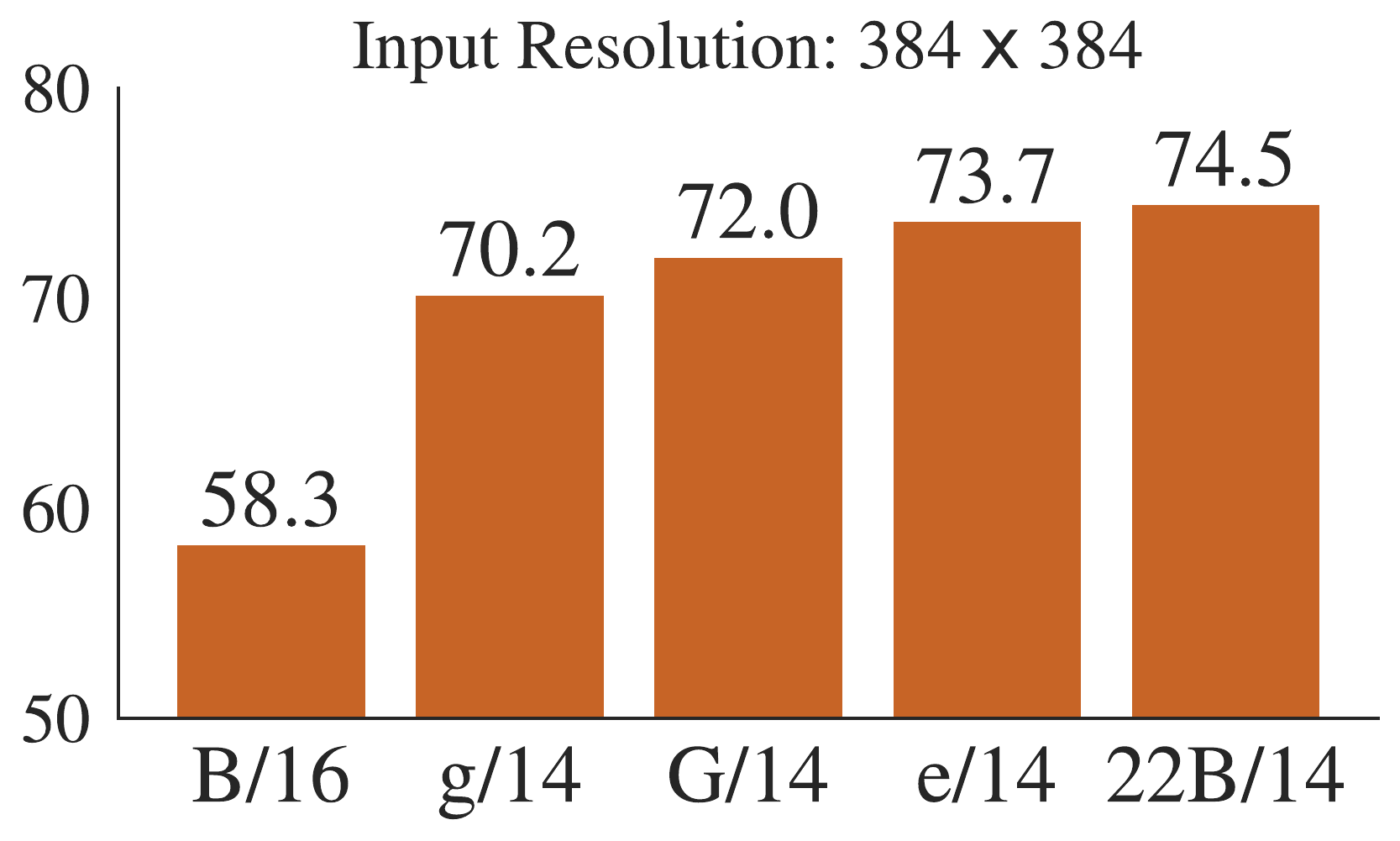}}
    \caption{Linear probing on iNaturalist 2017 with different input resolutions. \chonk leads to significant accuracy improvement especially when the input size is small.}
    \label{fig:inat}
\end{figure}

We further test linear separability on the fine-grained classification dataset, iNaturalist 2017~\citep{cui2018large}. It has 5,089 find-grained categories, belonging to 13 super-categories. Unlike ImageNet, the image numbers in different categories are not balanced. The long-tail distribution of concepts is more challenging for classification. We compare \chonk with the other ViT variants. 
Similar to the linear probing on ImageNet, we use SGD with 0.001 starting learning rate and no weight decay to optimize the models and train for 30 epochs with cosine learning rate schedule with 3 epochs of linear warm-up. We test both $224$px and $384$px input resolutions. \cref{fig:inat} shows the results. We observe that \chonk significantly improves over the other ViT variants, especially with the standard $224$px input resolution. This suggests the large number of parameters in \chonk are useful for extracting detailed information from the images. 

\subsubsection{Zero-shot via locked-image tuning}
\label{sect::lit}
\paragraph{Experimental setup.}
Following the Locked-image Tuning (LiT)~\citep{lit} protocol, we train a text tower contrastively to match the embeddings produced by the frozen \chonk model.
With this text tower, we can easily perform zero-shot classification and zero-shot retrieval tasks.
We train a text Transformer with the same size as ViT-g~\citep{zhai2022scaling} on the English subset of the WebLI dataset~\citep{chen2022pali} for 1M steps with a 32K batch size.
The images are resized to $288$px, and the text is tokenized to 16 tokens using a SentencePiece~\citep{kudo-richardson-2018-sentencepiece} tokenizer trained on the English C4 dataset.

\paragraph{Results.}
\cref{tab:lit} shows the zero-shot transfer results of \chonk against CLIP~\citep{clip}, ALIGN~\citep{align}, BASIC~\citep{basic}, CoCa~\citep{yu2022coca}, LiT~\citep{lit} with ViT-g~\citep{zhai2022scaling} and ViT-e~\citep{chen2022pali} models. 
The bottom part of \cref{tab:lit} compares three ViT models using the LiT recipe.
On all the ImageNet test sets, \chonk{} achieves either comparable or better results. 
Notably, zero-shot results on the ObjectNet test set is highly correlated with the ViT model size. 
The largest \chonk{} sets the new SOTA on the challenging ObjectNet test set.
\Cref{app:lit} shows zero-shot classification examples on OOD images.

\begin{table}
\centering
\caption{Zero-shot transfer results on ImageNet (variants).}
\resizebox{.5\textwidth}{!}{%
\setlength\tabcolsep{4pt} 
\begin{tabular}{@{}lcccccc@{}}
 \toprule
 Model & IN & IN-v2 & IN-R & IN-A & ObjNet & ReaL \\
 \midrule
 CLIP & 76.2 & 70.1 & 88.9 & 77.2 & 72.3 & - \\
 ALIGN & 76.4 & 70.1 & 92.2 & 75.8 & 72.2 & - \\
 BASIC & 85.7 & 80.6 & 95.7 & 85.6 & 78.9 & - \\
 CoCa & 86.3 & 80.7 & 96.5 & 90.2 & 82.7 & - \\
 \midrule
 LiT-g/14 & 85.2 & 79.8 & 94.9 & 81.8 & 82.5 & 88.6 \\
 LiT-e/14 & 85.4 & 80.6 & 96.1 & 88.0 & 84.9 & 88.4 \\
 LiT-22B & 85.9 & 80.9 & 96.0 & 90.1 & 87.6 & 88.6 \\
 \bottomrule
\end{tabular}
}
\label{tab:lit}
\end{table}

\subsubsection{Out-of-distribution}
\label{subsec:robustness_ood}


\paragraph{Experimental setup.} We construct a label-map from JFT to ImageNet, and label-maps from ImageNet to different out-of-distribution datasets, namely ObjectNet~\citep{barbu2019objectnet}, ImageNet-v2~\citep{recht2019imagenet_v2} ImageNet-R~\citep{hendrycks2020imagenet_r}, and ImageNet-A~\citep{hendrycks2021imagenet_a}. ImageNet-R and ImageNet-A use the same 200 label subspace of ImageNet (constructed in such a way that misclassifications would be considered egregious~\citep{hendrycks2021imagenet_a}), while ObjectNet has 313 categories, of which we only consider the 113 ones overlapping with the ImageNet label space. For ObjectNet and ImageNet-A we do an aspect-preserving crop of the central 75\% of the image, for the other datasets we first resize them to a square format and then take a 87.5\% central crop. Image input resolution is 224px for pre-trained checkpoints and 384px, 518px, 560px for models fine-tuned on ImageNet.

\begin{figure}[tbp]
    \centering
    \includegraphics[width=0.6\textwidth]{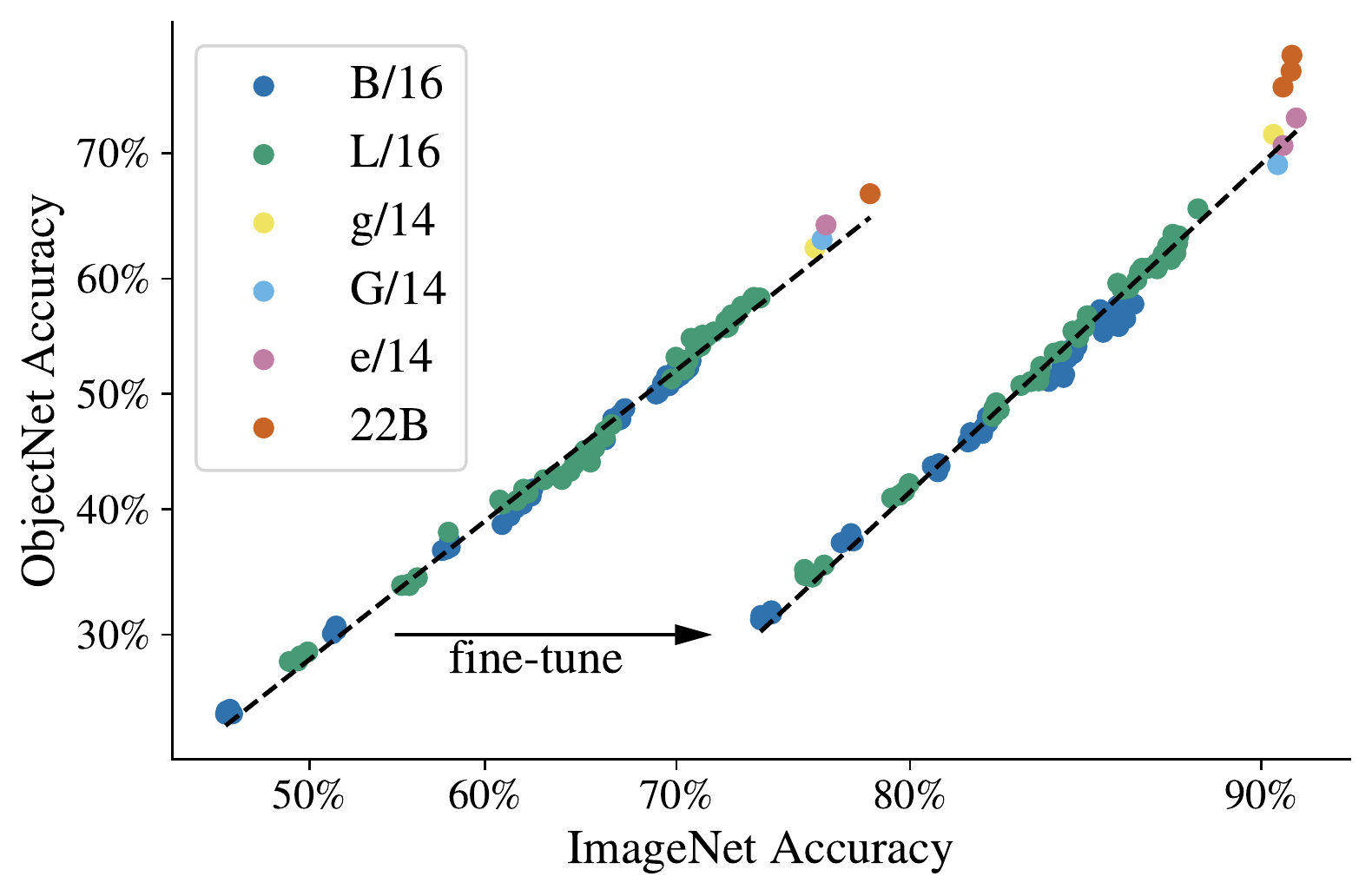}
    \caption{
        OOD classification performance. Axes are log-scaled as proposed in~\citep{taori2020measuring_robustness}. ViT-B and ViT-L are trained on subsets of varying size and varying number of steps on JFT~\citep{zhai2022scaling}. Fine-tuning boosts both ImageNet and ObjectNet performance, but the increase is more pronounced for in-domain data, which decreases effective robustness~\citep{andreassen2021effective_robustness}, visible as a rightwards shift on the plot. Same data as in \cref{tab:robustness_ood}.
    }
    \label{fig:robustness_objectnet}
\end{figure}

\paragraph{Results.} We can confirm results from~\citep{taori2020measuring_robustness,djolonga2021robustness_transferability,kolesnikov2020big} that scaling the model increases out-of-distribution performance in line with the improvements on ImageNet. This holds true for models that have only seen JFT images, and for models fine-tuned on ImageNet. In both cases,  \chonk continues the trend of better OOD performance with larger models (\cref{fig:robustness_objectnet}, \cref{tab:robustness_ood}).
While fine-tuning boosts accuracy on both ImageNet and out-of-distribution datasets, the effective robustness~\citep{andreassen2021effective_robustness} decreases (\cref{fig:robustness_objectnet}).
Even though ImageNet accuracy saturates, we see a significant increase on ObjectNet from ViT-e/14 to \chonk.

%% file: arxiv/text/4-2-eval_transfer_dense_prediction.tex
\subsection{Transfer to dense prediction}
Transfer learning for dense prediction is critical especially since obtaining pixel-level labels can be costly. In this section, we investigate the quality of captured geometric and spatial information by the \chonk model (trained using image-level classification objective) on semantic segmentation and monocular depth estimation tasks.

\subsubsection{Semantic segmentation}
\paragraph{Experimental setup.}
We evaluate \chonk as a backbone in semantic segmentation on three benchmarks: ADE20K~\citep{zhou2017scene}, Pascal Context~\citep{mottaghi2014role} and Pascal VOC~\citep{everingham2010pascal}.
We analyze the performance in two scenarios:
first, using a limited amount of data for transfer;
second (in~\cref{app:semantic_segmentation}), comparing end-to-end fine-tuning \textit{versus} a frozen backbone with either a linear decoder~\citep{strudel2021segmenter} or UperNet~\citep{xiao2018unified}.
The number of additional parameters ($\approx1$M for linear and $\approx783$M for UperNet) is negligible compared to the size of the backbone. We use a fixed resolution ($504$px) and report single scale evaluation.

\paragraph{Results.}
We compare \chonk to the ViT-L of DeiT-III~\citep{touvron2022deit} and ViT-G of \citet{zhai2022scaling}, when only a fraction of the ADE20k semantic segmentation data is available. 
We use the linear decoder and end-to-end fine-tuning.
From \cref{tab:semseg_fewshot}, we observe that our \chonk backbone transfers better when seeing only few segmentation masks.
For example, when fine-tuning with only 1200 images (i.e.\ $1/16$) of ADE20k training data, we reach a performance of 44.7 mIoU, an improvement of $+8.6$ mIoU over DeiT-III Large~\citep{touvron2022deit} and $+2.3$ mIoU over ViT-G~\citep{zhai2022scaling}.
When transferring with more data, the performance of ViT-G and \chonk converge.

\begin{table}[t]
    \caption{
      Fewshot semantic segmentation on ADE20k, when only a fraction of the training set is used.
We report mean IoU for semantic segmentation on the validation set.
Transfer is done with end-to-end fine-tuning and a linear decoder, following~\citet{strudel2021segmenter}.
We average over 3 runs.
}
\centering
\small
  \setlength{\tabcolsep}{4pt}
      \resizebox{.6\textwidth}{!}{%
    \begin{tabular}{@{} l c c c c c @{}}
      \toprule
	    Fraction of ADE20k train data      & $1/16$ & $1/8$ & $1/4$ & $1/2$ & $1$ \\
      \midrule
      ViT-L~\citep{touvron2022deit} & 36.1 & 41.3 & 45.6 & 48.4 & 51.9 \\
      ViT-G~\citep{zhai2022scaling} & 42.4 & 47.0 & 50.2 & 52.4 & \textbf{55.6} \\
      \chonk (Ours) & \textbf{44.7} & \textbf{47.2} & \textbf{50.6} & \textbf{52.5} & 54.9 \\
      \bottomrule
 \end{tabular}
 }
    \label{tab:semseg_fewshot}
\end{table}

\subsubsection{Monocular depth estimation}

\paragraph{Experimental setup.}
We largely mirror the set-up explored in \citet{ranftl2021vision} and train their Dense Prediction Transformer (DPT) on top of frozen \chonk backbone features obtained from the Waymo Open real-world driving dataset~\citep{sun2020scalability}.
Here we use only a single feature map (of the last layer) to better manage the high-dimensional ViT features.
We also explore a much simpler ``linear'' decoder as a lightweight readout.
In both cases we predict $\log(1 + \text{depth})$ obtained from sparse LiDAR as the target and use Mean Squared Error (MSE) as the decoder training loss. 
We quantify performance using standard depth estimation metrics from the literature~\citep{hermann2020self, eigen2014depth} and also report MSE. We use a resolution of $224\times224$. 
Remaining details are deferred to \Cref{app:de}.

\begin{figure}[tbp]
    \centering
    \subfigure[Semantic segmentation]{\includegraphics[width=0.315\linewidth]{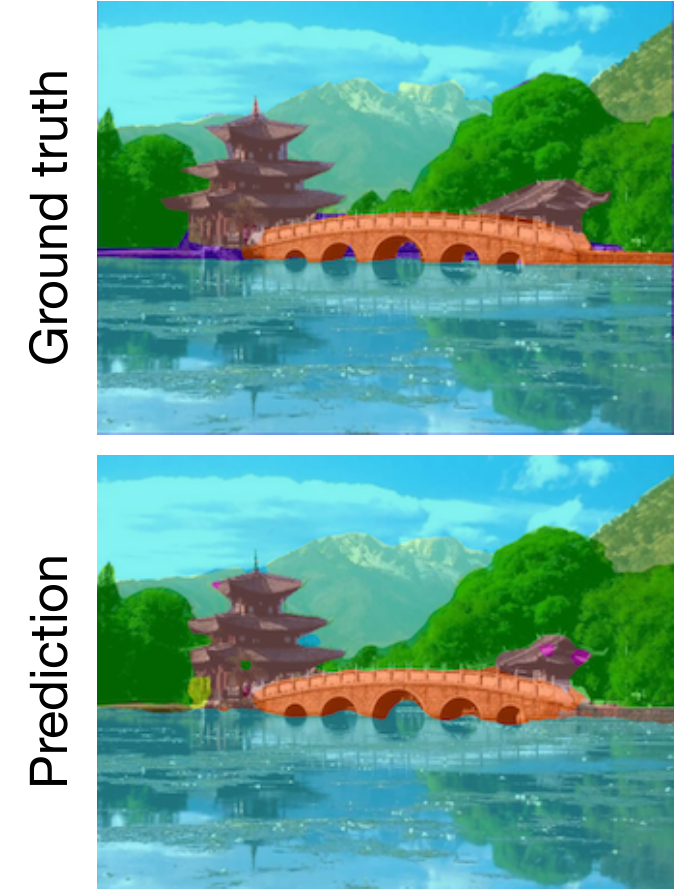}}\quad\quad \quad\quad
    \subfigure[Depth estimation]{\includegraphics[width=0.2955\linewidth]{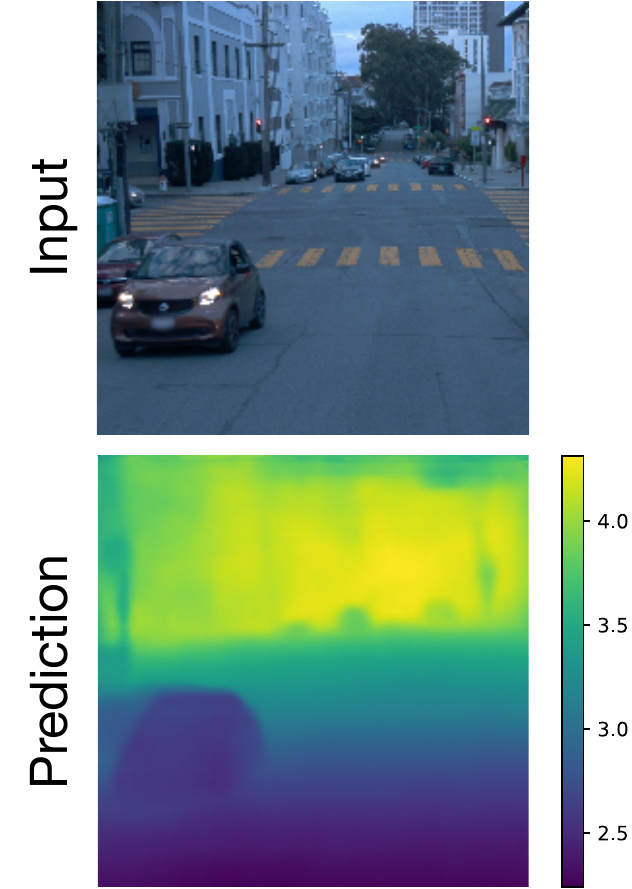}}
    \caption{Dense prediction from frozen \chonk features.}
    \label{fig:dense_prediction}
\end{figure}

\begin{table}[t]
\caption{
{Monocular depth estimation} from frozen ViT features using different decoders on the Waymo Open dataset.}
\centering
\small
\setlength{\tabcolsep}{4pt}
\resizebox{.55\textwidth}{!}{%
\begin{tabular}{@{} ll ccccc  @{}}
  \toprule
   & & & & \multicolumn{3}{c}{$\delta$ $\uparrow$} \\
   \cmidrule(l{2pt}r{2pt}){5-7}
    & Model      & MSE $\downarrow$ & AbsRel $\downarrow$ & $< 1.1$& $< 1.25$& $< 1.25^2$\\
  \midrule
  \multirow[c]{3}{*}{\rotatebox{90}{DPT}} & ViT-L & 0.027 & 0.121 & 0.594 & 0.871 & 0.972 \\ 
  & ViT-e & 0.024 & 0.112 & 0.631 & 0.888 & 0.975 \\ 
  & \chonk & \textbf{0.021} & \textbf{0.095} & \textbf{0.702} & \textbf{0.909} & \textbf{0.979} \\ 
  \midrule
  \multirow[c]{3}{*}{\rotatebox{90}{Linear}} & ViT-L & 0.060 & 0.222 & 0.304 & 0.652 & 0.926 \\ 
  & ViT-e & 0.053 & 0.204 & 0.332 & 0.687 & 0.938 \\ 
  & \chonk & \textbf{0.039} & \textbf{0.166} & \textbf{0.412} & \textbf{0.779} & \textbf{0.960} \\ 
  \bottomrule
\end{tabular}
}
\label{tab:depth_estimation}
\end{table}

\paragraph{Results.}
\Cref{tab:depth_estimation} summarizes our main findings.
From the top rows (DPT decoder), we observe that using \chonk features yields the best performance (across all metrics) compared to different backbones. By comparing the \chonk backbone to ViT-e (a smaller model but trained on the same data as \chonk) we find that scaling the architecture improves performance. Further, comparing the ViT-e backbone to ViT-L (a similar architecture to ViT-e but trained on less data) we find that these improvements also come from scaling the pre-training data. These findings demonstrate that both the greater model size and the greater dataset size contribute substantially to the improved performance.
Using the linear decoder, it can be observed again that using \chonk features yields the best performance.
The gap between DPT and linear decoding suggests that while enough geometric information is retained in the ViT features, only some of it is available for a trivial readout.
We report qualitative results in \Cref{fig:dense_prediction} and \Cref{fig:dpt_depth_predictions_full,fig:dpt_depth_predictions_full_error} in \cref{app:de}.


%% file: arxiv/text/4-3-eval_transfer_video.tex
\subsection{Transfer to video classification}
\paragraph{Experimental setup.}
We evaluate the quality of the representations learned by \chonk by adapting the model pretrained on images for video classification.
We follow the ``factorised encoder'' architecture of~\citet{arnab2021vivit}: Our video model consists of an initial ``spatial transformer'', which encodes each frame of the video independently of each other.
Thereafter, the representation from each frame is pooled into a single token, which is then fed to a subsequent ``temporal transformer'' that models the temporal relations between the representations of each frame.

Here, we initialize the ``spatial transformer'' with the pretrained weights from ViT-22B and freeze them, as this represents a computationally efficient method of adapting large-scale models for video, and also because it allows us to effectively evaluate the representations learned by pretraining \chonk. Exhaustive experimental details are included in \cref{sec:app_video_classification}. The temporal transformer is lightweight both in terms of parameters (only 63.7M parameters compared to the 22B frozen parameters in the spatial transformer), and FLOPs as it operates on a \emph{single} token per frame.

\paragraph{Results.}
\begin{table}[t]
    \caption{
        Video classification results.
        We evaluate the \chonk representations by freezing the backbone, and training a small transformer to aggregate frozen, per-frame representations.
        \chonk outperforms the largest previous vision backbone, ViT-e~\citep{chen2022pali} which contains 4 billion parameters and is also pretrained on JFT.
    }
    \centering{
    \resizebox{.55\textwidth}{!}{%
    \begin{tabular}{@{}lcc@{}}
        \toprule
                             & Kinetics 400 & Moments in Time \\
        \midrule
        \multicolumn{3}{l}{\textit{Frozen backbone}}                   \\
        CoCA$^*$                 &   88.0           &    47.4             \\
        ViT-e               &   86.5           &     43.6           \\  
        \chonk              &   88.0           &    44.9           \\  
        \midrule
        Fully finetuned SOTA &   91.1           &    49.0              \\ 
        \bottomrule
    \end{tabular}
    }}
    \begin{adjustwidth}{110pt}{110pt}
    \begin{spacing}{0.125}{\fontsize{8pt}{5pt}\selectfont $^*$Note that CoCA uses pre-pool spatial features and higher spatial resolution for both datasets. More details in \cref{sec:app_video_classification}.}\end{spacing}
    \end{adjustwidth}
    \label{tab:results_video}
\end{table}
\cref{tab:results_video} presents our results on video classification on the Kinetics 400~\citep{kay2017kinetics} and Moments in Time~\citep{monfort2019moments} datasets, showing that we can achieve competitive results with a frozen backbone.
We first compare to ViT-e~\citep{chen2022pali}, which has the largest previous vision backbone model consisting of 4 billion parameters, and was also trained on the JFT dataset.
We observe that our larger \chonk model improves by 1.5 points on Kinetics 400, and 1.3 points on Moments in Time. 
Our results with a frozen backbone are also competitive with CoCA~\citep{yu2022coca}, which performs a combination of contrastive and generative caption pretraining in comparison to our supervised pretraining, and uses many tokens per frame (vs.\ a single one produced by the pretrained frozen pooling) as well as a higher testing resolution.

Finally, we note that there is headroom for further improvement by full end-to-end fine-tuning. This is evidenced by the current state-of-the-art on Kinetics 400~\citep{wang2022internvideo} and Moments in Time~\citep{yu2022coca} which leverage a combination of large-scale video pretraining and full end-to-end fine-tuning on the target dataset. 

%% file: arxiv/text/4-4-eval_beyond_accuracy.tex
\subsection{Beyond accuracy on downstream tasks}
When studying the impact of scaling, there are important aspects to consider beyond downstream task performance. In this section, we probe \chonk's fairness, alignment with human perception, robustness, reliability, and calibration. We find that favorable characteristics emerge when increasing model size. Additional analysis on perceptual similarity and feature attribution can be found in \cref{app:perceptual_similarity} and \cref{app:feature_attribution}.

\subsubsection{Fairness}\label{sect::fairness}
Machine learning models are susceptible to unintended bias. For example, they can amplify spurious correlations in the training data~\citep{hendricks2018women,caliskan2017semantics,zhao2017men,wang2020towards} and result in error disparities~\citep{zhao2017men,buolamwini2018gender,deuschel2020uncovering}. 
Here, we identify how scaling the model size can help mitigate such issues, by evaluating the bias of \chonk  and ViT-\{L, g, G, e\} ~\citep{zhai2022scaling,chen2022pali} using demographic parity (DP) as a measure of fairness~\citep{dwork2012fairness,zafar2017fairness}.

\begin{figure}[hbt!]
    \centering
    \includegraphics[width=0.7\columnwidth]{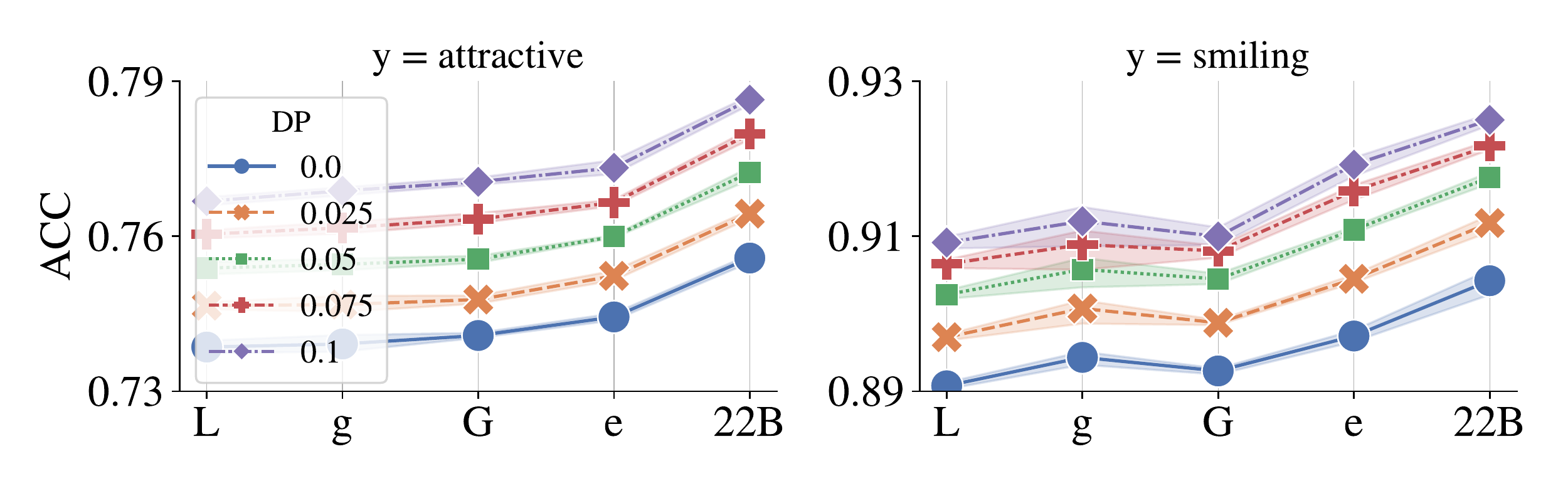}
    \includegraphics[width=0.7\columnwidth]{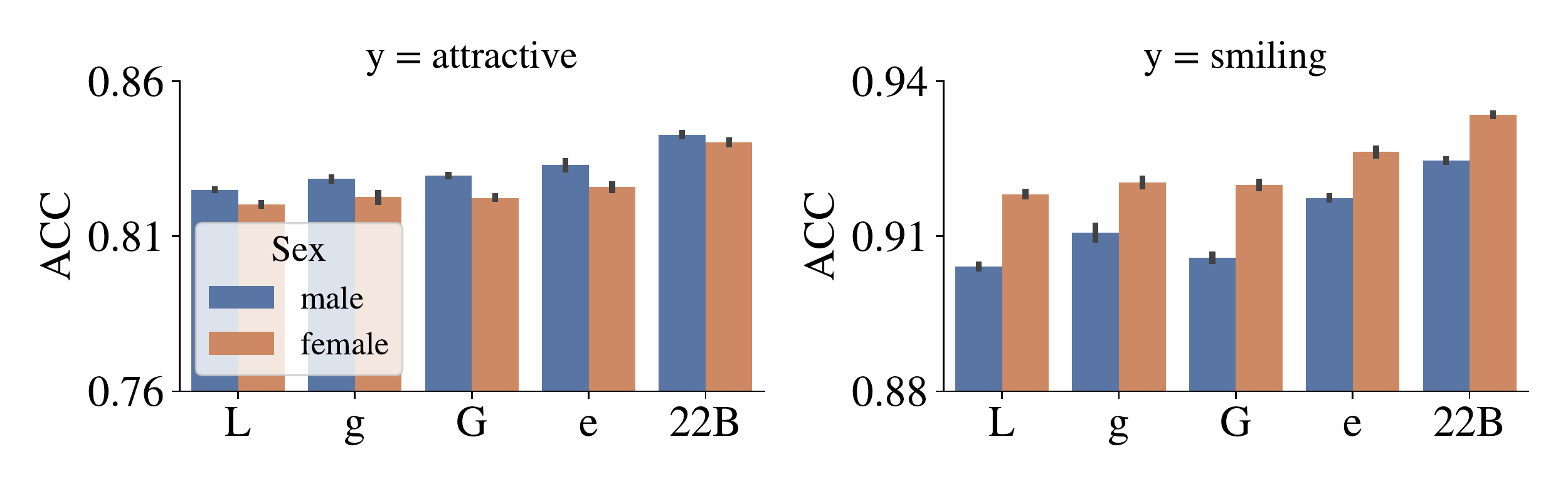}
    \includegraphics[width=0.7\columnwidth]{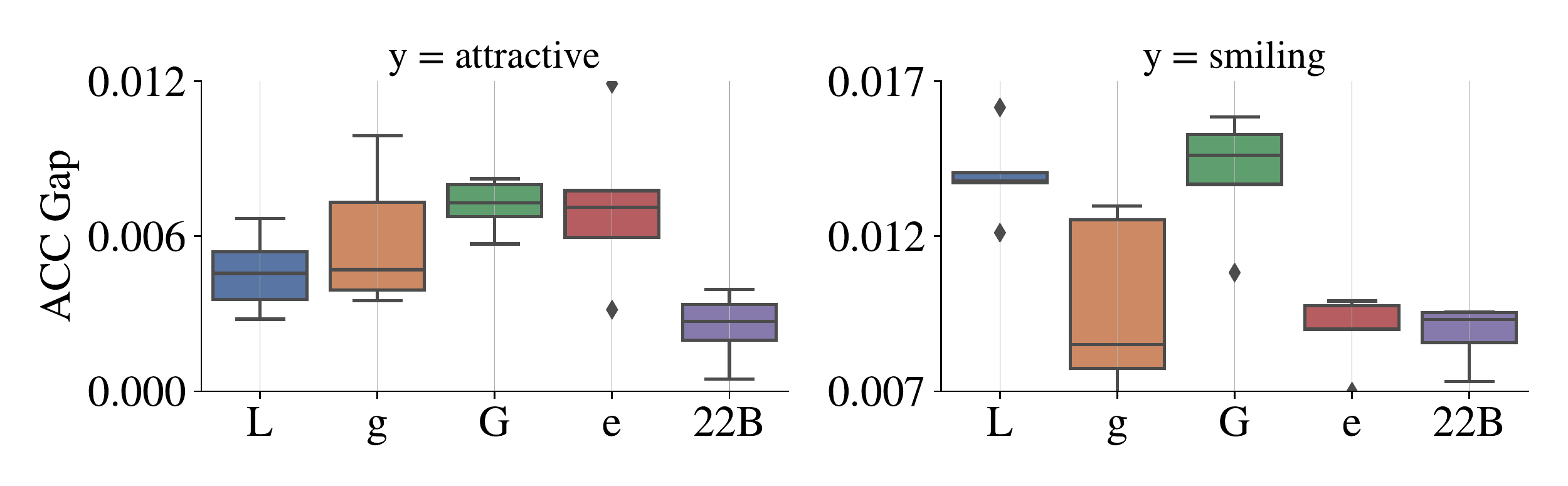}
    \vspace{-10pt}
    \caption{{\sc top:} Accuracy (ACC) for ViT variants \emph{after} debiasing for each DP level.  {\sc middle}: Accuracy for each subgroup in CelebA \emph{prior to} debiasing. {\sc bottom}:  $y$-axis is absolute difference in performance across the two subgroups: females and males. \chonk provides a more equitable performance, compared to smaller ViT architectures.}
    \label{fig:fairness:main}
    \vspace{-10pt}
\end{figure}


\paragraph{Experimental Setup.} We use CelebA~\citep{liu2015faceattributes} with binary gender as a sensitive attribute while the target is  ``attractive'' or ``smiling''. We emphasize that such experiments are carried out only to verify technical claims and shall by no means be interpreted as an endorsement of such vision-related tasks. We choose the latter attributes because they exhibit gender related bias as shown in \cref{fig:fairness:bias_original}.

We train a logistic regression classifier on top of the \chonk pretrained features for a total of $50$ epochs and batch size $256$, with a learning rate schedule of $0.01$ (first 25 epochs) and 0.001 (last 25 epochs).  After that, we debias using the randomized threshold optimizer (RTO) algorithm of~\citet{alabdulmohsin2021}, which was shown to be near-optimal and competitive with in-processing methods. 

\paragraph{Results.}
We observe that scale by itself does not impact DP, c.f.\ \cref{fig:fairness:bias_original}. This is perhaps not surprising, as the model is trained to reconstruct a chosen target so the level of DP in accurate models is similar to that of the data itself.

However, scaling to \chonk offers benefits for fairness in other aspects. First, scale offers a more favorable tradeoff ---  performance improves with scale subject to any prescribed level of bias constraint.
This is consistent with earlier observations reported in the literature~\citep{alabdulmohsin2021}.  Second, all subgroups tend to benefit from the improvement in scale. Third, \chonk reduces disparities in performance across subgroups. 
\cref{fig:fairness:main} summarizes results for classification accuracy and \cref{app:fairness} for expected calibration error (ECE)~\citep{naeini2015obtaining,guo2017calibration} and OC-AUC~\citep{kivlichan2021measuring}.

\subsubsection{Human Alignment}
\label{subsec:error consistency}

\begin{figure}
    \centering
    \vspace{-5pt}
    \includegraphics[width=\textwidth]{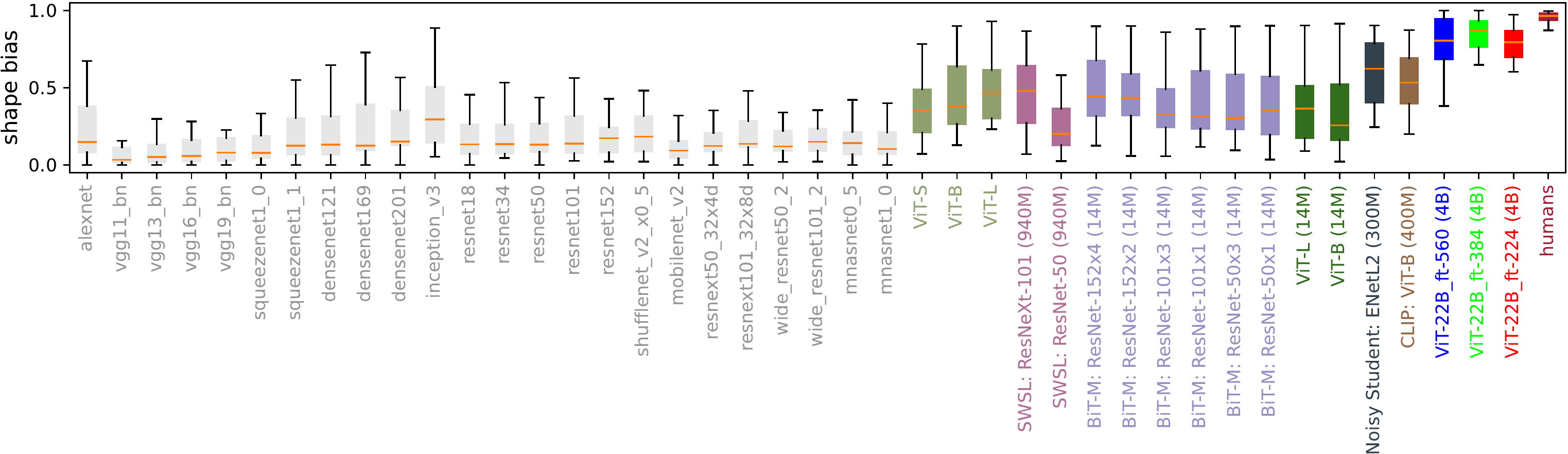}
    \vspace{-0.8cm}
    \caption{Shape bias: many vision models have a low shape / high texture bias, whereas \chonk fine-tuned on ImageNet (\textcolor{red}{red}, \textcolor{green}{green}, \textcolor{blue}{blue} trained on 4B images as indicated by brackets after model names, unless trained on ImageNet only) have the highest shape bias recorded in a ML model to date, bringing them closer towards a human-like shape bias.}
    \label{fig:shape_bias}
\vspace{-10pt}
\end{figure}

How well do \chonk classification decisions align with human classification decisions? Using the {\tt model-vs-human} toolbox~\citep{geirhos2021partial}, we evaluate three \chonk models fine-tuned on ImageNet with different resolutions (224, 384, 560). Accross all toolbox metrics, \chonk is SOTA: \chonk-224 for highest OOD robustness (\cref{fig:model_vs_human_benchmark_a}), \chonk-384 for the closest alignment with human classification accuracies (\cref{fig:model_vs_human_benchmark_b}), and \chonk-560 for the largest error consistency (i.e.\ most human-like error patterns, \cref{fig:model_vs_human_benchmark_d}). The \chonk models have the highest ever recorded shape bias in vision models: while most models have a strong texture bias (approx.\ 20--30\% shape bias / 70--80\% texture bias)~\citep{geirhos2019imagenet}; humans are at 96\% shape / 4\% texture bias and \chonk-384 achieves a previously unseen 87\% shape bias / 13\% texture bias (\cref{fig:shape_bias}). Overall, \chonk measurably improves alignment to human visual object recognition.

\subsubsection{Plex - pretrained large model extensions}
\citet{tran2022plex} comprehensively evaluate the reliability of models through the lens of uncertainty, robustnes (see \cref{subsec:robustness_ood}) and adaptation (see \cref{sect::lit}).
We focus here on the first aspect of that benchmark. 
To this end, we consider (1) the OOD robustness under
covariate shift with ImageNet-C~\citep{hendrycks2019benchmarking}, which we evaluate not only with the accuracy but also uncertainty metrics measuring the calibration (NLL, ECE) and the selective prediction~\citep{el2010foundations} (OC-AUC, see \cref{sect::fairness}), and (2) open-set recognition---also known as OOD detection~\citep{fort2021exploring}, which we evaluate via the AUROC and AUPRC, with Places365 as the OOD dataset~\citep{hendrycks2019scaling}; for more details, see \cref{app:plex}.


In \cref{tab:plex}, we report the performance of ViT-L and ViT-22B (both with resolution 384) fine-tuned on ImageNet. 
To put in perspective the strong gains of ViT-22B, we also show Plex-L, a ViT-L equipped with the two components advocated by \citet{tran2022plex}, viz, efficient-ensemble~\citep{wen2019batchensemble} and heteroscedastic layers~\citep{collier2021correlated}.
We discuss the challenges and the results of the usage of those components at the 22B scale (Plex-22B) in \cref{app:plex}.


\begin{table}[t]
    \caption{ViT-22B evaluated on some representative metrics from the Plex reliability benchmark~\citep{tran2022plex}$^*$.}
\centering
  \setlength{\tabcolsep}{4pt}
  \resizebox{.6\textwidth}{!}{%
    \begin{tabular}{@{} l @{} c ccc c ccc @{}}
      \toprule
       && \multicolumn{4}{c}{IN-C (mean over shifts)} && \multicolumn{2}{c}{IN vs.~Places365} \\
\cmidrule{3-6}\cmidrule{8-9}
	      Metrics && ACC $\uparrow$ & NLL $\downarrow$ & ECE $\downarrow$ & OC-AUC $\uparrow$ && AUROC $\uparrow$ & AUPRC $\uparrow$ \\
      \midrule
      ViT-L/32$^*$ &&  70.1 & 1.28 &  0.05 & 0.91 && 0.83 &  0.96 \\
      Plex-L/32$^*$ &&  71.3 & 1.21 & 0.02 &  0.91 && 0.83 &  0.97 \\
      ViT-22B && \textbf{83.7} & \textbf{0.63} &  \textbf{0.01} & \textbf{0.97} && \textbf{0.88} &  \textbf{0.98}  \\
      \bottomrule
 \end{tabular}}
    \label{tab:plex}
\end{table}

\subsubsection{Calibration}
\label{subsec:calibration}
Along with the robustness of \cref{subsec:robustness_ood}, it is also natural to wonder how the calibration property of ViT evolves as the scale increases. To this end, we focus on the study of \citet{minderer2021revisiting} that we extend with \chonk.

\begin{figure}[tbp]
    \centering
    \subfigure[Unscaled]{\includegraphics[width=0.35\textwidth]{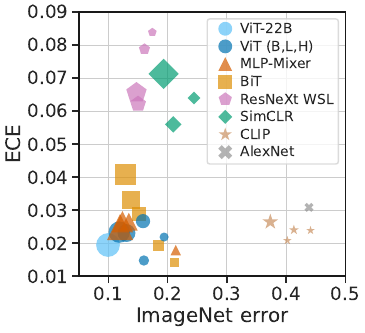}}\quad\quad 
    \subfigure[Temperature-scaled]{\includegraphics[width=0.35\textwidth]{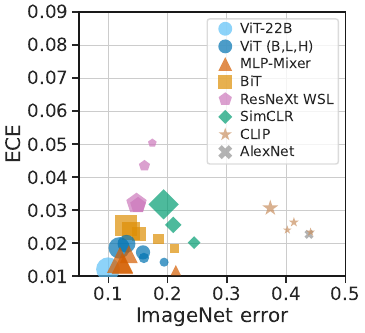}}
    \caption{\chonk (light-blue circle) improves the Pareto frontier of the accuracy vs.~the calibration (ECE). Left/right panels are without/with temperature scaling, respectively.}
    \label{fig:calibration}
\end{figure}

In \cref{fig:calibration}, we consider \chonk fine-tuned on ImageNet (resolution 384) and report the error (i.e., one minus accuracy) versus the calibration, as measured by the expected calibration error (ECE)~\citep{naeini2015obtaining,guo2017calibration}.
We see how \chonk remarkably improves the tradeoff between accuracy and calibration. The conclusion holds both without (left) and with (right) a temperature-scaling of the logits that was observed to better capture the calibration trends across model families~\citep{minderer2021revisiting}. More details can be found in \cref{app:calibration}.

\subsubsection{Distillation}
We perform model distillation~\citep{hinton2015distilling} to compress the \chonk into smaller, more widely usable ViTs.
We distill \chonk into ViT-B/16 and ViT-L/16 by following the procedure of~\citet{beyer2022knowledge}.
Using ImageNet-finetuned (at 384px) \chonk, we annotated 500 random augmentations and mixup transforms of each ImageNet image with \chonk logits.
Then, we minimize the KL divergence between the student and the teacher predictive distributions.
We train for 1000 epochs after initializing the student architecture from checkpoints pre-trained on JFT.
The results are shown in \cref{tab:distillation_imagenet}, and we see that we achieve new SOTA on both the ViT-B and ViT-L sizes.

\begin{table}[tbp]
    \caption{
      Distillation results, finetuned at 384 resolution.
}
\centering
  \setlength{\tabcolsep}{4pt}
  \resizebox{0.6\textwidth}{!}{%
    \begin{tabular}{@{} l l c @{}}
      \toprule
	    Model      & &
	    ImageNet1k 
	    \\
      \midrule
      \multirow[c]{4}{*}{\rotatebox{90}{ViT-B/16}}
      &~\citep{dosovitskiy2020image} (JFT ckpt.) & 84.2 \\
      &~\citep{zhai2022scaling} (JFT ckpt.) & 86.6 \\
      &~\citep{touvron2022deit} (INet21k ckpt.) & 86.7 \\
      & Distilled from \chonk (JFT ckpt.) & \textbf{88.6} \\
      \midrule
          \multirow[c]{4}{*}{\rotatebox{90}{ViT-L/16}}
      &~\citep{dosovitskiy2020image} (JFT ckpt.) & 87.1 \\
      &~\citep{zhai2022scaling} (JFT ckpt.) & 88.5 \\
      &~\citep{touvron2022deit} (INet21k ckpt.) & 87.7 \\
      & Distilled from \chonk (JFT ckpt.) & \textbf{89.6} \\ 
      \bottomrule
 \end{tabular}
 }
    \label{tab:distillation_imagenet}
\end{table}

%% file: arxiv/text/5-conclusion.tex
\section{Conclusion}
\label{sec:conclusion}
We presented \chonk, the currently largest  vision transformer model at 22 billion parameters.
We show that with small, but critical changes to the original architecture, we can achieve both excellent hardware utilization and training stability, yielding a model that advances the SOTA on several benchmarks.
In particular, great performance can be achieved using the frozen model to produce embeddings, and then training thin layers on top.
Our evaluations further show that \chonk is more aligned with humans when it comes to shape and texture bias, and offers benefits in fairness and robustness, when compared to existing models.




%% file: arxiv/text/6-appendix.tex
\appendix
\onecolumn
\input{arxiv/text/appendix/lit}
\input{arxiv/text/appendix/model}
\input{arxiv/text/appendix/model_card}

\input{arxiv/text/appendix/image_classification}
\input{arxiv/text/appendix/dense_prediction}
\input{arxiv/text/appendix/video_classification}
\input{arxiv/text/appendix/fairness}
\input{arxiv/text/appendix/calibration}
\input{arxiv/text/appendix/plex}
\input{arxiv/text/appendix/analysis}

%% file: arxiv/text/appendix/lit.tex
\section{Zero-shot Classification Examples}
\label{app:lit}

\begin{figure}[hb]
\newlength{\figwidth}\setlength{\figwidth}{0.45\textwidth}
\centering
\resizebox{0.92\columnwidth}{!}{%
\begin{tabularx}{\textwidth}{ cc }
\includegraphics[width=\figwidth,valign=t]{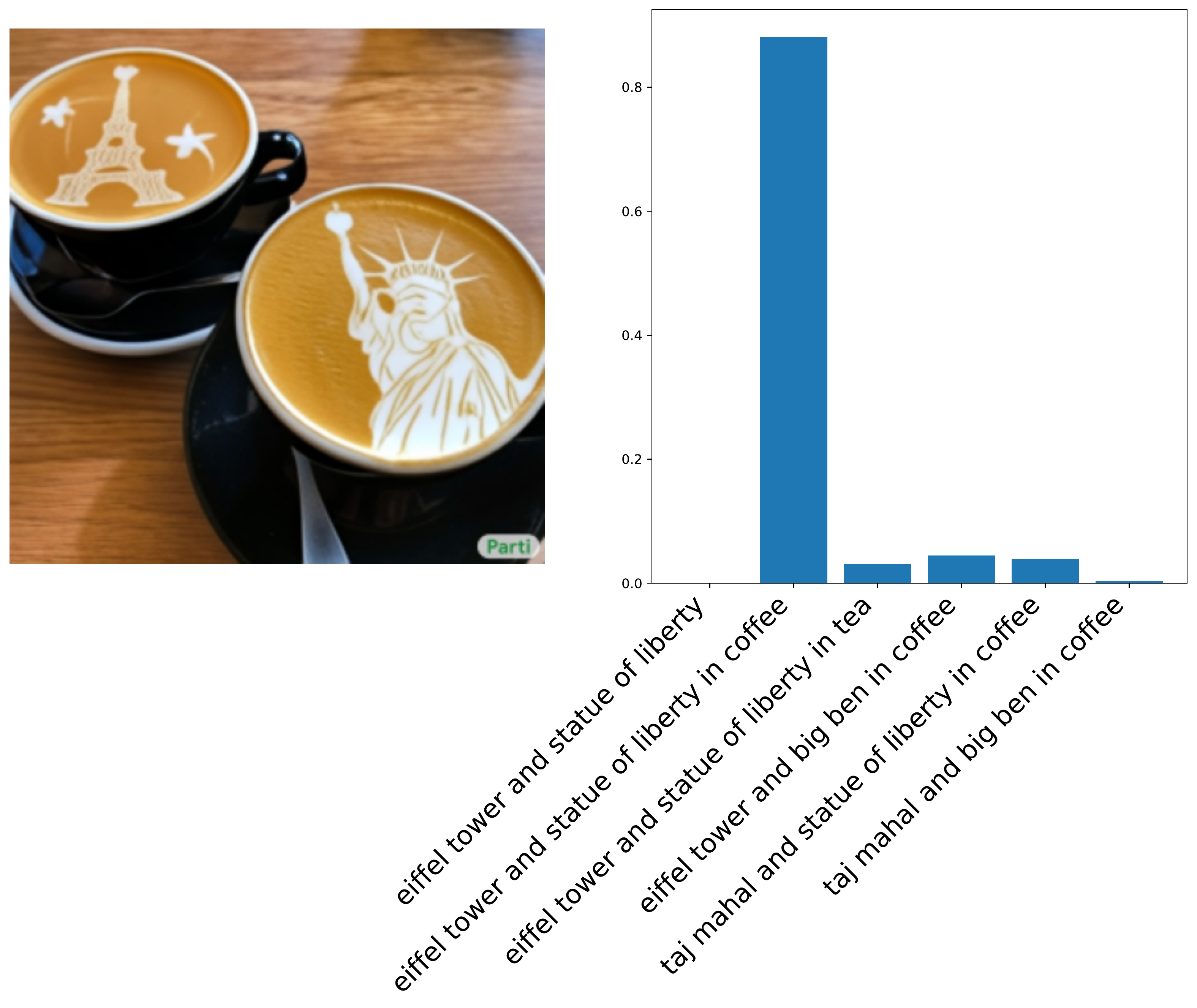} &
\includegraphics[width=\figwidth,valign=t]{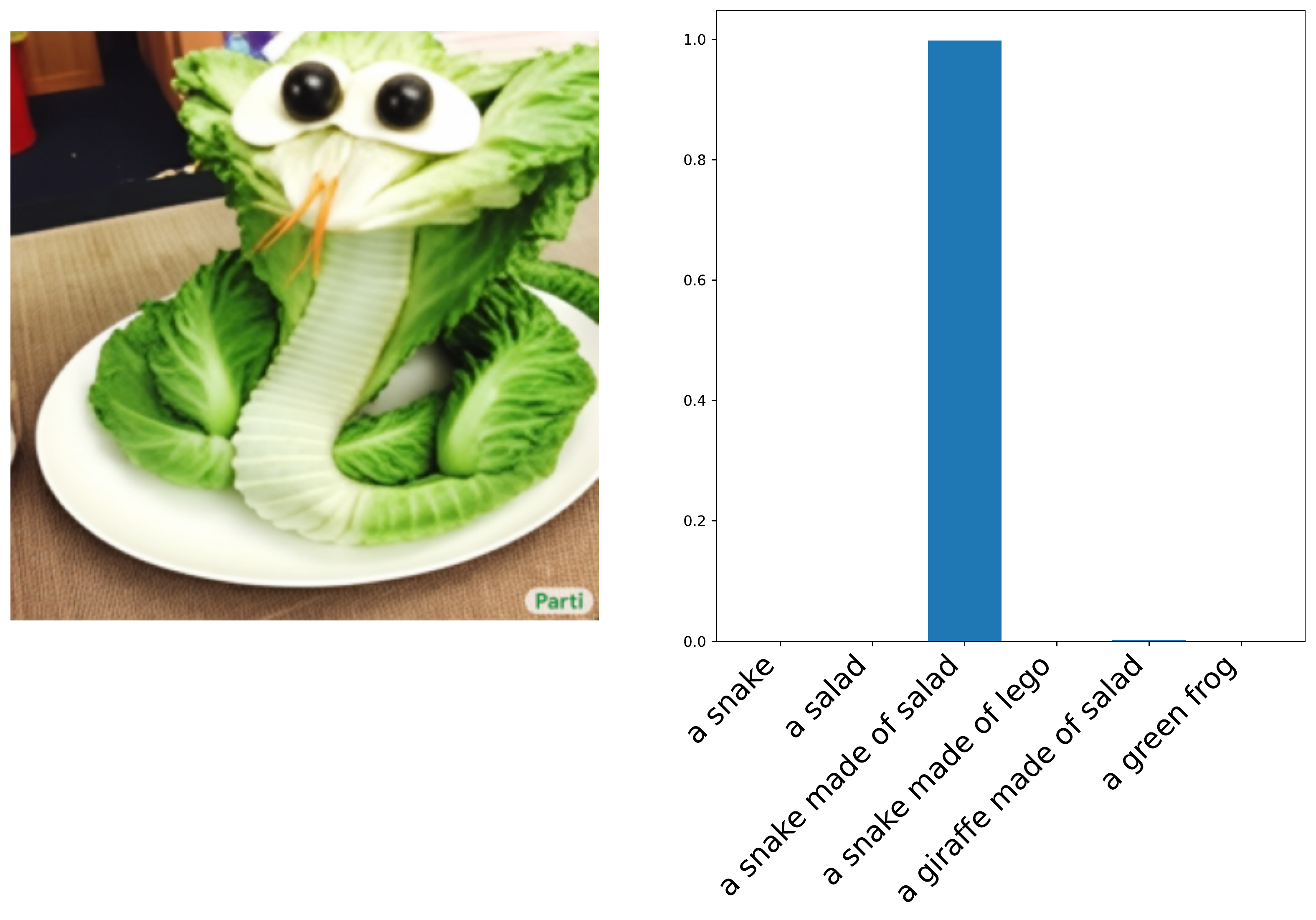}\\
\includegraphics[width=\figwidth,valign=t]{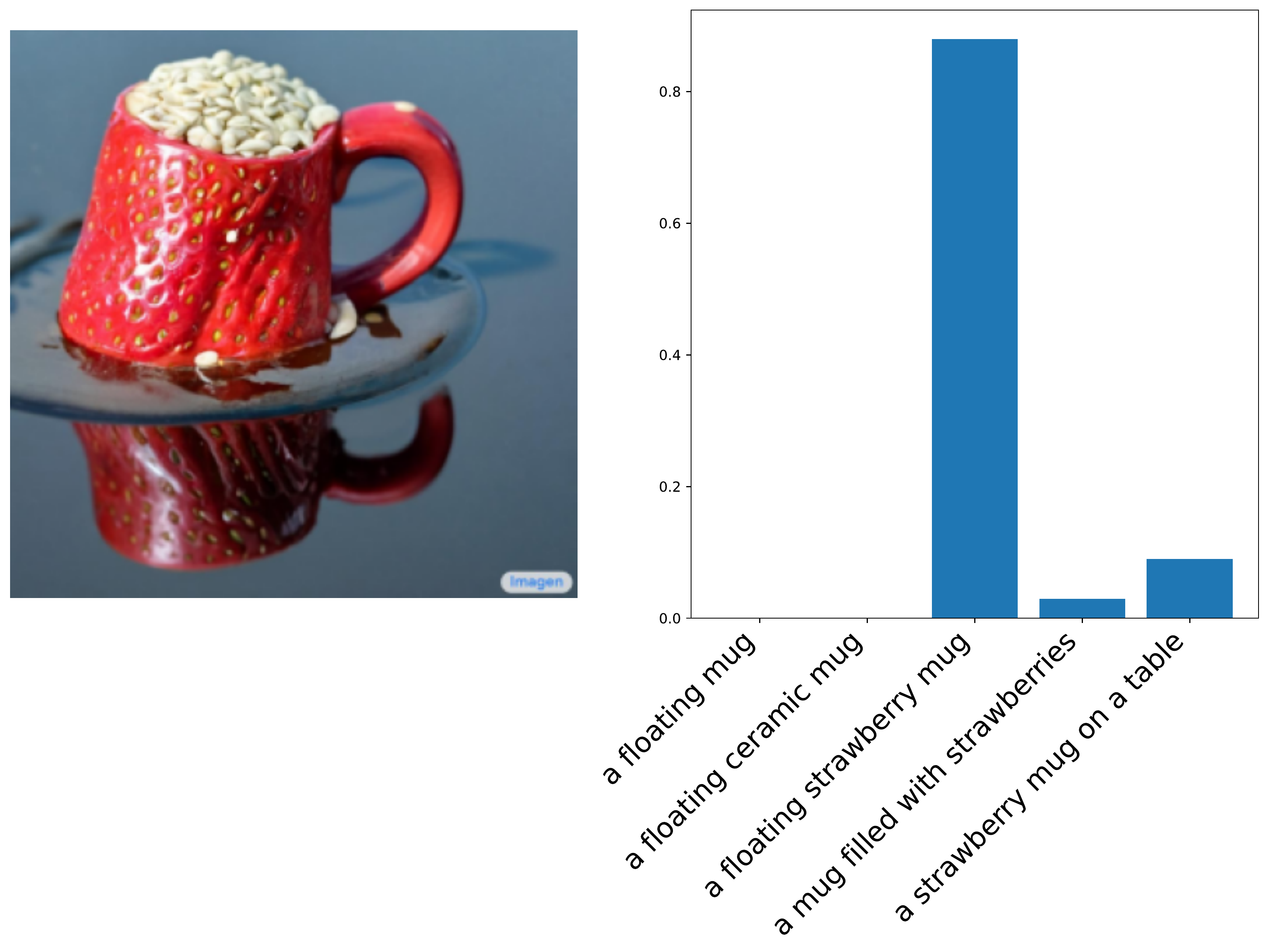} &
\includegraphics[width=\figwidth,valign=t]{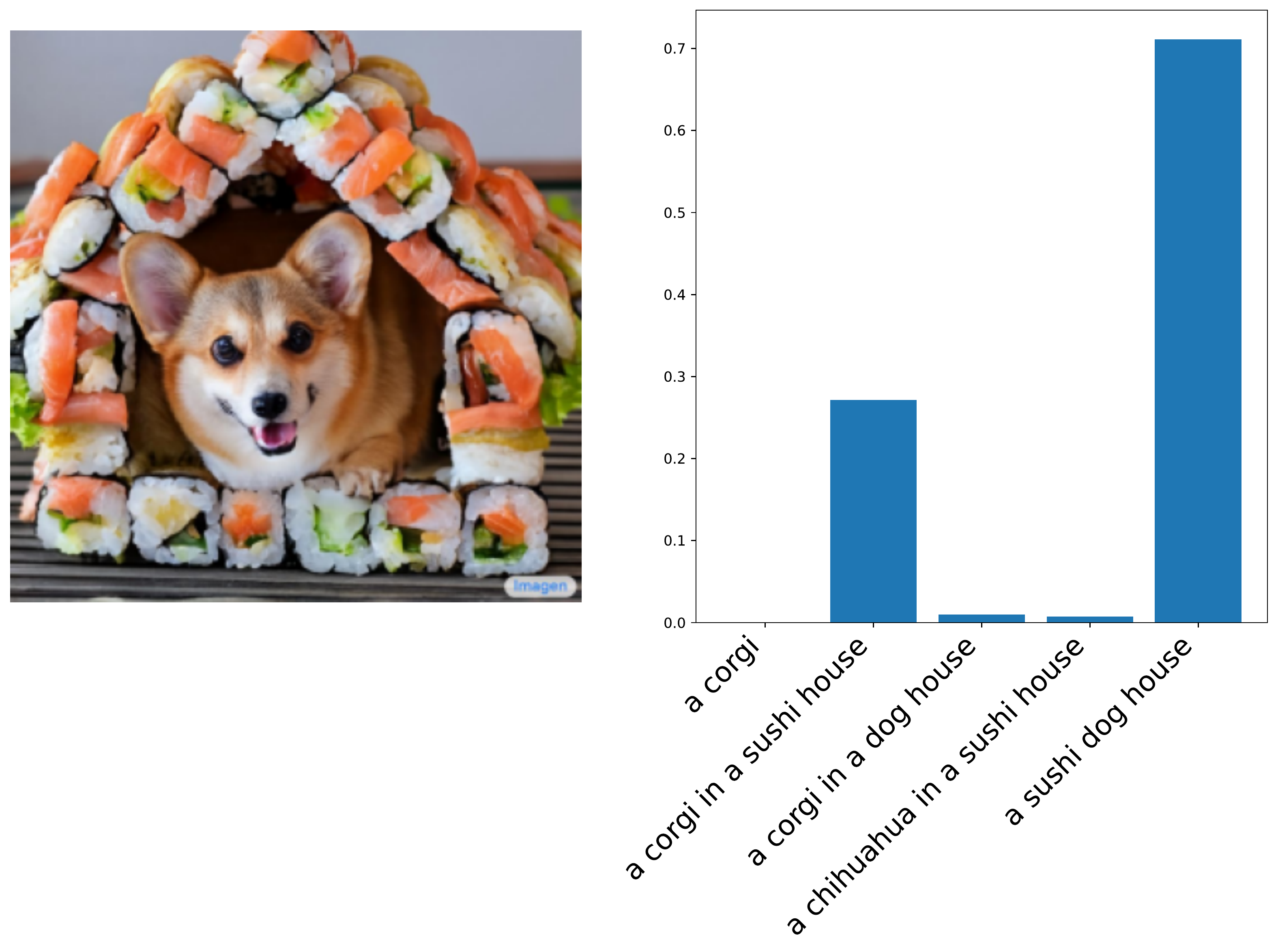}\\
\includegraphics[width=\figwidth,valign=t]{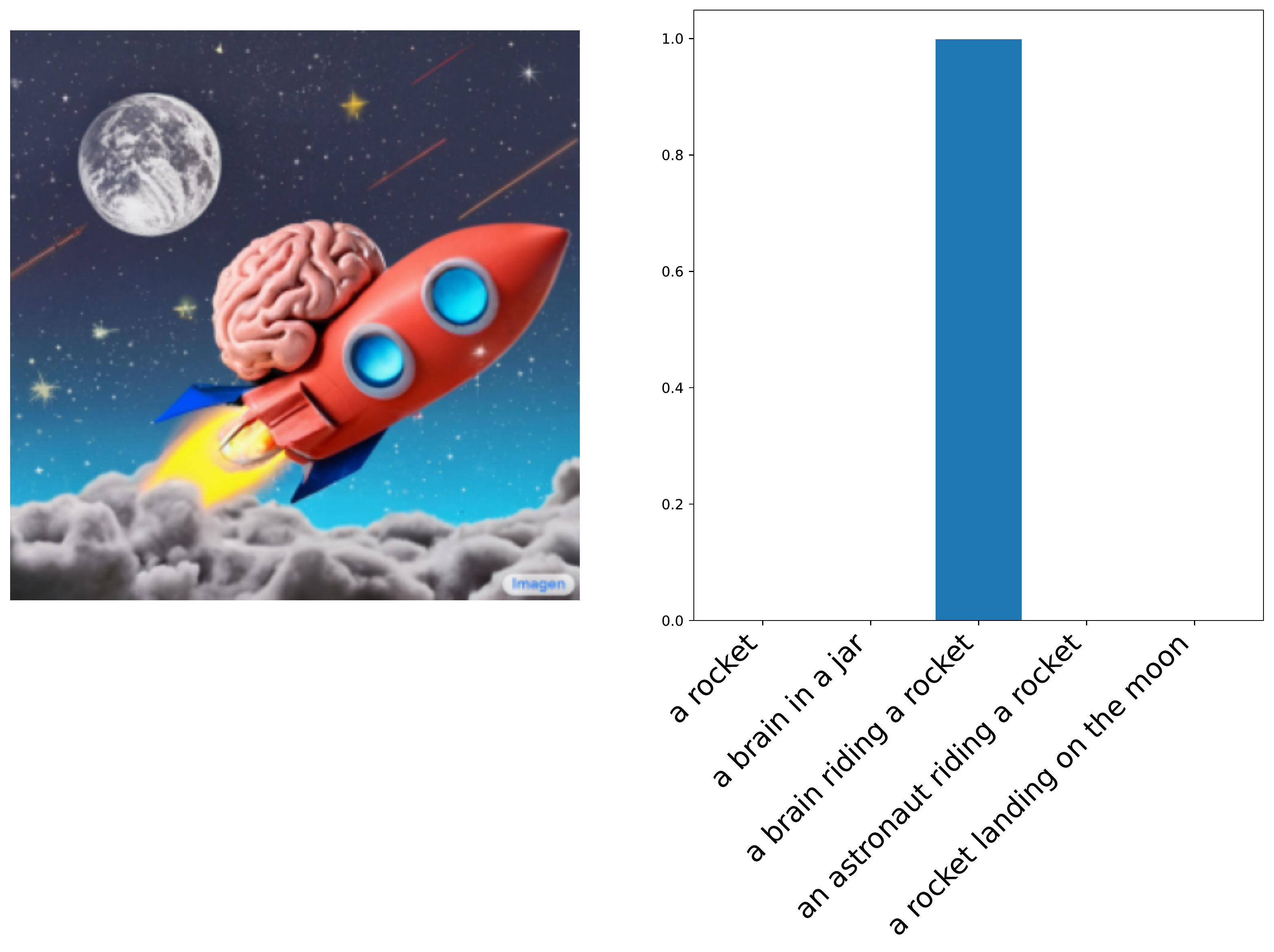} &
\includegraphics[width=\figwidth,valign=t]{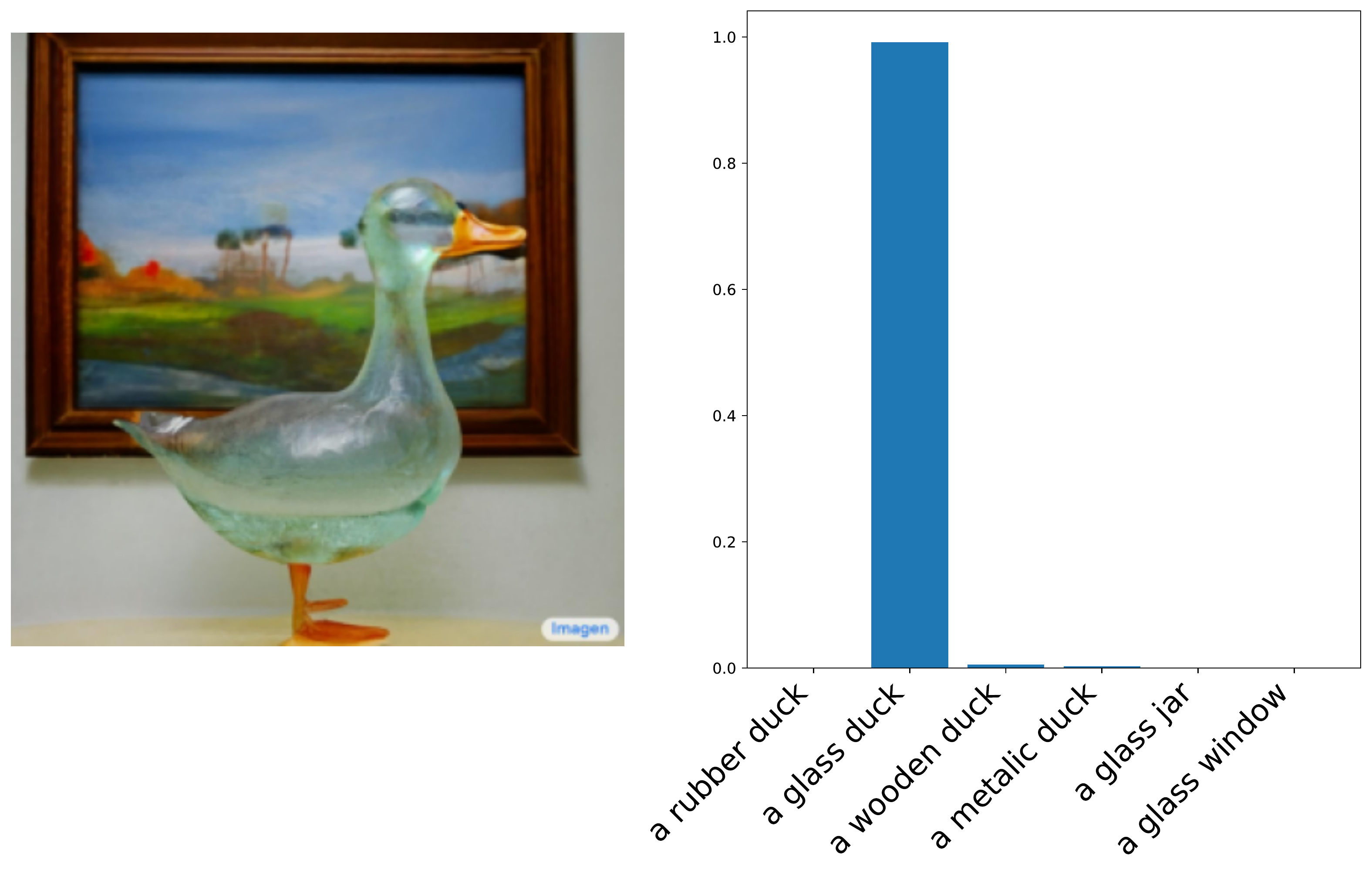}\\
\end{tabularx}%
} 
\caption{Examples of zero-shot classification results on images generated by the Parti~\citep{parti} and Imagen~\citep{imagen} models.
These examples contain unusual objects/scenes that do not occur in the training distribution.
\label{tab:lit_examples}
}
\end{figure}

\Cref{tab:lit_examples} contains example zero-shot classifications of generated images.
These images were provided by the Parti~\citep{parti} and Imagen~\citep{imagen} models.
The training data for the ViT-22B vision backbone and the LiT text backbone was created before these models were trained, therefore these images are not present in the training data. 
Further, the objects and scenes contained in these images are highly out-of-distribution relative to the distribution of natural images on the web.

\clearpage
\newpage

%% file: arxiv/text/appendix/model.tex
\section{Scalability}
\label{app:scalability}
When scaling up the default ViT architecture, we encountered training instability in ViT at Adam $1e^{-3}$. Initially, the loss would decrease as normal, but within 2000 steps the loss steadily increased. \cref{fig:stability_8b} shows the behavior of attention logits during training for an 8B parameter model. Without normalization, attention logits quickly grow to over $50000$ in magnitude, resulting in one-hot attention weights after the softmax, and subsequently unstable training losses and gradients.

\begin{figure}[htb]
    \centering
    \includegraphics[width=0.7\columnwidth]{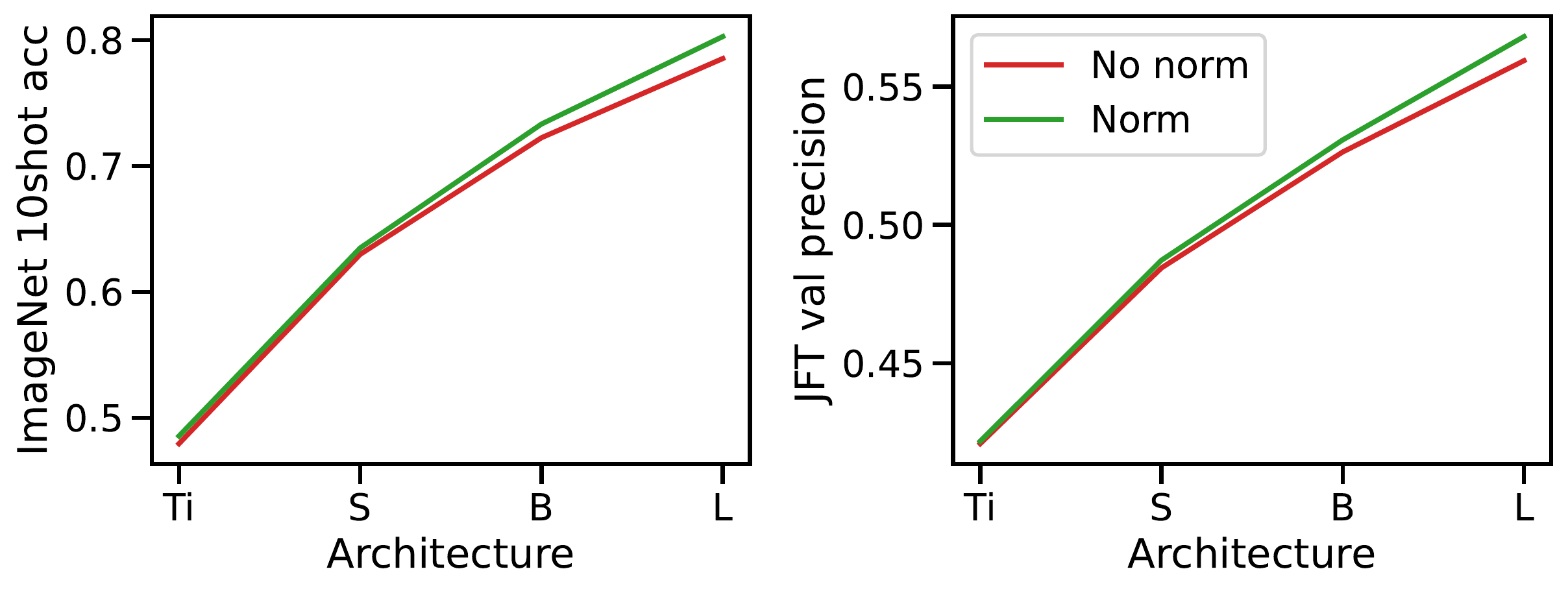}
    \caption{Training models with and without query/key normalization; those that do not have normalization are trained with lower learning rates at larger scale, whereas those with normalization have a consistent learning rate of 1e-3.}
    \label{fig:arch_sweep}
\end{figure}

To avoid instability, the learning rate of ViT was originally reduced with increasing model scale, from 1e-3 down to 4e-4 for ViT-H~\citep{dosovitskiy2020image}. We retrain models up to ViT-L, comparing models trained similar to ViT, to models which have the normalization/reduced precision. For the latter, the learning rate is kept at 1e-3 and not reduced for larger models. With the QK-normalization, the higher 1e-3 learning rate remains stable. The results, shown in~\cref{fig:arch_sweep}, demonstrate increasing benefits with scale, likely due to enabling the larger learning rate.

%% file: arxiv/text/appendix/model_card.tex
\newpage  
\section{Model Card}
\label{app:model_card}

\cref{tab:model_card} presents the model card~\citep{mitchell2019model} of the \chonk model.

{\fontsize{9pt}{9pt}\selectfont
\begin{longtable}[c]{ p{.05\textwidth} | p{.65\textwidth} }
\caption{Model Card of \chonk model.}
\label{tab:model_card}
\endfirsthead
\endhead
\toprule
\multicolumn{2}{c}{\textbf{Model Summary}}      \\ \toprule
\multicolumn{1}{l|}{Model Architecture} & 
Dense encoder-only model with 22 billion parameters. Transformer model architecture with variants to speed up and stabilize the training. For details, see Model Architecture~(\cref{sec:model_architecture}).  \\ \midrule
\multicolumn{1}{l|}{Input(s)} & 
The model takes images as input. %
\\ \midrule
\multicolumn{1}{l|}{Output(s)} & 
The model generates a class label as output during pretraining. %
\\
\toprule
\multicolumn{2}{c}{\textbf{Usage}}      \\ \toprule
\multicolumn{1}{l|}{Application} & 
The primary use is research on computer vision applications as a feature extractor that can be used in image recognition (finetuning, fewshot, linear-probing, zeroshot), dense prediction (semantic segmentation, depth estimation), video action recognition and so on. On top of that, \chonk is used in research that aim at understanding the impact of scaling vision transformers. 
\\ 
\midrule
\multicolumn{1}{l|}{Known Caveats} & 
When using \chonk, similar to any large scale model, it is difficult to understand how the model arrived at a specific decision, which could lead to lack of trust and accountability.

Moreover, we demonstrated that \chonk is less prone to unintentional bias and enhances current vision backbones by reducing spurious correlations. However, this was done through limited studies and particular benchmarks. Besides, there is always a risk of misuse in harmful or deceitful contexts when it comes to large scale machine learning models. 

\chonk should not be used for downstream applications without a prior assessment and mitigation of the safety and fairness concerns specific to the downstream application. We recommend spending enough time and energy on mitigation the risk at the downstream application level.  
\\
\toprule

\multicolumn{2}{c}{\textbf{System Type}}      \\ \toprule
\multicolumn{1}{l|}{System Description} & This is a standalone model.  \\ \midrule
\multicolumn{1}{l|}{Upstream Dependencies} & None.  \\ \midrule
\multicolumn{1}{l|}{Downstream Dependencies} & None. \\
\toprule
\multicolumn{2}{c}{\textbf{Implementation Frameworks}}      \\ \toprule
\multicolumn{1}{l|}{
\begin{minipage}{0.2\textwidth}
Hardware \& Software: \\ Training
\end{minipage}
} & 
Hardware: TPU v4~\citep{jouppi2020domain}. 
\vspace{0.05in}
Software: JAX~\citep{jax2018github}, Flax~\citep{flax2020github}, Scenic~\citep{dehghani2021scenic}.    
\\ \midrule
\multicolumn{1}{l|}{
\begin{minipage}{0.2\textwidth}
Hardware \& Software: \\ Deployment
\end{minipage}
} & Hardware: TPU v4~\citep{jouppi2020domain}. 
\vspace{0.05in}
Software: Scenic~\citep{dehghani2021scenic}. \\ \midrule
\multicolumn{1}{l|}{Compute Requirements} & \chonk was trained on 1024 TPU V4 chips for 177K steps.
\\
\hline 
\pagebreak  
\toprule
\multicolumn{2}{c}{\textbf{Model Characteristics}}      \\ \toprule
\multicolumn{1}{l|}{Model Initialization} & The model is trained from a random initialization.  \\ \midrule
\multicolumn{1}{l|}{Model Status} & This is a static model trained on an offline dataset.  \\ \midrule
\multicolumn{1}{l|}{Model Stats} & \chonk model has 22 billion parameters. \\
\toprule
\multicolumn{2}{c}{\textbf{Data Overview}}      \\ \toprule
\multicolumn{1}{l|}{Training Dataset} & \chonk is trained on a version of JFT~\citep{sun2017revisiting}, extended to contain around 4B images~\citep{zhai2022scaling}. See \cref{sec:training} for the description of datasets used to train \chonk.
\\ \midrule
\multicolumn{1}{l|}{Evaluation Dataset} & We evaluate the \chonk on a wide variety of tasks and report the results on each individual tasks and datasets~\citep{dehghani2021benchmark}. Specifically, we evaluate the models on:
ADE20K~\citep{zhou2017scene},
Berkeley Adobe Perceptual Patch Similarity (BAPPS)~\citep{zhang2018perceptual},
Birds~\citep{WahCUB_200_2011},
Caltech101~\citep{li_andreeto_ranzato_perona_2022},
Cars~\citep{KrauseStarkDengFei-Fei_3DRR2013},
CelebA~\citep{liu2015faceattributes},
Cifar-10~\citep{krizhevsky2009learning},
Cifar-100~\citep{krizhevsky2009learning},
CLEVR/count~\citep{johnson2017clevr},
CLEVR/distance~\citep{johnson2017clevr},
ColHist~\citep{kather2016multi},
DMLab ~\citep{beattie2016deepmind},
dSprites/location~\citep{dsprites17},
dSprites/orientation~\citep{dsprites17},
DTD~\citep{cimpoi2014describing},
EuroSAT~\citep{helber2019eurosat},
Flowers102~\citep{nilsback2008automated},
ImageNet~\citep{deng2009imagenet},
Inaturalist~\citep{cui2018large},
ImageNet-v2~\citep{recht2019imagenet_v2},
ImageNet-R~\citep{hendrycks2020imagenet_r}, 
ImageNet-A~\citep{hendrycks2021imagenet_a},
ImageNet-C~\citep{hendrycks2019benchmarking},
ImageNet-ReaL-H~\citep{tran2022plex},
Kinetics 400~\citep{kay2017kinetics},
KITTI~\citep{geiger2013vision},
Moments in Time~\citep{monfort2019moments},
ObjectNet~\citep{barbu2019objectnet}, 
Pascal Context~\citep{mottaghi2014role},
Pascal VOC~\citep{everingham2010pascal},
Patch Camelyon~\citep{teh2019metric},
Pets~\citep{parkhi2012cats},
Places365~\citep{zhou2017places},
Resisc45~\citep{cheng2017remote},
Retinopathy ~\citep{kaggle-diabetic-retinopathy},
SmallNORB/azimuth~\citep{lecun2004learning},
SmallNORB/elevation~\citep{lecun2004learning},
Sun397~\citep{xiao2010sun},
SVHN~\citep{netzer2011reading},
UC Merced~\citep{yang2010bag},
Waymo Open real-world driving dataset~\citep{sun2020scalability}.
\\
\hline 
\pagebreak  
\toprule
\multicolumn{2}{c}{\textbf{Evaluation Results}}  \\ 
\toprule
\multicolumn{1}{l|}{Benchmark Information} & 
\vspace{-5pt}
\begin{itemize}
    \addtolength{\itemsep}{-1ex}
    \item \textbf{Quality of transfer to downstream tasks.}
    \begin{itemize}
    \addtolength{\itemsep}{-1ex}
    \item Transfer to image classification (via linear probing, zero-shot transfer, OOD transfer, fewshot transfer, Head2Toe transfer, and fine-tuning).
    \begin{itemize}
    \addtolength{\itemsep}{-1ex}
    \item Datasets Used:
            Birds~\citep{WahCUB_200_2011},
            Caltech101~\citep{li_andreeto_ranzato_perona_2022},
            Cars~\citep{KrauseStarkDengFei-Fei_3DRR2013},
            CelebA~\citep{liu2015faceattributes},
            Cifar-10~\citep{krizhevsky2009learning},
            Cifar-100~\citep{krizhevsky2009learning},
            CLEVR/count~\citep{johnson2017clevr},
            CLEVR/distance~\citep{johnson2017clevr},
            ColHist~\citep{kather2016multi},
            DMLab ~\citep{beattie2016deepmind},
            dSprites/location~\citep{dsprites17},
            dSprites/orientation~\citep{dsprites17},
            DTD~\citep{cimpoi2014describing},
            EuroSAT~\citep{helber2019eurosat},
            Flowers102~\citep{nilsback2008automated},
            ImageNet~\citep{deng2009imagenet},
            iNaturalist~\citep{cui2018large},
            ImageNet-v2~\citep{recht2019imagenet_v2},
            ImageNet-R~\citep{hendrycks2020imagenet_r}, 
            ImageNet-A~\citep{hendrycks2021imagenet_a},
            ImageNet-C~\citep{hendrycks2019benchmarking},
            ImageNet-ReaL-H~\citep{tran2022plex},
            Kinetics 400~\citep{kay2017kinetics},
            KITTI~\citep{geiger2013vision},
            ObjectNet~\citep{barbu2019objectnet},
            Patch Camelyon~\citep{teh2019metric},
            Pets~\citep{parkhi2012cats},
            Places365~\citep{zhou2017places},
            Resisc45~\citep{cheng2017remote},
            Retinopathy ~\citep{kaggle-diabetic-retinopathy},
            SmallNORB/azimuth~\citep{lecun2004learning},
            SmallNORB/elevation~\citep{lecun2004learning},
            Sun397~\citep{xiao2010sun},
            SVHN~\citep{netzer2011reading},
            UC Merced~\citep{yang2010bag}.
    \end{itemize} 
    \vspace{+0.1cm}
    \item Transfer to video classification.
    \begin{itemize}
    \addtolength{\itemsep}{-1ex}
    \item Datasets Used: Kinetics 400~\citep{kay2017kinetics}, Moments in Time~\citep{monfort2019moments}.
    \end{itemize} 
    \vspace{+0.1cm}
    \item Transfer to dense prediction.
    \begin{itemize}
    \addtolength{\itemsep}{-1ex}
    \item Semantic segmentation
            \begin{itemize}
            \addtolength{\itemsep}{-1ex}
            \item Datasets Used: ADE20k~\citep{zhou2017scene}, Pascal Context~\citep{mottaghi2014role}, Pascal VOC~\citep{everingham2010pascal}.
            \end{itemize} 
    \vspace{+0.1cm}
    \item Depth estimation
            \begin{itemize}
            \addtolength{\itemsep}{-1ex}
            \item Dataset used: Waymo Open real-world driving dataset~\citep{sun2020scalability}.
            \end{itemize} 
    \end{itemize} 
    \end{itemize}  
    \item \textbf{Quality of learned features.}
    \begin{itemize}
    \addtolength{\itemsep}{-1ex}
        \item Fairness. 
            \begin{itemize}
            \addtolength{\itemsep}{-1ex}
            \item Dataset used: CelebA~\citep{liu2015faceattributes}.
            \end{itemize} 
        \vspace{+0.1cm}
        \item Human alignment.
            \begin{itemize}
            \addtolength{\itemsep}{-1ex}
            \item We used the {\tt model-vs-human} toolbox~\citep{geirhos2021partial}.
            \end{itemize} 
        \vspace{+0.1cm}
        \item Calibration.
            \begin{itemize}
            \addtolength{\itemsep}{-1ex}
            \item We follow the setup of \citet{minderer2021revisiting}. We use ImageNet validation set~\citep{deng2009imagenet}.
            \end{itemize} 
        \vspace{+0.1cm}
        \item Perceptual similarity.
            \begin{itemize}
            \addtolength{\itemsep}{-1ex}
            \item Dataset used:  Berkeley Adobe Perceptual Patch Similarity (BAPPS) dataset~\citep{zhang2018perceptual}
            \end{itemize} 
        \vspace{+0.1cm}
        \item Feature attribution. 
            \begin{itemize}
            \addtolength{\itemsep}{-1ex}
            \item Dataset used: ImageNet~\citep{deng2009imagenet}.
            \end{itemize} 
    \end{itemize} 
\end{itemize}  \vspace{-5pt} \\
\midrule
\multicolumn{1}{l|}{Evaluation Results} & All results are reported in ~\cref{sec:evaluation}.    \\ 
\hline 
\pagebreak  
\toprule
\multicolumn{2}{c}{\textbf{Model Usage \& Limitations}}      \\ \toprule
\multicolumn{1}{l|}{Sensitive Use} & \chonk should not be used for any unacceptable vision model use cases. For example: for detecting demographic human features for non-ethical purposes, as a feature extractor used to condition on and generate toxic content, or for captcha-breaking. We also do not approve use of \chonk in applications like surveillance, law enforcement, healthcare, or hiring and employment, and self-driving cars without putting measures in place to mitigate the ethical risks.
\\ \midrule
\multicolumn{1}{l|}{Known Limitations} & \chonk is designed for research. The model has not been tested in settings outside of research that can affect performance, and it should not be used for downstream applications without further analysis on factors in the proposed downstream application.  
\\ \midrule
\multicolumn{1}{l|}{
\begin{minipage}{0.2\textwidth}
Ethical Considerations \\ \& Risks
\end{minipage}
} & 
In order to train \chonk, we conducted an analysis of sensitive category associations on the JFT-4B dataset as described in \citet{aka2021measuring}. This process involved measuring the per label distribution of sensitive categories across the raw data, cleaned data, and models trained on this data, as well as labels verified by human raters. To further enhance the data quality, human raters also assisted in removing offensive content from the dataset. Our analysis using standard fairness benchmarks shows that \chonk increases performance for all subgroups while minimizing disparities among them. However, it is important to note that there may be situations where utilizing \chonk could pose ethical concerns. Therefore, we recommend conducting custom ethical evaluations for any new applications of \chonk.
\\
\bottomrule
\end{longtable}
}
\newpage  

%% file: arxiv/text/appendix/image_classification.tex
\section{Transfer to image classification: More results and addition details}

\subsection{Linear probing with L-BFGS}\label{app:linear}
An alternative to doing linear probing with SGD is to use the convex optimization technique, L-BFGS~\citep{byrd1995limited}. It is very effective and has strict convergence guarantees. We compare SGD and L-BFGS for a variety of ViT models using the ImageNet-1k datasset. Specifically, we precompute image embeddings by resizing input images to 224px resolution and then solve the multiclass logistic regression problem with L-BFGS. We also sweep the L2 regularization parameter and select the optimal one using 20000 holdout images from the training data (approximately 2\% of the training data). In \cref{tab:app:bfgs} we compare the resulting model with the SGD baseline from the main text. It demonstrates that L-BFGS matches or lags behind SGD approach, so we selected the latter technique for our core experiments.

\begin{table}[h]
    \caption{Comparison of SGD and L-BFGS for linear probing on ImageNet-1k. The numbers indicate top-1 accuracy.}
\centering
\small
  \setlength{\tabcolsep}{4pt}
    \begin{tabular}{@{} l c c c c c c c @{}}
      \toprule
	    Linear Probing & B/32 & B/16 & L/16 & g/14 & G/14 & e/14 & 22B \\
      \midrule
      L-BFGS & 79.94 & 84.06 & 86.64 & 88.46 & 88.79 & 89.08 & 89.27 \\
      SGD & 80.18 & 84.20 & 86.66 & 88.51 & 88.98 & 89.26 & 89.51 \\
      \bottomrule
 \end{tabular}
    \label{tab:app:bfgs}
\end{table}

\label{app:image_classification}

\subsection{Out of distribution classification}
\label{app:ood}

\begin{table}[th]
    \caption{
        OOD Classification. Results from models fine-tuned on ImageNet (top half), and models that were only trained on JFT and evaluated with a label-map (bottom half).
        Models with ``ema'' are fine-tuned with Polyak averaging, similar to~\citep{dosovitskiy2020image}.
        B/16, L/16, g/14, and G/14 are from~\citep{zhai2022scaling}, and e/14 is from~\citep{chen2022pali}.
        IN$^\dagger$ uses same resize without crop like in the original publication.
        See \cref{fig:robustness_objectnet} and \cref{subsec:robustness_ood} for discussion of the results, and details about datasets and pre-processing.
    }
    \centering
    \setlength{\tabcolsep}{5pt}
    \begin{tabular}{@{} l l c c c c c c @{}}
        \toprule
        Model & Fine-tuned & IN & IN$^\dagger$ & INv2 & ObjectNet & IN-R & IN-A \\
        \midrule
        e/14 ema & 560px &   \textbf{90.70} & \textbf{90.84} &  84.38 &     72.53 & 94.49 & 88.44 \\
        22B ema & 560px &   90.62 &              - &  \textbf{84.65} &     \textbf{76.70} & \textbf{95.05} & \textbf{89.12} \\
        22B & 560px &   90.60 & - &  84.38 &     75.69 & 94.62 & 88.55 \\
        22B & 384px &   90.44 & - &  84.28 &     74.64 & 94.44 & 87.95 \\
        e/14 ema & 384px &   90.44 & - &  83.95 &     70.56 & 93.56 & 87.16 \\
        G/14 ema & 518px &   90.33 & 90.47 &  83.53 &     69.14 & 94.22 & 86.95 \\
        g/14 ema & 518px &   90.25 & 90.11 &  83.61 &     71.36 & 93.37 & 86.12 \\
        L/16 & 384px &   88.60 & - &  80.74 &     65.73 & 90.32 & 78.65 \\
        B/16 & 384px &   87.02 & - &  78.21 &     57.83 & 82.91 & 66.08 \\
        \midrule
        22B & - &    \textbf{78.5} & - &   \textbf{72.5} &      \textbf{66.9} &  \textbf{91.6} &  \textbf{79.9} \\
        e/14 & - &    76.7 & - &   71.0 &      64.4 &  90.6 &  75.3 \\
        G/14 & - &    76.6 & - &   70.9 &      63.3 &  90.2 &  75.1 \\
        g/14 & - &    76.3 & - &   70.5 &      62.6 &  89.8 &  73.7 \\
        L/16 & - &    73.9 & - &   66.9 &      58.4 &  86.8 &  64.6 \\
        B/16 & - &    70.7 & - &   62.9 &      52.9 &  81.4 &  51.1 \\
        \bottomrule
    \end{tabular}
    \label{tab:robustness_ood}
\end{table}

\subsection{Head2Toe}
The cost of fine-tuning a model during transfer learning goes up with increased model size and often requires the same level of resources as training the model itself. Linear probing on the other hand is much cheaper to run, however it often performs worse than fine-tuning. Recent work showed that training a linear classifier on top of the intermediate features can provide significant gains compared to using the last-layer only, especially for target tasks that are significantly different from the original pre-training task~\citep{evci22h2t,Adler2020CrossDomainFL, khalifa2022}. 

In \cref{tab:h2t_vtab} we compare Head2Toe~\citep{evci22h2t} with Linear probe on common vision benchmarks and VTAB-1k~\citep{zhai2019large}. We include Finetuning results as a comparison point. We use a simplified version of Head2Toe with no feature selection. Experimental details are shared below. Head2Toe achieves 7\% better results on VTAB-1k, however fails to match the full finetuning performance (-6\%). On other benchmarks (CIFARs, Flowers and Pets), all methods perform similarly potentially. Head2Toe improves over Linear only for the Cifar-100 task. For the remaining tasks it either achieves the same performance or worse (Pets).

All experiments presented here use images with the default resolution of 224. Head2Toe uses the following intermediate features: (1) output of each of the 48 blocks, (2) features after the positional embedding, (3) features after the pooling head (4) pre-logits and logits. We average each of these features among the token dimension and concatenate them; resulting in a 349081 dimensional feature vector. In contrast, linear probe uses the 6144 dimensional prelogit features, which makes Head2Toe training roughly 50 times more expensive. However, given the extraordinary size of the original model, Head2Toe requires significantly less FLOPs and memory\footnote{on the order of 1000x, the exact value depends on number of classes} compared to fine-tuning. For all tasks (4 standard and 19 VTAB-1k), we search over 2 learning rates (0.01, 0.001) and 2 training lengths (500 and 10000 (2500 for VTAB-1k) steps) using the validation set.

\begin{table}[htbp]
    \centering
    \caption{Frozen evaluation using linear and Head2Toe (H2T) probe on the VTAB-1k benchmark and four other image classification tasks. We report mean accuracies averaged using 3 seeds.}
    \label{tab:h2t_vtab}
    \begin{tabular}{@{}lrrrrrrrr@{}}
    \toprule
    Method &  VTAB-Average &  Natural & Specialized &  Structured &  CIFAR-10 & CIFAR-100 &   Flowers  &      Pets   \\
    \midrule
    Finetuning   & \textbf{76.71} &  \textbf{89.09} & 87.08 & \textbf{61.83} & \textbf{99.63} &    \textbf{95.96} & 97.59  &  \textbf{99.75}\\
    Linear   & 63.15 &  80.86 & 87.05 & 35.70 &  99.37 &    93.39 &  \textbf{99.75} &  98.15 \\
    H2T & 70.12 &  84.60 & \textbf{88.61} & 48.19 &  99.45 &  94.11 &  99.69 &  97.46 \\
    \bottomrule
    \end{tabular}
%
 
\end{table}

\subsection{Few-shot}
\begin{figure*}[h]
    \centering
    \includegraphics[width=0.95\columnwidth]{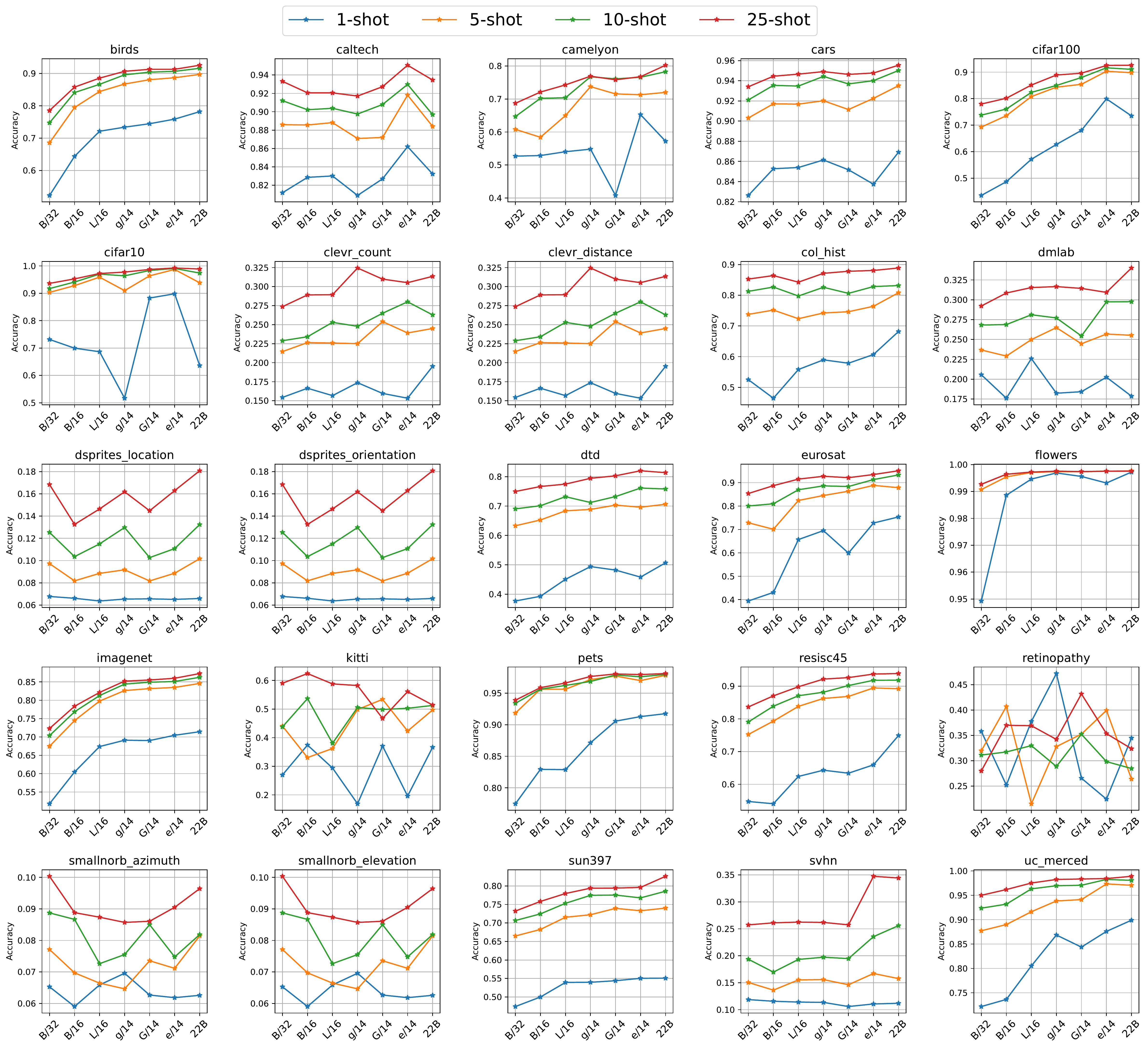}
    \vspace{-10pt}
    \caption{Few-shot transfer with 1, 5, 10, and 25 shots on 25 vision tasks~\citep{abnar2021exploring}.}
    \vspace{-15pt}
    \label{fig:fewshot}
\end{figure*}

We replicate the experimental setup of ~\citep{abnar2021exploring} to evaluate the \chonk model and baselines on 25 tasks (\cref{table:fewshot_datasets}) using few-shot transfer setups. The results of few-shot transfer of different models using 1, 5, 10, and 25 shots are presented in \cref{fig:fewshot}. 
Scaling up can improve performance in many tasks, but in some cases, downstream accuracy does not improve with increased scale. This may be due to the higher dimension of the representation from the \chonk model, which may require more regularization as the size of the head grows to prevent overfitting. Further study is needed to investigate this. 

\begin{table}[ht]
    \centering
    \caption{Summary of datasets used in our few-shot experiments in \cref{fig:fewshot}}
    {\fontsize{9}{9}\selectfont
    \resizebox{0.95\columnwidth}{!}{%
    \begin{tabular}{l p{12cm} p{5cm}}
     \textbf{Dataset}    & \textbf{Description} & \textbf{Reference} \\ \toprule
      ImageNet   & 1.28M labelled natural images. &~\citep{deng2009imagenet}\\ \midrule
      Caltech101 &The task consists in classifying pictures of objects (101 classes plus a background clutter class), including animals, airplanes, chairs, or scissors. The image size varies, but it typically ranges from 200-300 pixels per edge. &\citep{li_andreeto_ranzato_perona_2022} \\ \midrule
     CIFAR-10 & The task consists in classifying natural images (10 classes, with 6000 training images each). Some examples include apples, bottles, dinosaurs, and bicycles. The image size is 32x32.& \url{https://www.cs.toronto.edu/~kriz/cifar.html}\\ \midrule
      CIFAR-100 & The task consists in classifying natural images (100 classes, with 500 training images each). Some examples include apples, bottles, dinosaurs, and bicycles. The image size is 32x32.& \url{https://www.cs.toronto.edu/~kriz/cifar.html}\\ \midrule
      DTD & The task consists in classifying images of textural patterns (47 classes, with 120 training images each). Some of the textures are banded, bubbly, meshed, lined, or porous. The image size ranges between 300x300 and 640x640 pixels. &~\citep{cimpoi2014describing}\\ \midrule
      Pets & The task consists in classifying pictures of cat and dog breeds (37 classes with around 200 images each), including Persian cat, Chihuahua dog, English Setter dog, or Bengal cat. Images dimensions are typically 200 pixels or larger.& \url{https://www.robots.ox.ac.uk/~vgg/data/pets/} \\ \midrule
      Sun397 & The Sun397 task is a scenery benchmark with 397 classes and, at least, 100 images per class. Classes have a hierarchy structure and include cathedral, staircase, shelter, river, or archipelago. The images are (colour) 200x200 pixels or larger.& \url{https://vision.princeton.edu/projects/2010/SUN/}\\ \midrule
      Flowers102 & The task consists in classifying images of flowers present in the UK (102 classes, with between 40 and 248 training images per class). Azalea, Californian Poppy, Sunflower, or Petunia are some examples. Each image dimension has at least 500 pixels.& \url{https://www.robots.ox.ac.uk/~vgg/data/flowers/102/}\\ \midrule
      SVHN & This task consists in classifying images of Google’s street-view house numbers (10 classes, with more than 1000 training images each). The image size is 32x32 pixels.& \url{http://ufldl.stanford.edu/housenumbers/}\\ \midrule
      CLEVR/count & CLEVR is a visual question and answer dataset designed to evaluate algorithmic visual reasoning. We use just the images from this dataset, and create a synthetic task by setting the label equal to the number of objects in the images.& ~\citep{johnson2017clevr}\\ \midrule
      CLEVR/distance & Another synthetic task we create from CLEVR consists of predicting the depth of the closest object in the image from the camera. The depths are bucketed into size bins. &~\citep{johnson2017clevr}\\ \midrule
      Retinopathy & The Diabetic Retinopathy dataset consists of image-label pairs with high-resolution retina images, and labels that indicate the presence of Diabetic Retinopathy (DR) in a 0-4 scale (No DR, Mild, Moderate, Severe, or Proliferative DR). & \url{https://www.kaggle.com/c/diabetic-retinopathy-detection/data} \\ \midrule 
      Birds &  Image dataset with photos of 200 bird species (mostly North American). &~\citep{WahCUB_200_2011} \\ \midrule
      
      Patch Camelyon & The Patch Camelyon dataset contains 327,680 images of histopathologic scans of lymph node sections. The classification task consists in predicting the presence of metastatic tissue in given image (i.e., two classes). All images are 96x96 pixels. &~\citep{teh2019metric}\\ \midrule
      
        Resisc45 & The Remote Sensing Image Scene Classification (RESISC) dataset is a scene classification task from remote sensing images. There are 45 classes, containing 700 images each, including tennis court, ship, island, lake, parking lot, sparse residential, or stadium. The image size is RGB 256x256 pixels. &~\citep{cheng2017remote} \\ \midrule
        EuroSAT & The task consists in classifying Sentinel-2 satellite images into 10 different types of land use (Residential, Industrial, River, Highway, etc). The spatial resolution corresponds to 10 meters per pixel, and the image size is 64x64 pixels. &~\citep{helber2019eurosat} \\ \midrule
        dSprites/location & The dSprites dataset was originally designed to assess disentanglement properties of unsupervised learning algorithms. In particular, each image is a 2D shape where six factors are controlled: color, shape, scale, rotation, and (x,y) center coordinates. Images have 64x64 black-and-white pixels. This task consists in predicting the x (horizontal) coordinate of the object. The locations are bucketed into 16 bins & \url{https://github.com/deepmind/dsprites-dataset/}\\ \midrule
        dSprites/orientation & We create another task from dSprites consisting in predicting the orientation of each object, bucketed into 16 bins. & \url{https://github.com/deepmind/dsprites-dataset/https://github.com/deepmind/dsprites-dataset/} \\ \midrule
        SmallNORB/azimuth & The Small NORB dataset contains images of 3D-toys from 50 classes, including animals, human figures, airplanes, trucks, and cars. The image size is 640x480 pixels. In this case, we define labels depending on the azimuth (angle of horizontal deviation), in intervals of 20 degrees (18 classes). &~\citep{lecun2004learning}\\ \midrule
        SmallNORB/elevation & Another synthetic task we create from Small NORB consists in predicting the elevation in the image. There are 9 classes, corresponding to 9 different elevations ranging from 30 to 70 degrees, in intervals of 5 degrees & ~\citep{lecun2004learning} \\ \midrule
        DMLab & The DMLab (DeepMind Lab) is a set of control environments focused on 3D navigation and puzzle-solving tasks. The Dmlab dataset contains frames observed by the agent acting in the DeepMind Lab environment, which are annotated by the distance between the agent and various objects present in the environment. The goal is to evaluate the ability of a visual model to reason about distances from the visual input in 3D environments. The Dmlab dataset consists of 360x480 color images in 6 classes. The classes are {close, far, very far} × {positive reward, negative reward} respectively. &~\citep{beattie2016deepmind} \\ \midrule
        KITTI & The KITTI task consists in predicting the (binned) depth to the vehicle (car, van, or truck) in the image. There are 4 bins / classes. &~\citep{geiger2013vision}\\ \midrule
        ColHist & Classification of textures in colorectal cancer histology. Each example is a 150 x 150 x 3 RGB image of one of 8 classes. & \url{https://www.tensorflow.org/datasets/catalog/colorectal_histology} \\ \midrule
        UC Merced & 21 class land use image dataset & \url{https://usdahsi.ucmerced.edudatasets/landuse.html} \\
        \midrule
        Cars &  The Cars dataset contains 16,185 images of 196 classes of cars. The data is split into 8,144 training images and 8,041 testing images, where each class has been split roughly in a 50-50 split. Classes are typically at the level of Make, Model, Year, e.g. 2012 Tesla Model S or 2012 BMW M3 coupe.& \url{http://ai.stanford.edu/~jkrause/cars/car_dataset.html} \\ \bottomrule

    \end{tabular}
    }}
    \label{table:fewshot_datasets}
\end{table}

\clearpage
\newpage

%% file: arxiv/text/appendix/dense_prediction.tex
\section{Transfer to dense prediction: More results and addition details.}

\subsection{Semantic segmentation: frozen \textit{versus} fine-tuning.}

\label{app:semantic_segmentation}
In this experiment, we evaluate the effect of fine-tuning \textit{versus} freezing the \chonk backbone when transferring to semantic segmentation.
The results are shown in \cref{tab:semseg_frozen_finetune}.
We observe that for the linear decoder fine-tuning results in much better performance than using frozen features.
For the UperNet decoder, however, the gap between fine-tuning and freezing the backbone is much smaller.
This can be explained by the fact that UperNet has $\sim870$ times more parameters than the linear model.
\cref{fig:dense_prediction} shows qualitative results using Upernet.

\begin{table}[h]
    \caption{
      {Frozen \textit{versus} fine-tuning transfer of \chonk to semantic segmentation.} 
We report mean IoU on the validation set of 3 popular datasets, namely ADE20k (``A-150'')~\citep{zhou2017scene}, Pascal Context (``P-60'')~\citep{mottaghi2014role}, Pascal VOC and (``P-20'')~\citep{everingham2010pascal}, for different protocols:
(i) frozen \textit{versus} finetuned backbone;
(ii) linear~\citep{strudel2021segmenter} \textit{versus} UperNet~\citep{xiao2018unified} decoder.
}
\centering
    \begin{tabular}{@{} l c ccc c ccc @{}}
      \toprule
      Decoder && \multicolumn{3}{c}{Linear} && \multicolumn{3}{c}{UperNet} \\
\cmidrule{3-5}\cmidrule{7-9}
	      Dataset && A-150 & P-60 & P-20 && A-150 & P-60 & P-20 \\
      \midrule
      \chonk frozen && 34.6 & 38.6 & 65.0 && 52.7 & 58.7 & 78.7 \\
      \chonk fine-tuned\hspace{-1em} && 54.9 & 61.6 & 79.0 && 55.3 & 62.3 & 80.1\\
      \bottomrule
 \end{tabular}
    \label{tab:semseg_frozen_finetune}
\end{table}

\subsection{Monocular Depth Estimation}
\label{app:de}

\subsubsection{Dataset}

We pre-process Waymo Open video and LiDAR data to obtain RGB frames and associated sparse depth images. The camera frames are extracted from the front-facing camera mounted on the vehicle, while the sparse depth images are obtained by projecting the LiDAR point cloud of a single time step onto the camera frame. We use the camera and LiDAR calibration parameters to compute the distance of each LiDAR point to the camera. For training, we normalize the depth targets using a $\mathrm{log}(1+x)$ transformation; we undo this transformation for metric computation. As the signal is sparse, we mask out any pixels in the depth image for which there is no signal during loss computation.
We evaluate on the first 5120 validation set frames from the front facing camera.

We sub-sample videos to 5 fps, and crop and resize frames to $224\times224$ resolution (both RGB inputs and depth targets). The LiDAR projection is done after cropping and resizing, to retain a high-quality signal. For ViT-L, we upscale the RGB input frames to $256\times256$ resolution to account for the larger patch size, while keeping the same information content as for ViT-e and ViT-22B, which both use a patch size of 14. For evaluation frames, we use a simple center-crop. For training, we use Inception-style~\citep{szegedy2015going} random-resized crops as our only form of data augmentation. We ensure that at least $20\%$ of the original frame is retained after cropping.

For efficiency reasons, we pre-compute ViT-22B feature maps for 1,024,000 randomly sampled and augmented frames from the training set, which amounts to approx.~6.4 epochs of training data. When training the decoder, we iterate over these pre-computed feature maps in random order, irrespective of the number of training steps used. We follow the same protocol for all compared models.

\subsubsection{Decoder Architectures}

\paragraph{Dense Prediction Transformer.}

We largely follow the design of~\citep{ranftl2021vision}, using four reassemble and fusion blocks that processes the $16\times16$ ViT feature map at $(4\times4)$, $(8\times8)$, $(16\times16)$, and $(32\times32)$ spatial resolutions.
We use $64$ features at each stage and thus can omit the $1\times1$ projection convolution in the fusion block. 
The final fusion stage feeds into a monocular depth estimation head, where we use the default $128$ features and adjust the final re-sampling stage to yield the desired resolution of $224\times224$.
Similar to~\citep{ranftl2021vision}, we do not consider dropout or batchnorm for depth estimation.

For efficiency purposes we reuse the same $16\times16$ ViT feature map at each stage.
We empirically verified that this did not significantly impact results and our implementation of DPT using four ViT-22B feature maps (from layers 12, 24, 36, and 48) normalized using LayerNorm obtained similar scores to what was reported in Table~\ref{tab:depth_estimation}: $0.021$ MSE, $0.098$ AbsRel, $0.686$ $\delta < 1.1$, 0.906 $\delta < 1.25$, 0.979 $\delta < 1.25^2$.
Directly feeding pre-norm feature maps led to instabilities.

\begin{figure}[h]
    \centering
    \subfigure{\includegraphics[width=0.15\textwidth]{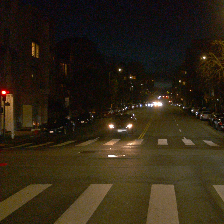}} \
    \subfigure{\includegraphics[width=0.2\textwidth]{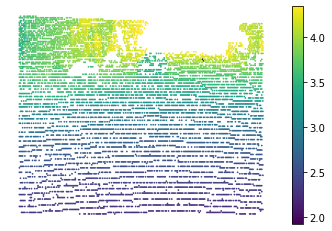}} \
    \subfigure{\includegraphics[width=0.2\textwidth]{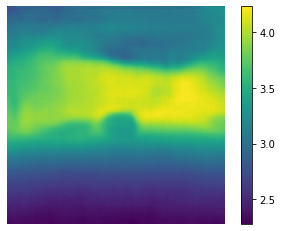}} \
    \subfigure{\includegraphics[width=0.2\textwidth]{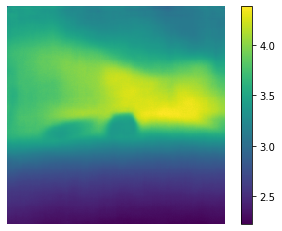}} \
    \subfigure{\includegraphics[width=0.2\textwidth]{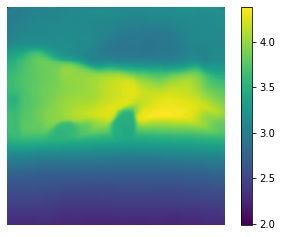}}
    \subfigure{\includegraphics[width=0.15\textwidth]{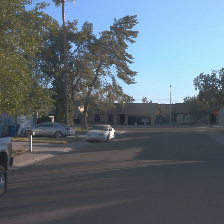}} \
    \subfigure{\includegraphics[width=0.2\textwidth]{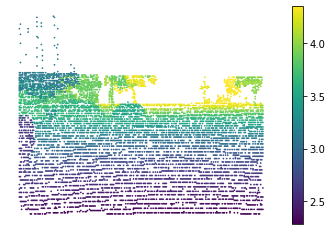}} \
    \subfigure{\includegraphics[width=0.2\textwidth]{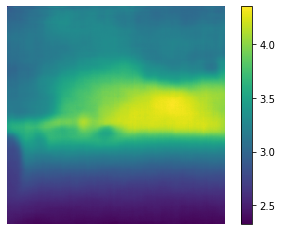}} \
    \subfigure{\includegraphics[width=0.2\textwidth]{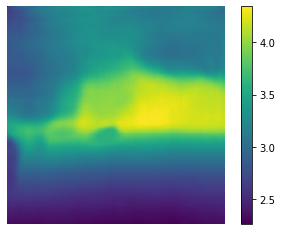}} \
    \subfigure{\includegraphics[width=0.2\textwidth]{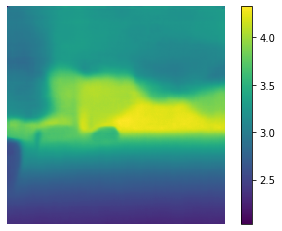}}
    \addtocounter{subfigure}{-10}
    \subfigure[Input frame]{\includegraphics[width=0.15\textwidth]{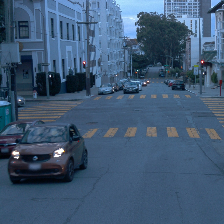}} \
    \subfigure[Sparse depth target]{\includegraphics[width=0.2\textwidth]{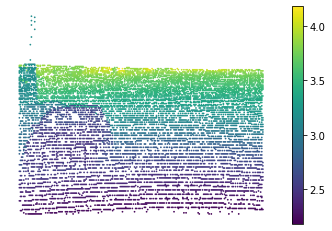}} \
    \subfigure[DPT (ViT-L)]{\includegraphics[width=0.2\textwidth]{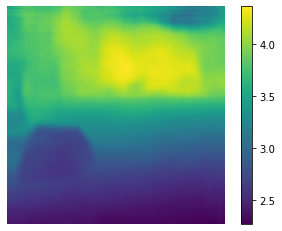}} \
    \subfigure[DPT (ViT-e)]{\includegraphics[width=0.2\textwidth]{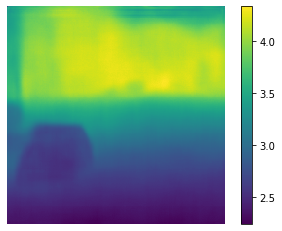}} \
    \subfigure[DPT (ViT-22B)]{\includegraphics[width=0.2\textwidth]{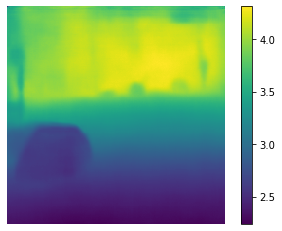}}
    \caption{
    {Monocular depth estimation}: (c, d, e) show estimated depth by a DPT head applied on ViT features, (a) shows the input frame and (b) shows the sparse ground-truth depth maps. Notice how (eg. in the third row) the model manages to go well beyond the available depth targets and makes what appear to be reasonable predictions for far-away cars, even though they are well out of LiDAR range. Ground-truth depth and depth predictions are visualized in $\log(1 + \text{depth})$ space.}
    \label{fig:dpt_depth_predictions_full}
\end{figure}

\begin{figure}[t!]
    \centering
    \subfigure{\includegraphics[width=0.15\textwidth]{figs/depth_prediction/DPT-1-ViT-22b-48459029-2/1_video.png}} \
    \subfigure{\includegraphics[width=0.2\textwidth]{figs/depth_prediction/DPT-1-ViT-22b-48459029-2/1_depth_scatter.png}} \
    \subfigure{\includegraphics[width=0.2\textwidth]{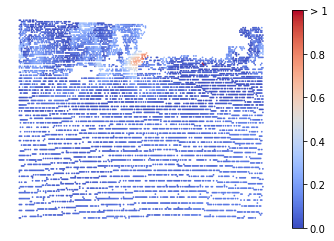}} \
    \subfigure{\includegraphics[width=0.2\textwidth]{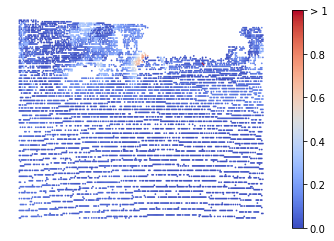}} \
    \subfigure{\includegraphics[width=0.2\textwidth]{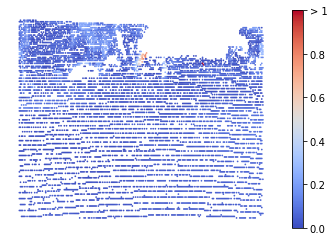}}
    \subfigure{\includegraphics[width=0.15\textwidth]{figs/depth_prediction/DPT-1-ViT-22b-48459029-2/2_video.png}} \
    \subfigure{\includegraphics[width=0.2\textwidth]{figs/depth_prediction/DPT-1-ViT-22b-48459029-2/2_depth_scatter.png}} \
    \subfigure{\includegraphics[width=0.2\textwidth]{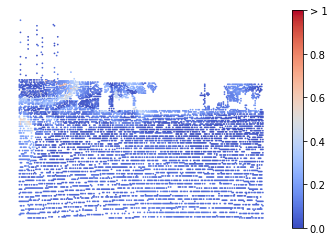}} \
    \subfigure{\includegraphics[width=0.2\textwidth]{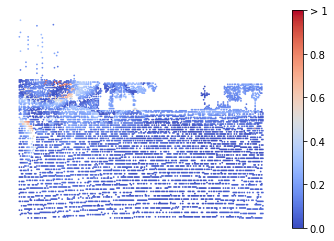}} \
    \subfigure{\includegraphics[width=0.2\textwidth]{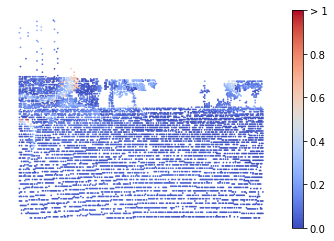}}
    \addtocounter{subfigure}{-10}
    \subfigure[Input frame]{\includegraphics[width=0.15\textwidth]{figs/depth_prediction/DPT-1-ViT-22b-48459029-2/3_video.png}} \
    \subfigure[Sparse depth target]{\includegraphics[width=0.2\textwidth]{figs/depth_prediction/DPT-1-ViT-22b-48459029-2/3_depth_scatter.png}} \
    \subfigure[DPT (ViT-L)]{\includegraphics[width=0.2\textwidth]{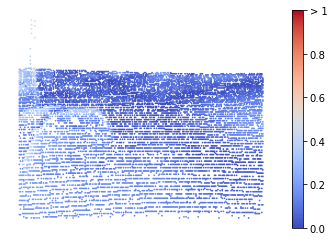}} \
    \subfigure[DPT (ViT-e)]{\includegraphics[width=0.2\textwidth]{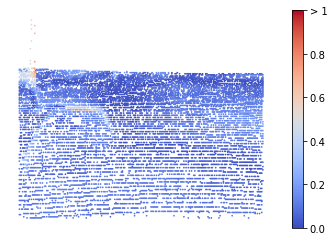}} \
    \subfigure[DPT (ViT-22B)]{\includegraphics[width=0.2\textwidth]{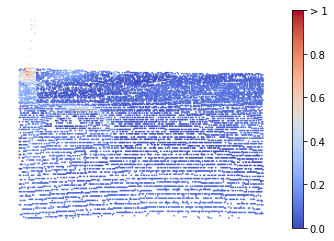}}
    \caption{
    {Monocular depth estimation errors}: (c, d, e) show absolute depth estimation errors by a DPT head applied on ViT features (for points where the ground truth is available), (a) shows the input frame and (b) shows the sparse ground truth depth maps. Notice how the outline of the car is clearly visible (light blue) from the sparse error maps in the third row when using ViT-L or ViT-e, while using \chonk leads to fewer errors in this region (darker blue). Ground-truth depth and absolute prediction errors are visualized in $\log(1 + \text{depth})$ space.}
    \label{fig:dpt_depth_predictions_full_error}
\end{figure}

\paragraph{Linear decoder.}

The linear decoder processes $16\times 16 \times 6144~$ ViT-22B feature maps using two transpose convolution layers of stride 2 and kernel size 5, followed by a $1\times 1$ convolution that outputs a low resolution depth map.
This is resized to $224\times 224$ using bilinear interpolation and clipped at $0$ to be valid predictions for $log\left(1 + \mathrm{depth}\right)$
The intermediate feature maps have $1536$ and $768$ dimensions each.
Linear activations are used in between layers so that the whole decoder is end-to-end linear.
This performed marginally better than a single $11\times 11$ stride $4$ transpose convolution layer, although the single layer decoder should be equally powerful in theory.
We suspect that this has to do with how hyper-parameters have been empirically optimized for smaller kernel sizes.

For the ViT-e and ViT-L baselines, the linear decoder is exactly the same except for a much smaller input feature dimension ($1792$ for ViT-e and $1024$ for ViT-L).
Thus the linear decoder on top of ViT-22B has more capacity than the same on top of ViT-e or ViT-l.
We controlled for this in two ways: (a) using a $1\times 1$ convolution on the ViT-22B features, we down-project them to $1792$ dimensions to match the feature map size of ViT-e, or (b) using a large hidden dimension ($4096$ in ViT-e's decoder and $6144$ in ViT-L's decoder) after the first convolution transpose layer, we approximately matched the number of parameters across the three models. In control (a), performance stayed roughly the same at $0.165$ relative absolute error (AbsRel) for ViT-22B. In control (b) performance for baselines did not change substantially in terms of relative absolute error, $0.208$ for ViT-e and $0.222$ for ViT-L. We therefore report results without these controls in~\cref{tab:depth_estimation}.

\subsubsection{Training Details}

We train the decoder for 300k steps with a batch size of 64 using Adam~\citep{kingma2014adam} and clip the gradients to a global norm value of 0.05 to stabilize training.
We linearly increase the learning rate for 2500 steps to 0.0002 (starting from 0) and then decay the learning rate with a cosine schedule~\citep{loshchilov2016sgdr} back to 0.

\subsubsection{Metrics}

We quantify performance using the following standard depth estimation metrics from the literature~\citep{hermann2020self, eigen2014depth}, and also report the MSE loss on the validation set:
AbsRel measures the mean absolute error between the ground truth and predicted depth relative to the ground truth, while the inlier fraction metrics ($\delta$) measure the fraction of valid pixels within a certain 
percentage
from ground truth.
All metrics were measured after undoing the log transformation.

\subsubsection{Qualitative Results}

We report qualitative depth predictions by DPT from different ViT backbones in \Cref{fig:dpt_depth_predictions_full}, and absolute prediction errors in \Cref{fig:dpt_depth_predictions_full_error}.

%% file: arxiv/text/appendix/video_classification.tex
\section{Video Classification}
\label{sec:app_video_classification}

We sample 128 and 32 frames with a stride of 2 frames from Kinetics 400 videos~\citep{kay2017kinetics} and Moments in Time~\citep{monfort2019moments} videos, respectively. For both ViT-22B and ViT-e we rely in the frozen, pre-trained models and use the pre-logit feature representation to extract a single embedding per frame, resulting in a token sequences of length 128 and 32, respectively, which are then processed by a shallow transformer model equipped with a class-token classifier.

This is in contrast to CoCa~\citep{yu2022coca}, which uses one token per image patch for their video classification experiments and a resolution of 576px (compared 224px in our experiments), resulting in much longer token sequences. We explored using one token per image patch (i.e. unpooled features) in preliminary experiments, but found that this leads to inferior performance. One potential reason for this could be that CoCa applies a contrastive loss to a pooled feature representation, and additionally feeds the unpooled token sequences to a generative decoder, which might lead to a different structure in the unpooled representation than the supervised classification loss used to pretrain ViT-22B and ViT-e.

To facilitate experimentation, we pre-compute frozen features for the two ViT variants we consider, using the same augmentations as~\citep{arnab2021vivit}. To improve the robustness of our model and prevent overfitting we feed the entire training set ten times, with different data augmentations for every pass. We train for 30 epochs on these precomputed features with a batch size of 256 using SGD with momentum and with a cosine schedule including a linear warmup of 2.5 epochs. We sweep the following hyperparameters and corresponding value ranges to train our video model: $\{1, 2\}$ transformer layers of width $\{1024, 2048, 4096\}$, using a learning rate in $\{10^{-1}, 10^{-2}\}$ and a weight decay in $\{10^{-3}, 10^{-2}\}$.

%% file: arxiv/text/appendix/fairness.tex
\section{Fairness}\label{app:fairness}
\newlength{\fairnessfigurewidth}\setlength{\fairnessfigurewidth}{0.6\columnwidth}

\begin{figure}[htbp]
    \centering
    \includegraphics[width=\fairnessfigurewidth]{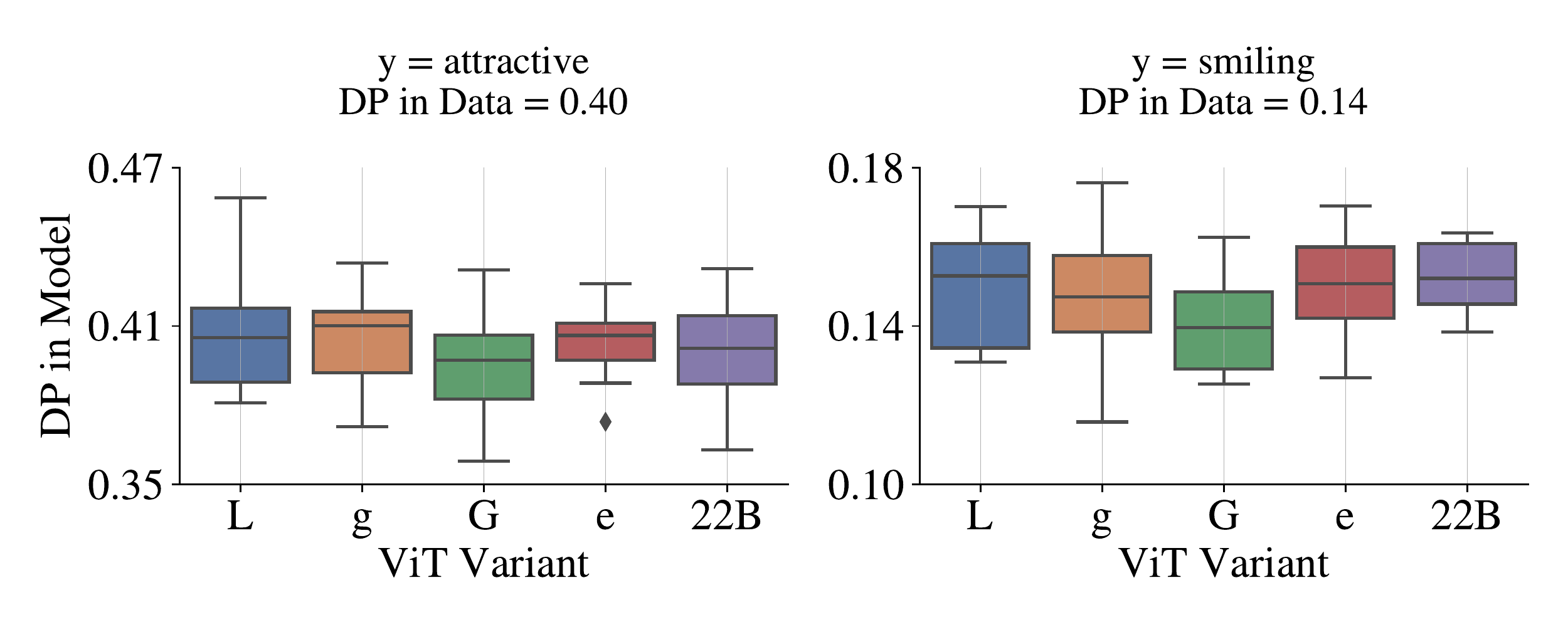}
    \vspace{-10pt}
    \caption{DP in the model often reflects DP in the data in the absence of bias mitigation. In this figure, binary sex is the sensitive attribute and linear heads are trained to predict other attributes in CelebA using pretrained features.
    }
    \label{fig:fairness:bias_original}
    \vspace{-10pt}
\end{figure}

\begin{figure}[htbp]
    \centering
    \includegraphics[width=\fairnessfigurewidth]{figs/fairness/perf_acc.v3.pdf}
    \includegraphics[width=\fairnessfigurewidth]{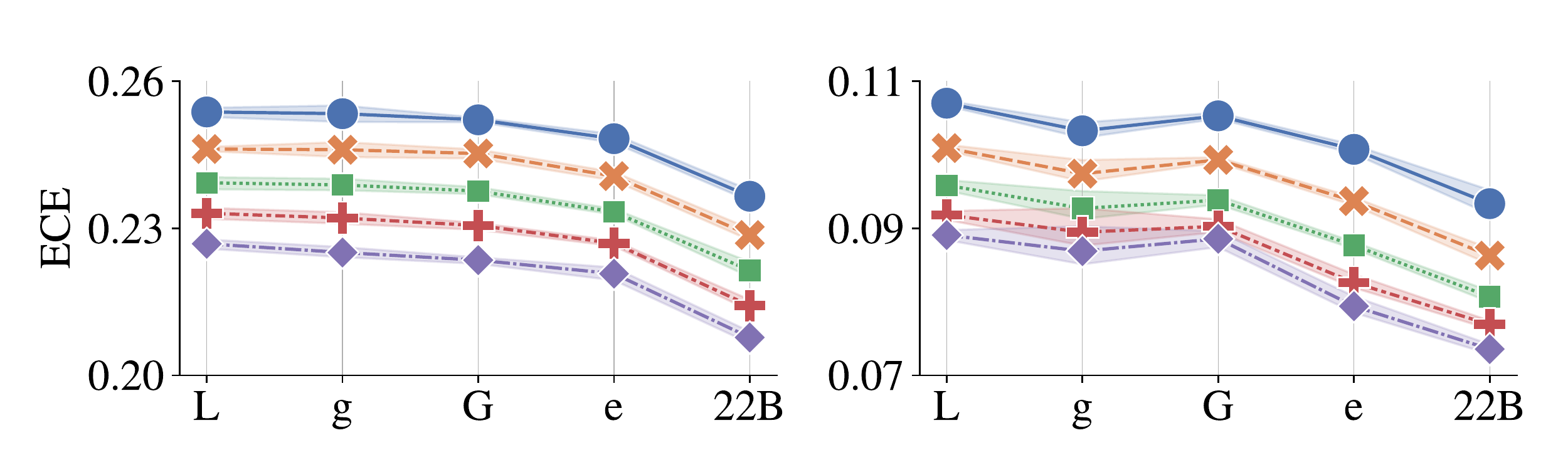}
    \includegraphics[width=\fairnessfigurewidth]{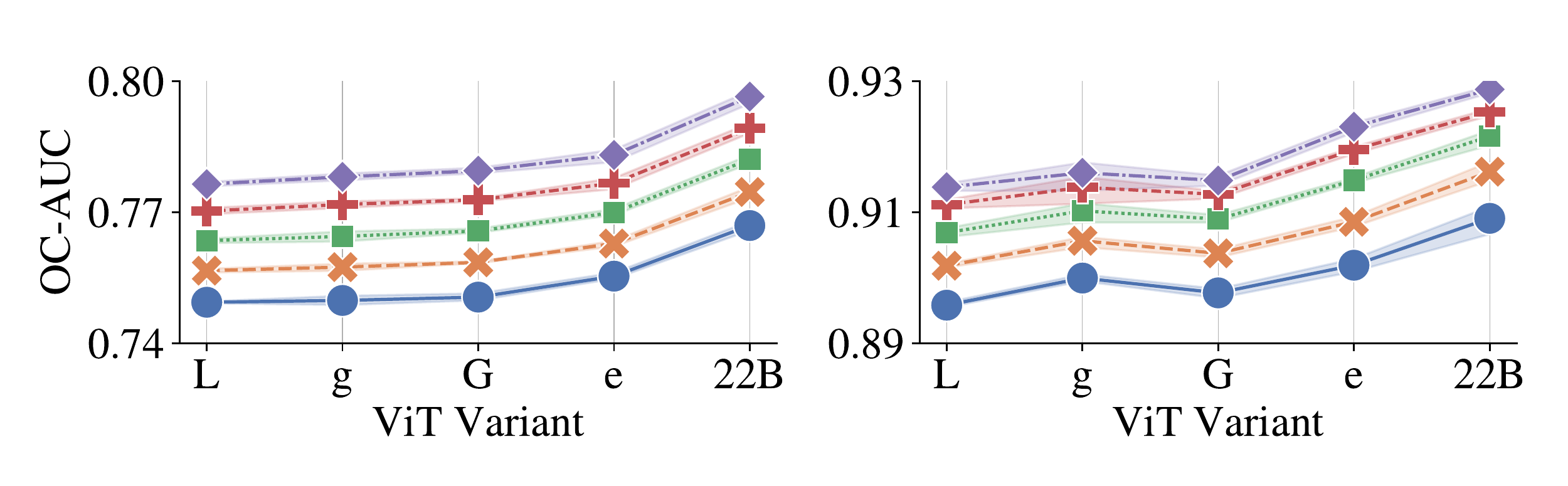}\vspace{-1em}
    \caption{Performance, in terms of either accuracy (top),  ECE (middle), or OC-AUC (bottom), is plotted for each ViT variant \emph{after} debiasing the model to meet the prescribed level of bias shown in the legends. Refer to \cref{sect::fairness} for details.}
    \label{fig:fairness:perf}
\end{figure}

\begin{figure}[htbp]
    \centering
    \includegraphics[width=\fairnessfigurewidth]{figs/fairness/all_benefit_acc_1.v3.pdf}
    \includegraphics[width=\fairnessfigurewidth]{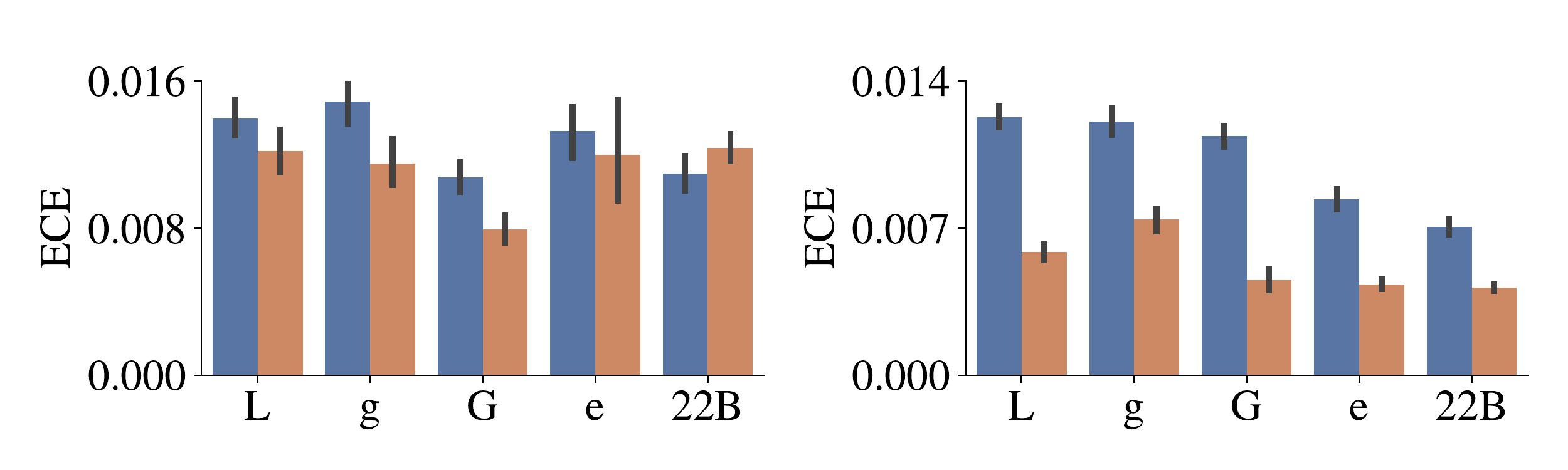}
    \includegraphics[width=\fairnessfigurewidth]{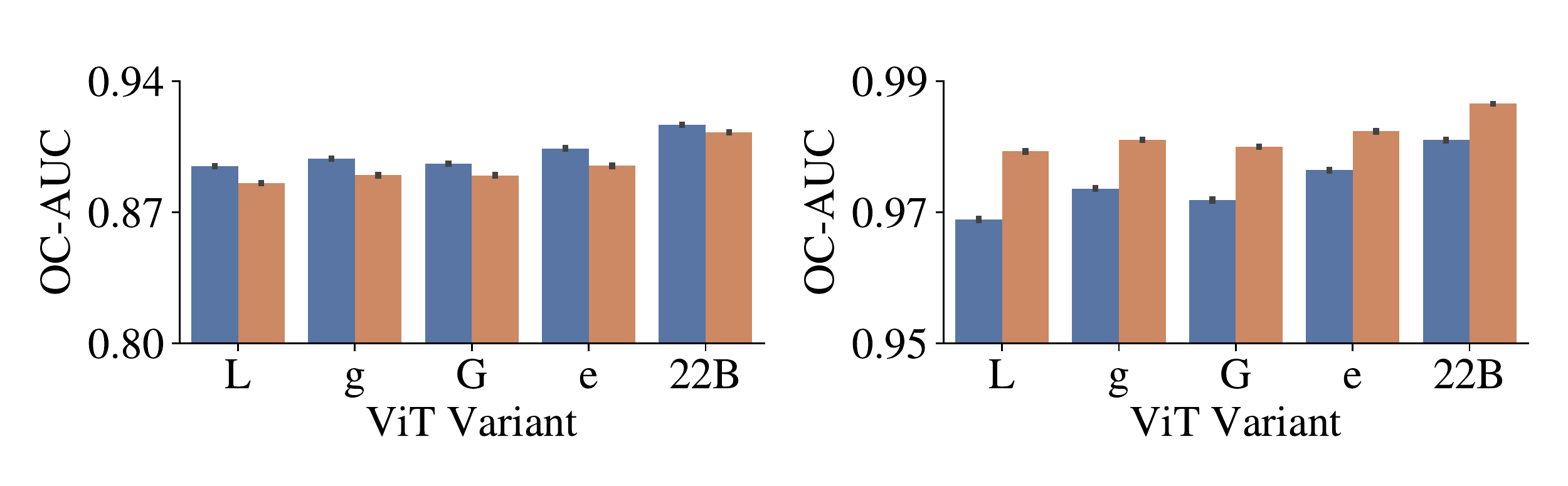}\vspace{-1em}
    \caption{Performance is plotted for each subgroup in CelebA \emph{prior to} bias mitigation. \chonk offers better performance overall across all three metrics, not just overall, but also within each subgroup separately. Refer to \cref{sect::fairness} for details.}
    \label{fig:fairness:allbenefit}
\end{figure}

\begin{figure}[htbp]
    \centering
    \includegraphics[width=\fairnessfigurewidth]{figs/fairness/parity_acc_1.v3.pdf}
    \includegraphics[width=\fairnessfigurewidth]{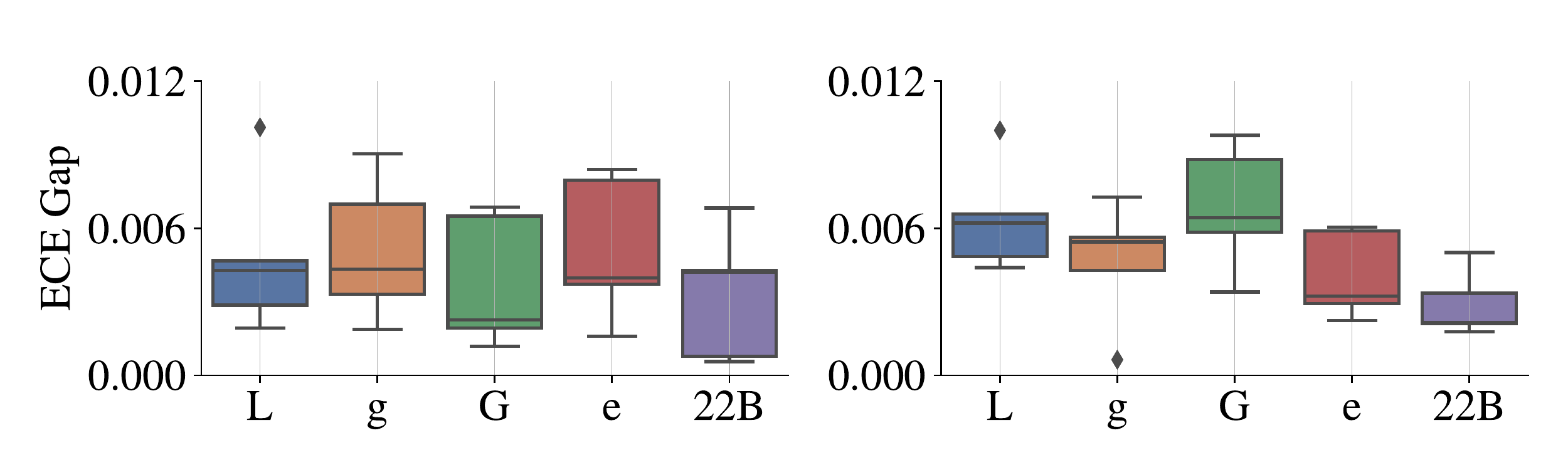}
    \includegraphics[width=\fairnessfigurewidth]{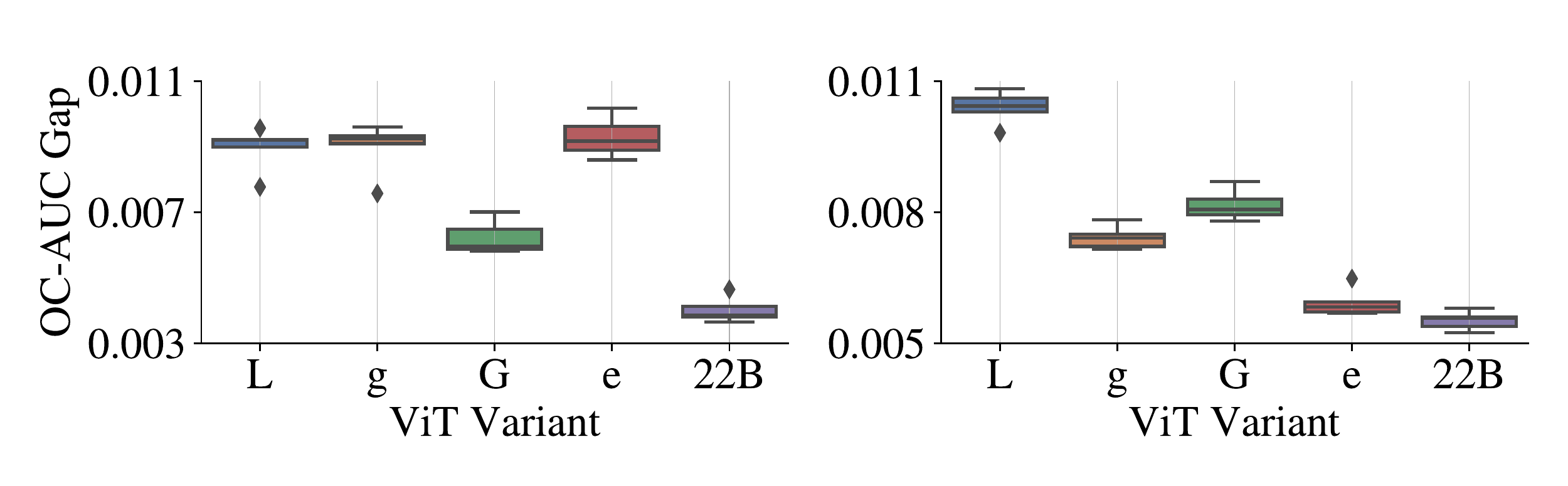}\vspace{-1em}
    \caption{The $y$-axis is the absolute difference in performance across the two subgroups: females and males. \chonk provides a more equitable performance, compared to earlier/smaller ViT architectures in all three metrics.}
    \label{fig:fairness:parity}
\end{figure}

We report the full experimental results described in \cref{sect::fairness} for all three evaluation (1) classification accuracy (denoted ACC), (2) expected calibration error (ECE)~\citep{naeini2015obtaining,guo2017calibration}, and (3) Oracle Collaborative AUC (OC-AUC)~\citep{kivlichan2021measuring}. 
ECE is used to measure calibration, while OC-AUC computes the four variables: binned true/false positives/negatives, as a function of a linearly spaced set of thresholds and score bins. The full results are presented in \cref{fig:fairness:perf}, \cref{fig:fairness:allbenefit}, and \cref{fig:fairness:parity}.

%% file: arxiv/text/appendix/calibration.tex
\section{Calibration}\label{app:calibration}

We precisely follow the setup of \citet{minderer2021revisiting}: Since temperature scaling~\citep{guo2017calibration} requires some held-out data, we use 20\% of the ImageNet validation set to learn the temperature parameter while we report the accuracy and expected calibration error on the remaining 80\%.

Moreover, since the expected calibration error is defined with respect to a probability distribution normalised over the classes, we use a \texttt{softmax} loss function during fine tuning. The \texttt{sigmoid} loss function is defined independently across the classes and does not yield the required normalisation. We use 20k steps together with a learning rate of 0.03. 

We reuse the plotting tools provided at \url{https://github.com/google-research/robustness_metrics/tree/master/robustness_metrics/projects/revisiting_calibration}.

%% file: arxiv/text/appendix/plex.tex
\section{Plex}
\label{app:plex}

\subsection{Details about the evaluation}

We start by providing some details about the datasets and the different evaluation protocols based on \citet{djolonga2020robustness}.

\paragraph{ImageNet-C~\citep{hendrycks2019benchmarking}.} This variant of the ImageNet dataset contains algorithmically generated corruptions (e.g., blur and noise) applied to the ImageNet test-set. The results that we report in the paper are averaged over the 16 corruptions and over their 5 different intensity levels.

\paragraph{OOD detection~\citep{hendrycks2019scaling,fort2021exploring}.} In this task, we try to classify whether a given test point belongs to the in-distribution dataset (in our case, ImageNet) or an out-of-distribution dataset (following \citet{hendrycks2019scaling, tran2022plex}, we take Places365 which consists of about 1.8 million images from 365 scene categories, where there are at most 5000 images per category~\citep{zhou2017places}).

To perform the detection, we use the maximum softmax probability (MSP)~\citep{hendrycks2019scaling, tran2022plex}. We evaluate the performance of the resulting binary classification task thanks to the AUROC and AUPRC.

\paragraph{Selective prediction.}
In this task, a model may defer its predictions to human experts when it is not confident enough.
In particular, this task jointly assesses a model’s predictive performance and quality of
uncertainty estimates~\citep{el2010foundations}. 
Following \citet{tran2022plex}, we measure the performance with the oracle collaborative AUC~\citep{kivlichan2021measuring}, with a review fraction of 0.5\% of all predictions.

\paragraph{Label uncertainty.} For this evaluation, we aim at demonstrating the ability of the model to capture the inherent ambiguity of image labels assigned by humans. Following \citet{tran2022plex}, we focus on the ImageNet ReaL-H dataset that exploits the human ratings from \citet{beyer2020we} to construct a label
distribution representing rater uncertainty for each image. The performance is measured by the negative log likelihood computed with respect to the soft labels (i.e., vectors in the simplex as opposed to the usual one-hot vectors).

\subsection{Details about the Plex architecture}

Plex~\citep{tran2022plex} calls for BatchEnsemble layers~\citep{wen2019batchensemble} to be added
in
the model architecture during both pre-training and fine-tuning.\footnote{In \citet{tran2022plex}, the BatchEnsemble layers are added only to a few of the last layers of the encoder in order to reduce the computational and memory cost. The efficient implementation of ViT-22B constrains us to apply BatchEnsemble layers \textit{throughout} the network.} Due to the high cost of training ViT-22B, we add the BatchEnsemble layers during the fine-tuning stage only. We replace all \texttt{Dense} layers in the ViT-22B, except for the \texttt{Dense} layer in the MLP layer for the pooling head with BatchEnsemble layers. \citet{tran2022plex} further suggest to replace the final \texttt{Dense} layer of the network with a heteroscedastic output head~\citep{collier2021correlated}. We thus follow this approach and evaluate both a heteroscedastic and BatchEnsemble final layer.

\subsection{Details about the hyperparameters}

All models were fine-tuned on ImageNet with a batch size 512. We swept over fine-tuning for 20k or 40k steps and learning rates of 0.01 and 0.03, with Plex models performing better at 40k fine-tuning steps---as already observed by \citet{tran2022plex}---and learning rate of 0.03. Two BatchEnsemble members were used, with a random sign initialization in the BatchEnsemble layer of -0.5.

For the experiments with a heteroscedastic output layer, 1k MC samples were used and the low-rank component of the covariance matrix employed 15 factors. Furthermore, we report results for a temperature parameter of 5 (after a hyperparameter search over the [0.5, 10] range).

Unlike most of the models in the rest of the paper, the models of this section are fine tuned with a \texttt{softmax} loss function. We do so to be consistent with the design choices of \citet{tran2022plex} and because a distribution normalised across the classes is required by several of the metrics employed (e.g., ECE). 

\subsection{Results of Plex-22B and challenges}

In \cref{tab:full_plex}, we report the results of ViT-L/32, Plex-L/32, ViT-22B and the extensions of Plex to the 22B scale, Plex-22B, with the BatchEnsemble (BE) and heteroscedastic (HET) heads. All the models are fine tuned with a resolution of 384. 

The main observation is that the increased scale of ViT-22B comes with substantial improvements across all metrics, except for the label uncertainty over ImageNet-ReaL-H.

More surprisingly, we can see that across all metrics (except for the label uncertainty over ImageNet-ReaL-H), the Plex-22B variants perform worse than the vanilla ViT-22B model. This observation does not extend the findings from \citet{tran2022plex} where Plex consistently leads to improvement at the S, B and L scales.

We believe that this surprising observation may be related to specific challenges faced at the 22B scale:
\begin{itemize}
    \item \textbf{Pre-training vs.~fine-tuning}: While \citet{tran2022plex} introduce BatchEnsemble layers already at pre-training time, the high training cost of ViT-22B forces us to only operate at fine-tuning time. In this regime, it may not be possible to properly learn the BatchEnsemble and heteroscedastic layers. 
    Moreover, while fine-tuning with standard ViT backbones enjoys a well-performing and robust recipe, namely  initializing the final \texttt{Dense} layer kernel to all zeros, we do not have an equivalent approach when adding the Plex components. 
    \item \textbf{Hyperparameter tuning}: Even though we already covered a reasonable combination of hyperparameters (fine-tuning duration, learning rate and temperature), it is possible that a finer-grained search is required to close the performance gap.
    \item \textbf{Numerical stability}: As discussed in \cref{sec:model_architecture}, it was required to use particular techniques to stabilize the training of ViT-22B. We hypothesise that similar techniques may have to be developed specifically for the Plex components (BatchEnsemble and heteroscedastic layers) to keep their efficiency at this scale.
\end{itemize}

\begin{table}[t]
    \caption{Evaluation on some representative metrics from the Plex reliability benchmark~\citep{tran2022plex}.}
\centering
  \setlength{\tabcolsep}{4pt}
  \resizebox{0.8\columnwidth}{!}{%
    \begin{tabular}{@{} l @{} c ccc c ccc cc @{}}
      \toprule
       && \multicolumn{4}{c}{IN-C (mean over shifts)} && \multicolumn{2}{c}{IN vs.~Places365} && \multicolumn{1}{c}{IN-ReaL-H} \\
\cmidrule{3-6}\cmidrule{8-9}\cmidrule{11-11}
	      Metrics && ACC $\uparrow$ & NLL $\downarrow$ & ECE $\downarrow$ & OC-AUC $\uparrow$ && AUROC $\uparrow$ & AUPRC $\uparrow$ && NLL $\downarrow$ \\
      \midrule
      ViT-L/32~\citep{tran2022plex} &&  70.1 & 1.28 &  0.05 & 0.91 && 0.83 &  0.96 && 1.09 \\
      Plex-L/32~\citep{tran2022plex} &&  71.3 & 1.21 & 0.02 &  0.91 && 0.83 &  0.97 && 1.03\\
      ViT-22B && 83.7 & 0.63 &  0.01 & 0.97 && 0.88 &  0.98  && 1.21\\
      Plex-22B [BE] && 81.0 &     0.98 &     0.18 & 0.95 && 0.86 &  0.98 && 0.94\\
      Plex-22B [HET] && 80.9 & 0.97 & 0.17 & 0.94 && 0.86 & 0.97 &&  0.93 \\
      \bottomrule
 \end{tabular}}
    \label{tab:full_plex}
\end{table}

%% file: arxiv/text/appendix/analysis.tex
\section{Error Consistency \& Human Alignment}
\begin{figure}[ht]
    \centering
    \subfigure[][OOD accuracy (higher = better).]{
        \label{fig:model_vs_human_benchmark_a}
        \includegraphics[height=1.6in]{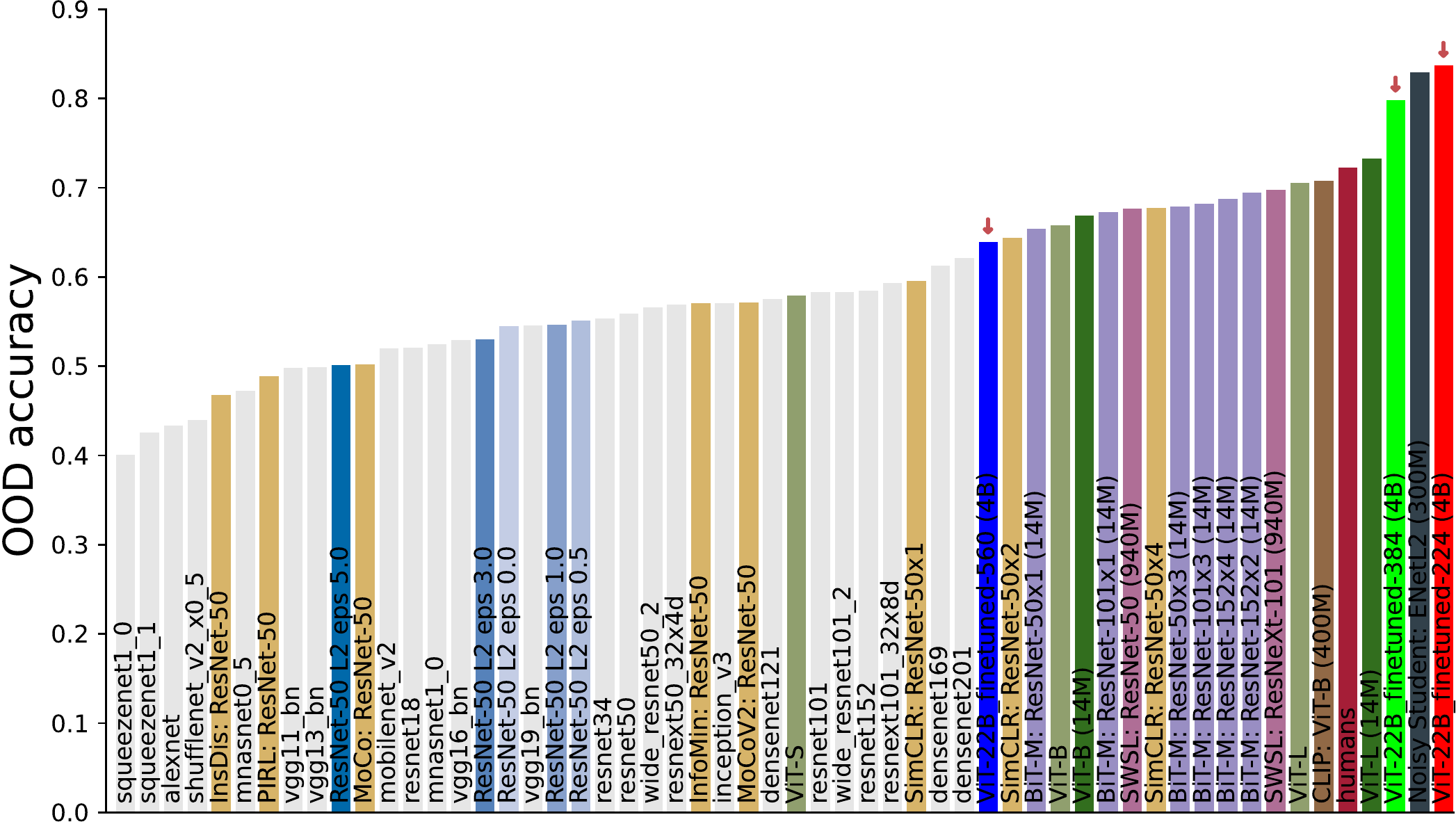}}
    \hspace{8pt}
    \subfigure[][Accuracy difference (lower = better).]{
        \label{fig:model_vs_human_benchmark_b}%
        \includegraphics[height=1.6in]{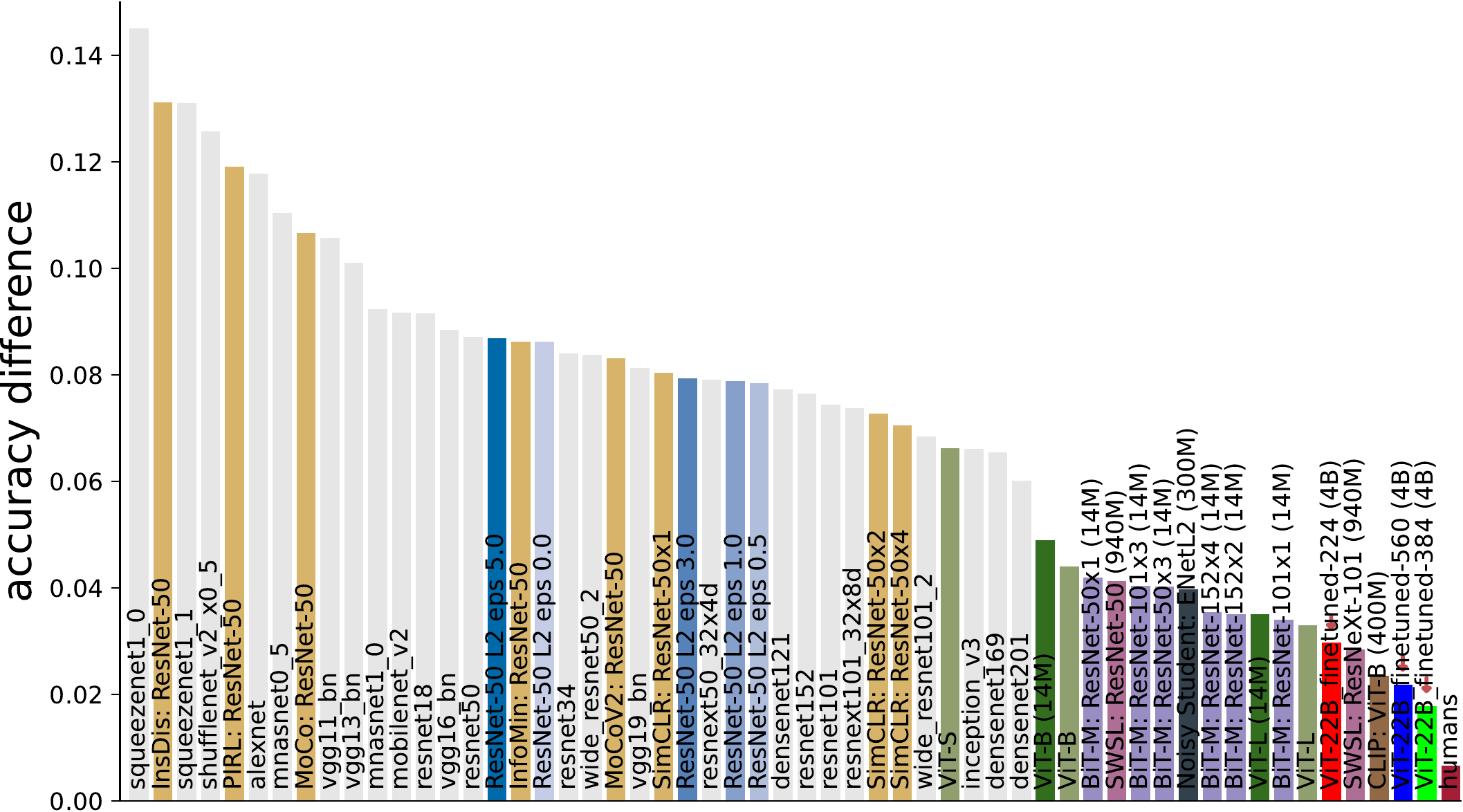}} \\
    \subfigure[][Observed consistency (higher = better).]{
        \label{fig:model_vs_human_benchmark_c}
        \includegraphics[height=1.6in]{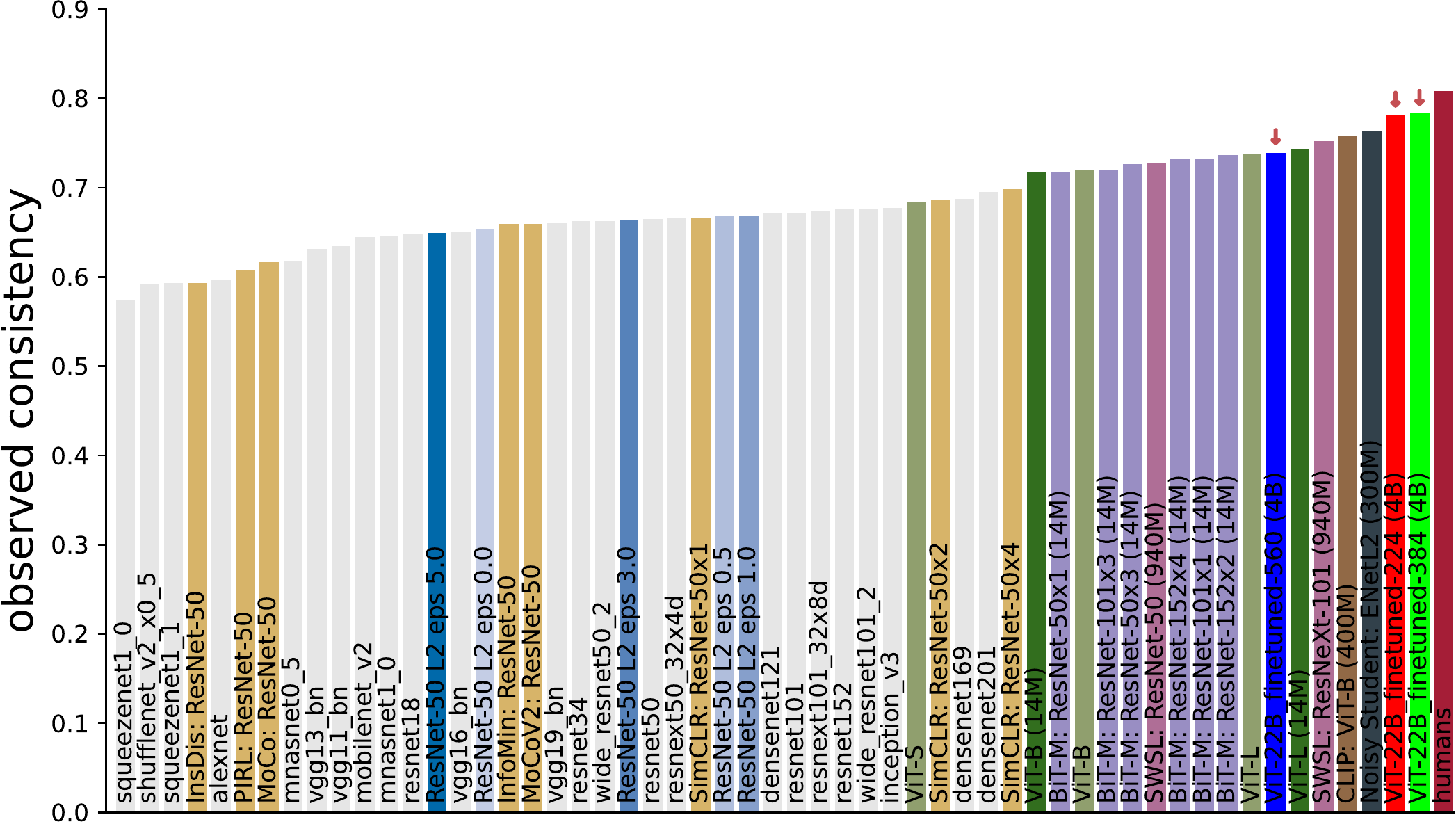}}
    \hspace{8pt}
    \subfigure[][Error consistency (higher = better).]{
        \label{fig:model_vs_human_benchmark_d}
        \includegraphics[height=1.6in]{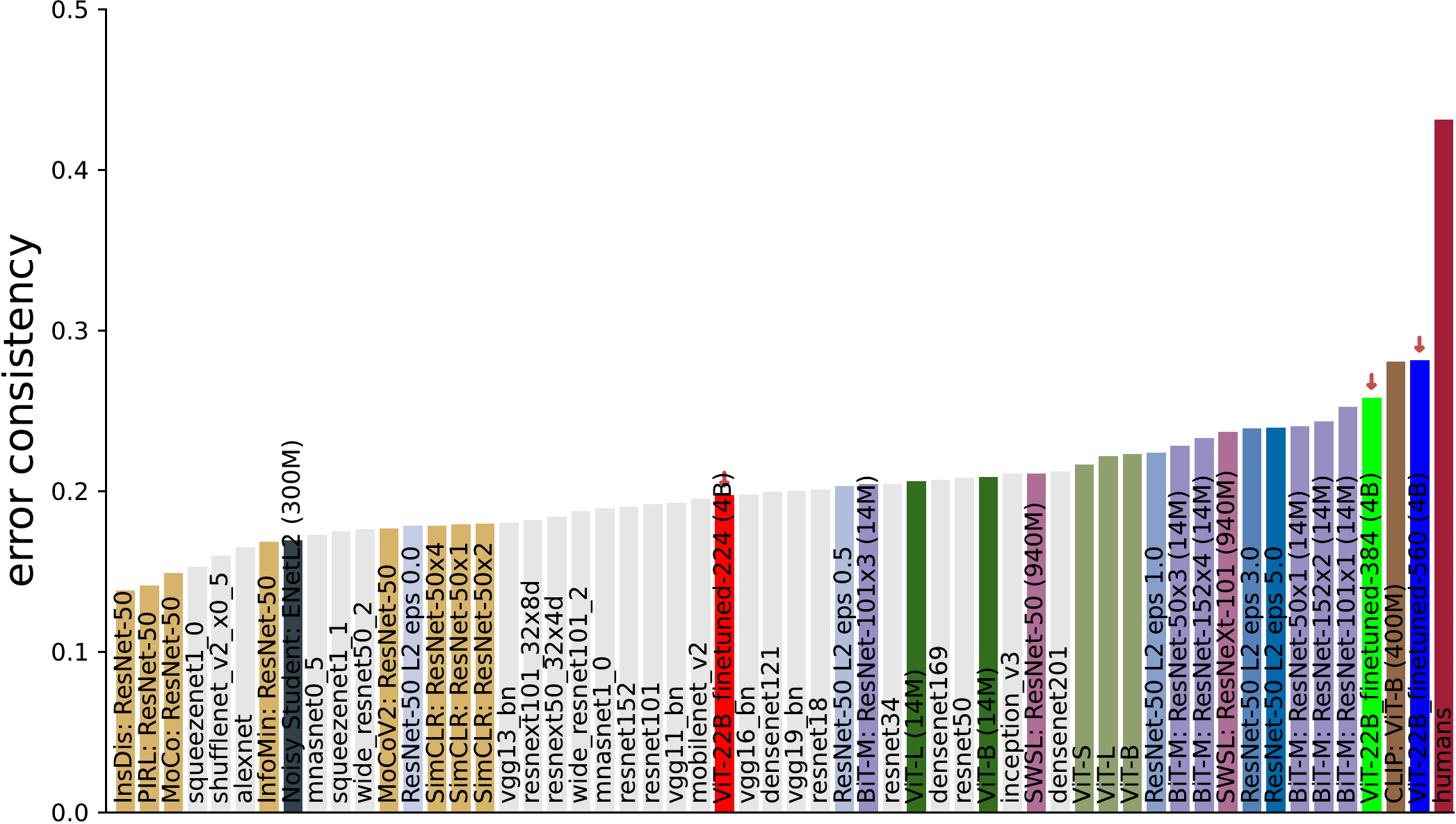}}
    \caption[]{Model-vs-human~\citep{geirhos2021partial} benchmarking results for ViT-22B models fine-tuned on ImageNet with different resolutions (indicated by a red arrow). Results are aggregated over 17 challenging out-of-distribution (OOD) datasets from~\citep{geirhos2021partial}. OOD accuracy in \cref{fig:model_vs_human_benchmark_a} is simply the accuracy across those datasets; accuracy difference in \cref{fig:model_vs_human_benchmark_b} denotes the difference to human accuracies (either too low or too high is penalized); observed consistency in \cref{fig:model_vs_human_benchmark_c} shows unnormalized image-level consistency~\citep[see][for details]{geirhos2021partial} while \cref{fig:model_vs_human_benchmark_d} shows error consistency (normalizes observed consistency by the consistency expected purely by change). Error consistency is only above zero if models and humans systematically agree in terms of which images are easy/difficult (correct/incorrect classification). Overall, while the three ViT-22B variants trained with different resolutions vary in their performance, a ViT-22B variant leads the leaderboard in all four metrics. Comparison models include standard CNNs (grey), adversarially trained models (blue), self-supervised models (orange) as well as other models evaluated by \citet{geirhos2021partial}.}
    \label{fig:model_vs_human_benchmark}
\end{figure}

In \cref{subsec:error consistency}, we described results for testing ViT-22B fine-tuned on ImageNet on the \texttt{model-vs-human} benchmark. In \cref{fig:model_vs_human_benchmark_a}, \cref{fig:model_vs_human_benchmark_b}, \cref{fig:model_vs_human_benchmark_c}, \cref{fig:model_vs_human_benchmark_d}, we provide additional benchmarking results.

\section{Perceptual similarity}
\label{app:perceptual_similarity}

\citet{kumar2022do} show a trade-off between the accuracy of latest ImageNet classifiers and their inherent ability to capture perceptual similarity. Here, we explore if large-scale classification on a more diverse training dataset than ImageNet can break the observed trade-off. To compare the perceptual similarity of \chonk with prior ImageNet-trained models, we make minor changes to adapt \chonk on low resolution $64 \times 64$ ImageNet. \chonk fine-tuned on ImageNet $64 \times 64$ achieves 84.2 accuracy on ImageNet $64 \times 64$ which is $16 \%$ better than the best models trained directly on ImageNet. As done in~\citep{zhang2018perceptual}, we assess the ability of \chonk to capture perceptual similarity using 48 intermediate representations. The perceptual score of \chonk (64.9) is much lower than all other models, indicating that models trained on large-scale classification also lie on the observed accuracy-perceptual similarity Pareto Frontier.

\begin{figure}[htbp]
    \centering
    \includegraphics[width=0.4\columnwidth]{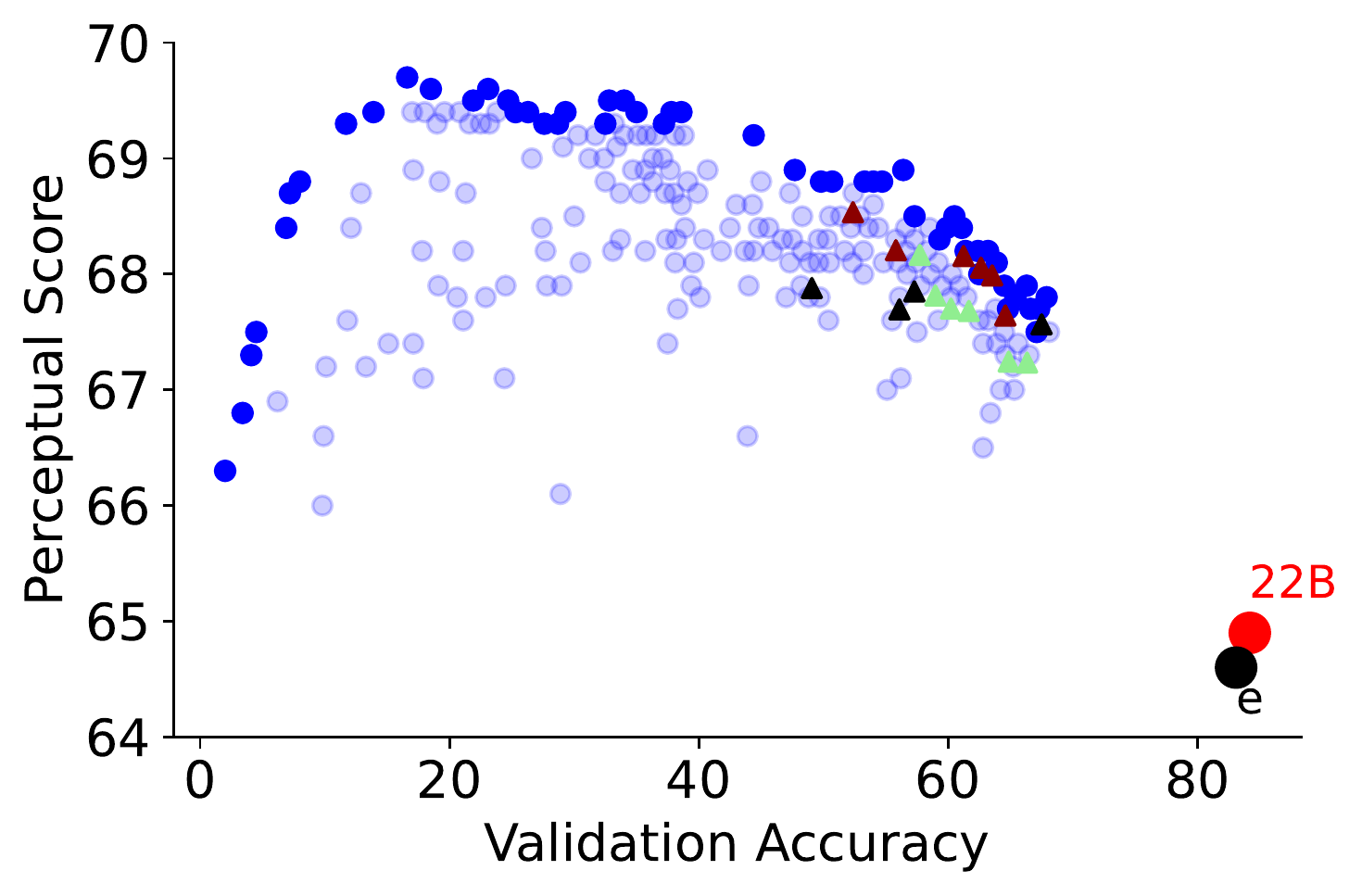}
    \caption{\chonk lies on the bottom-right of the accuracy-perceptual similarity tradeoff. It achieves the best validation accuracy on $64 \times 64$ ImageNet with the worst perceptual scores.}
    \label{fig:ps_vit22B}
\end{figure}

To make a fair comparison with the models in~\citep{kumar2022do}, we make minor changes to adapt ViT-22B on low resolution $64 \times 64$ ImageNet. Directly finetuneing ViT-22B on $64 \times 64$ images with the default patch-size of 14 leads to two undesirable consequences a) A low sequence length of 16 and b) Cropping of 8 pixels on the right borders. So, as proposed in~\citep{beyer2022flexivit}, we resize the trained embedding layer from the default patch-size of 14 to a patch-size of 8 that leads to a longer sequence length of 64. Then, we adapt standard finetuneing protocols.

We make three more observations: 1) Untrained ViT-22B gets a even lower Perceptual Score of 62.3, thus some amount of training is desirable 2) ViT-e lies in the same ballpark as ViT-22B with slightly lower accuracy and Perceptual Scores 3) ViT-22B with the newly proposed Mean Pool distance function~\citep{kumar2022do} can improve its Perceptual Score up to 66.2.

\section{Feature attribution analysis}
\label{app:feature_attribution}
\begin{figure}[h]
    \centering
    \includegraphics[width=0.5\textwidth]{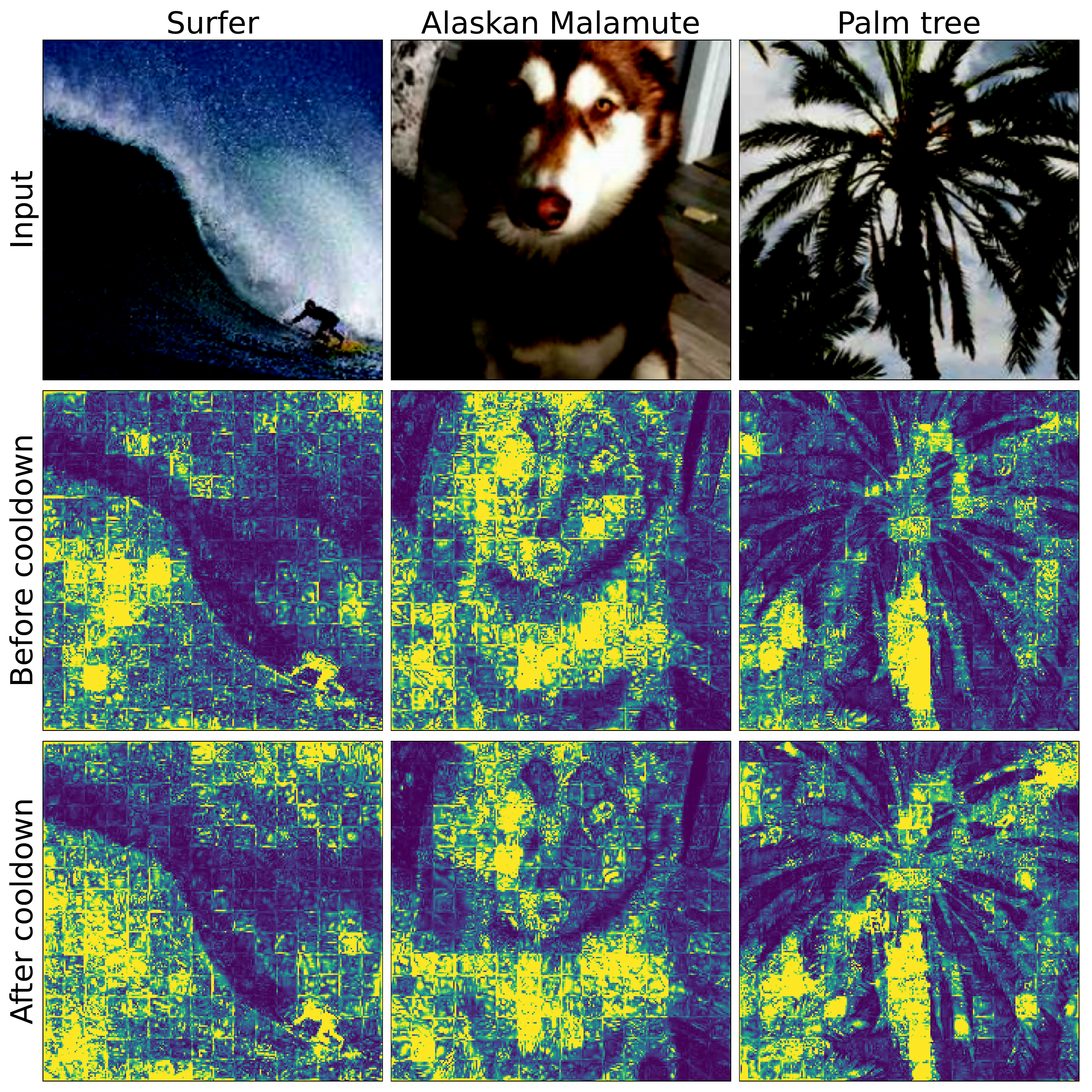}
    \caption{Saliency before and after model cooldown.}
    \label{fig:featureattribution}
\end{figure}

To get a better understanding on how \chonk arrives at its predictions we make use of gradient-based feature attribution methods (a.k.a.\ saliency maps). 
\cref{fig:featureattribution} shows the result of applying Integrated Gradients~\citep[][IG]{sundararajan2017axiomatic} to three example datapoints before and after \chonk cooldown. We find that using a gray (0.5) baseline and 1024 steps yields qualitatively the best results. 
The images show a subtle difference in how the two model checkpoints process the example inputs, where more yellow indicates a higher sensitivity.
We can also clearly see the patches in which \chonk processes input images. This means that the model is less sensitive around the edges of each patch, and suggests a path for future work to improve the model to better deal with patch edges.

